\documentclass{bmvc2k}
\usepackage{times}
\usepackage{amsmath}
\usepackage{amssymb}

\DeclareMathAlphabet{\pazocal}{OMS}{zplm}{m}{n}

\newcommand{\Lb}{\pazocal{L}}
\usepackage{wrapfig}
\usepackage{booktabs}
\usepackage{adjustbox}
\usepackage{array}
\usepackage{multirow}
\usepackage{dsfont}
\usepackage{scalerel}
\usepackage{lipsum}
\usepackage{wrapfig}
\usepackage{blindtext}
\usepackage[font=small]{caption, subcaption}

\newcommand\blfootnote[1]{%
  \begingroup
  \renewcommand\thefootnote{}\footnote{#1}%
  \addtocounter{footnote}{-1}%
  \endgroup
}

\newcolumntype{x}[1]{>{\centering\arraybackslash}p{#1}}

\usepackage[labeled,resetlabels]{multibib} 
\newcites{S}{References} 

\title{Improving Object Detection via Local-global Contrastive Learning}

\addauthor{Danai Triantafyllidou}{danaitri22@gmail.com}{$1^\mathbf{\ast}$,$\mathbf{\dag}$}
\addauthor{Sarah Parisot}{sarah.parisot@huawei.com}{1}
\addauthor{Ales Leonardis}{a.leonardis@cs.bham.ac.uk}{2,$\mathbf{\dag}$}
\addauthor{Steven McDonagh}{s.mcdonagh@ed.ac.uk}{3,$\mathbf{\dag}$}

\addinstitution{
Huawei Noah's Ark Lab
}
\addinstitution{
University of Birmingham\\
Birmingham, UK
}
\addinstitution{
 University of Edinburgh\\
 Edinburgh, UK\\
}

\runninghead{TRIANTAFYLLIDOU, PARISOT, LEONARDIS, MCDONAGH}{}

\def\eg{\emph{e.g}\bmvaOneDot}

\begin{document}

\maketitle

\blfootnote{$\mathbf{\dag}$ Work partially done at Huawei Noah’s Ark Lab}
\blfootnote{$\mathbf{\ast}$ Currently at Kittl Technologies}

\begin{abstract}

Visual domain gaps often impact object detection performance. Image-to-image translation can mitigate this effect, where contrastive approaches enable learning of the image-to-image mapping under unsupervised regimes. However, existing methods often fail to handle content-rich scenes with multiple object instances, which manifests in unsatisfactory detection performance. Sensitivity to such instance-level content is typically only gained through object annotations, which can be expensive to obtain. Towards addressing this issue, we present a novel image-to-image translation method that specifically targets cross-domain object detection. We formulate our approach as a contrastive learning framework with an inductive prior that optimises the appearance of object instances through spatial attention masks, 
implicitly delineating the scene into foreground regions associated with the target object instances and background non-object regions. Instead of relying on object annotations to explicitly account for object instances during translation, our approach learns to represent objects by contrasting local-global information. This affords investigation of an under-explored challenge: obtaining performant detection, under domain shifts, without relying on object annotations nor detector model fine-tuning. We experiment with multiple cross-domain object detection settings across three challenging benchmarks and report state-of-the-art performance. 
\newline 
\newline
\noindent Project page: \footnotesize{\url{https://local-global-detection.github.io}}

\end{abstract}

\section{Introduction}
\label{sec:intro}

Deep learning based object detection has become an indispensable part of many computer vision applications such as autonomous navigation. 
State-of-the-art detection models typically rely on large-scale annotated data in order to learn representative features and yet often fail to generalize well to new target domains that exhibit visual disparity, (such as foggy vs.~clear weather scenes), with common benchmarks typically reporting falls in detection accuracy in excess of $20\%$ (see Sec.~\ref{sec:experiments:eval:quant} for details). Performance drops drastically due to the domain shift problem. 
Image-to-image (I2I) translation aims to mitigate such domain gaps at the input level and thereby reduce the distribution shift in the visual domain. Such approaches enable an existing (i.e.~pre-trained) detector, trained on the source domain, to function well on \emph{source-like} images, translated from the target domain. It has been evidenced that this process is able to improve target domain detection performance~\cite{rodriguez_2019,zhang2019cycle,chen2020harmonizing}. 
\begin{figure*}
    \centering
    \begin{subfigure}
        \centering
        \includegraphics[width=3cm]{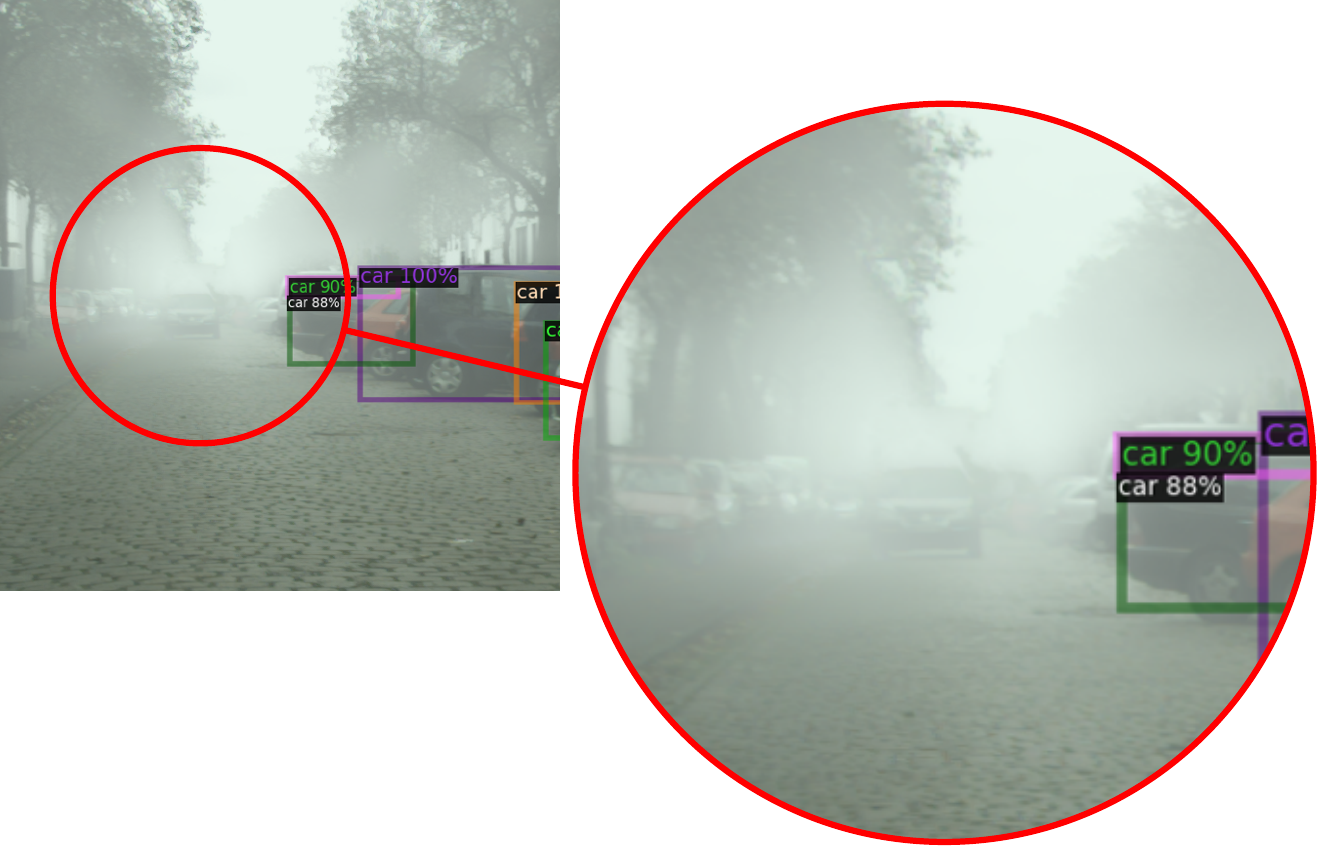}
    \end{subfigure}
    \begin{subfigure}  
        \centering 
        \includegraphics[width=3cm]{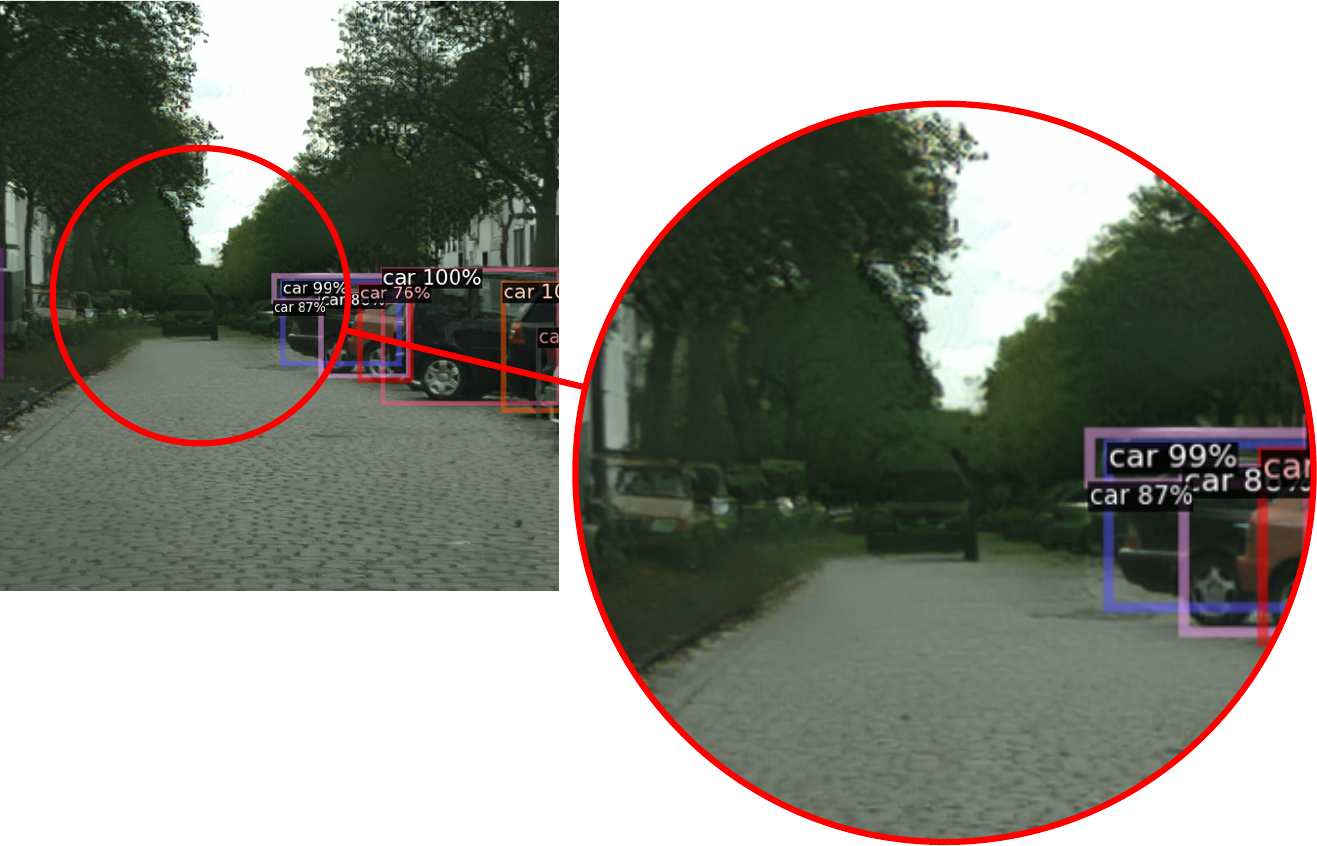}
    \end{subfigure}
    \begin{subfigure}   
        \centering 
        \includegraphics[width=3cm]{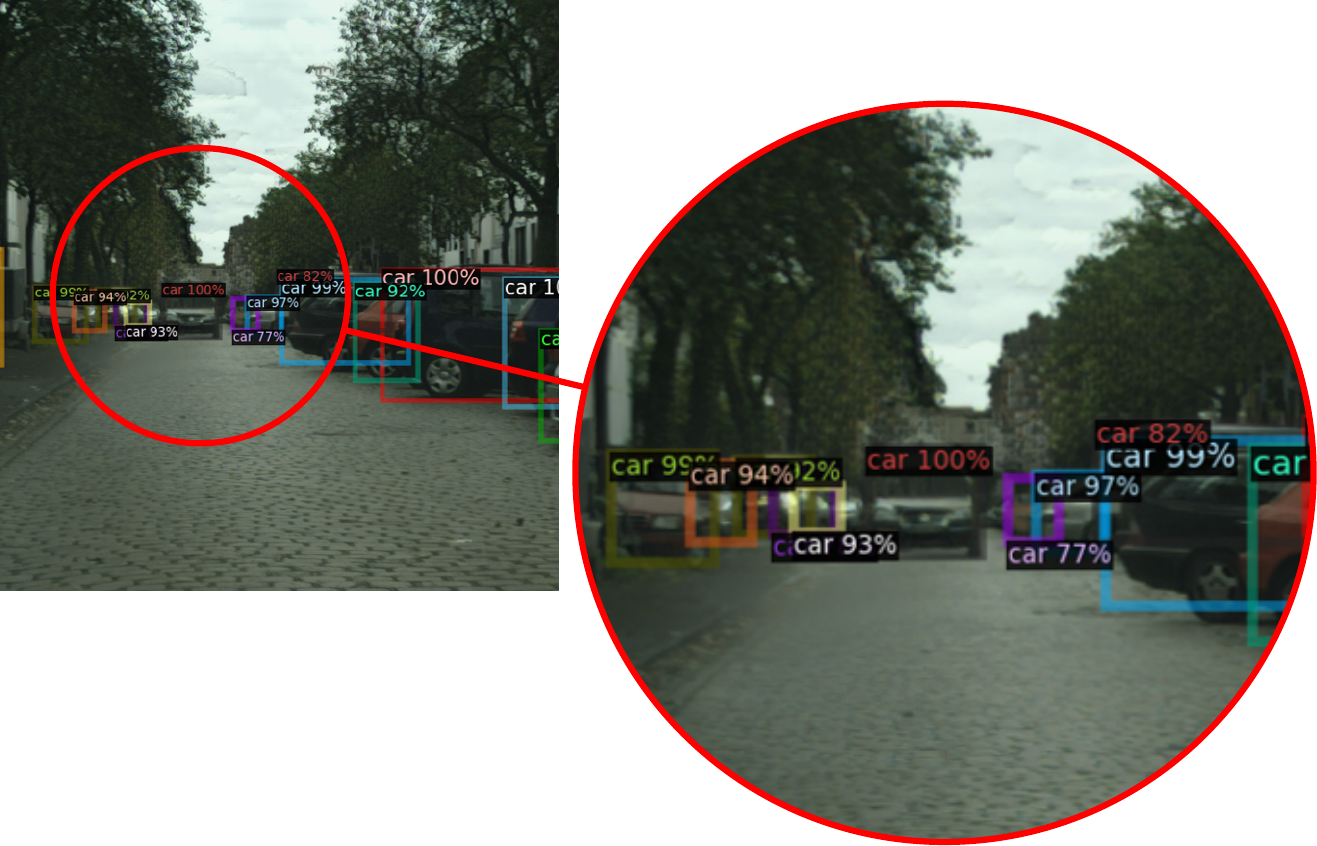}
    \end{subfigure}
    \caption{\textbf{Left:} visual domains, unseen during object detector training, hurt detection performance. \mbox{\textbf{Middle:}} global image-to-image translation (foggy $\rightarrow$ clear weather) provides some benefit to downstream detection performance, yet homogeneous image translation strategies result in small objects, with low contrast regions, that remain undetectable. \textbf{Right:} Our local-global approach is guided to better delineate objects during translation and thus cross-domain detection is improved.}
    {} 
    \label{fig:teaser}
\end{figure*}

The high costs related to (domain-wise) paired image data collection have steered community interest towards unpaired I2I translation. Pioneering unpaired image translation work has made use of cycle-consistency~\cite{zhu_park_isola_efros_2017} and shared latent space assumptions~\cite{jeong_kim_lee_sohn_2021b}. Such methods have become the de facto translation modules in many works. However, they can often lead to severe content distortions and shape deformation as they assume a bijective relationship between source and target domains~\cite{pang2021image}.
Such failures to ensure content preservation may in turn adversely affect performance in downstream object detection tasks~\cite{oza2021unsupervised}, where this effect is exemplified in Fig.~\ref{fig:teaser}.

Explicitly accounting for object instances has provided an intuitive direction for improving image translation in spatial regions that are critical to down-stream detection \cite{bhattacharjee2020dunit,shen2019towards,shen2021cdtd,jeong_kim_lee_sohn_2021a}. 
However, these works rely on object annotations in order to treat object and background spatial image regions distinctly in the target domain, which fundamentally limits their applicability. In cases where strong pre-trained detectors exist, 
yet object labels are inaccessible 
or are otherwise 
infeasible to obtain, such strategies become unsuitable. 
We offer a new perspective and alternatively consider the scenario where labels are unavailable, an under-explored and yet practical problem setting, where we further propose a method 
to account for this gap in the literature. 

Contrastive learning 
has emerged as a promising strategy for solving I2I translation, through the maximisation of mutual information between corresponding input and output patches~\cite{park2020contrastive,zheng_cham_cai_2021,hu_zhou_huang_shi_sun_li_2022}. While recent contrastive-based translation results report promising increases for standard image quality metrics~\cite{heusel2017gans,rabin2012wasserstein}, 
these approaches consider image translation as a \emph{global} task; i.e.~a translation problem where all image regions are treated uniformly. Framing translation in this manner can lead to unsatisfactory outcomes when considering object-rich images with complex local structures; the visual disparity between objects and background is often 
large.  We hypothesize that 
implicitly modeling background and foreground object regions can enhance translation quality in local salient areas, significantly improving downstream object detection. Additionally, we propose that the separation of foreground and background can be accomplished through local-global contrastive learning.

Motivated by this intuition, we propose a contrastive learning-based I2I translation framework for cross-domain object detection. We introduce an architectural inductive prior that optimises object instance appearance using spatial attention masks, effectively disentangling scenes into background and foreground regions. We note that while previous studies~\cite{10030413} have employed the ``background / foreground'' terminology to describe explicit content separation based on object annotations, our approach learns to delineate content under unsupervised conditions\footnote {Despite not using object annotations, our empirical results demonstrate effective separation of content, leading us to adopt this terminology.}. 
Inspired by the recent success of region-based representation learning~\cite{xie_ding_wang_zhan_xu_sun_li_luo_2021,wang_zhang_shen_kong_li_2021,yang_wu_zhou_lin_2021}, we alternatively rely on contrasting local and global views to learn discriminative representations and separate content. Our main contributions can be summarised as:

\begin{itemize}

\item We propose a novel attention guided I2I translation framework for cross-domain object detection. Our approach encourages the model to optimise local image region appearance without requiring object annotations and can be used in conjunction with a frozen pre-trained object detector.

\item We illustrate how the idea of local-global contrastive learning can be used to improve image-to-image translation for object detection: 
implicitly differentiating between objects and background image regions gives rise to 
robust translation of object image regions, amenable for detection tasks.

\item We conduct extensive experiments in common domain adaptation and detection settings, reporting 
state-of-the-art performance under three visual adaptation scenarios. 
\end{itemize}

\section{Related work}
\label{sec:related}

We briefly review topics most relevant to our core ideas and refer 
to~\cite{pang2021image,oza2021unsupervised} for extensive surveys on both image-to-image translation and 
domain adaptation for object detectors.
 
\noindent \textbf{Instance-aware I2I translation.}
Instance-aware I2I translation has recently garnered interest, towards enabling models to translate objects and background 
areas separately. 
Shen et al.~\cite{shen2019towards,shen2021cdtd} perform translation with distinct encoder-decoder blocks to generate separate object, background and global image style codes and provide the model with object-specific guidance in relation to translation. Following the idea of distinct instance region encodings, Bhattacharjee et al.~\cite{bhattacharjee2020dunit} propose to jointly learn image translation and detection, therefore focussing on certain objects during translation. A class-aware memory network was used in~\cite{jeong_kim_lee_sohn_2021a} to store features and retain individual object styles, thus improving translation for images with multiple objects. The recent work of~\cite{kim2022instaformer} performs instance-aware I2I 
using a transformer model, trained with contrastive learning. All of these works crucially assume access to object annotations during training in order to guide the translation.

\noindent \textbf{Trainable cross-domain object detectors.} In contrast to I2I, cross-domain detection solutions integrate  Unsupervised Domain Adaptation (UDA) techniques, within object detection pipelines. An extensive set of cross-domain object detector training strategies exist; adversarial feature learning for domain invariant representations, pseudo-labels for self-training, graph reasoning and domain randomization, among others~\cite{sindagi_oza_yasarla_patel_2020,chen_zheng_huang_ding_yu_2021,kim_choi_kim_kim_2019,roychowdhury_chakrabarty_singh_jin_jiang_cao_learned-miller_2019,xu2020cross,wu_han_zhu_yang_2021,rodriguez_2019,shen2021cdtd,hsu2020every,wang_liao_shao_2021,chen2020harmonizing,lin2021domain,wang2021domain,deng_li_chen_duan_2021,liu_li_yang_li_yuan_2022,zhao_wang_2022,rezaeianaran2021seeking,li2022sigma,li_liu_yuan_2023}. 
This tranche of works fundamentally involve training or otherwise adapting a detector model. However, adaptation strategies typically assume that source domain object labels remain available 
and such methods are also known to suffer from catastrophic forgetting problems~\cite{volpi2021continual}. We consider direct comparison with this class of methods under common experimental settings to offer useful insight, with respect to investigation of method efficacy and related trade-offs (further details are found in Sec.~\ref{sec:experiments}).

\noindent \textbf{Contrastive learning.} Contemporary self-supervised learning aims to exploit the underlying structures in the data and build unsupervised visual representations; either by solving generative pretext tasks (e.g.~colourisation, inpainting, jigsaw puzzles) or through contrastive learning. Contrastive learning has shown great potential when performing instance discrimination tasks \cite{pmlr-v119-chen20j,Zhuang2019LocalAF,Chen2020ImprovedBW}, where the objective is formed by generating different views of an image and maximising their similarity through data augmentation. The success of such methods is mainly attributed to the ability of contrastive learning to encode semantic priors across different images~\cite{purushwalkam2020demystifying}. 
More recently, there has been increased interest in region-based representation learning, shifting the focus to learning local descriptors that are relevant for dense prediction tasks such as image segmentation and object detection. 
Indeed, global-local and multi-scale crop strategies have proven popular in self-supervised and unsupervised (contrastive) learning scenarios where prevalent works include BYOL~\cite{grill2020bootstrap} and DINO~\cite{zhang2022dino}.
Learning region-level representations has been realised through image segmentation masks~\cite{van_gansbeke_vandenhende_georgoulis_van_gool_2021,10.5555/3495724.3497115,henaff_koppula_alayrac_van_denoord_vinyals_carreira_2021} and by 
contrasting between local patches and global image views~\cite{xie_ding_wang_zhan_xu_sun_li_luo_2021}. 
These recent successes lead contrastive learning to become a prevailing component of self-supervised learning and particularly successful in pre-training a strong feature extractor for several local and global discriminative tasks.

Park et al.~\cite{park2020contrastive} 
employed a contrastive approach to the I2I translation task and enforce their model to preserve structure in corresponding input and output spatial locations. Zheng~et al.~\cite{zheng_cham_cai_2021} further improve the structure consistency constraint by contrasting self-similarity patches. Huet et al.~\cite{hu_zhou_huang_shi_sun_li_2022} proposed an entropy based query selection mechanism, towards enabling feature selection that better reflects domain specific characteristics. Jung et al.~\cite{jung_kwon_ye_2022} enable semantic awareness in a contrastive setting through the exploitation of 
semantic relation consistencies across image patches. However we note that these methods lack in-built mechanisms to exploit \emph{instance-level} information, specifically relating to semantic objects in a scene. We foresee this as a potential shortcoming for downstream detection tasks and experimentally evidence this conjecture (Sec.~\ref{sec:experiments:eval:quant}).


\section{Method}
\label{sec:method}

\subsection{Preliminaries}
\label{sec:method:prelim}
\textbf{Self-supervised representations} 
are realised under a contrastive learning regime by considering a dictionary look-up task. Specifically, given an encoded query $q$, the task is to identify which single positive key $k_{\scaleto{+}{3pt}}$ matches the query $q$ among a set of encoded keys $\{k_0, k_1, ...\}$. The InfoNCE loss function~\cite{oord_li_vinyals_2019} is employed to attract $q$ close to $k_{\scaleto{+}{3pt}}$ while pushing it away from a set of alternative negative keys $\{k_{\scaleto{-}{3pt}}\}$:
\begin{equation}
\Lb^{\textsc{NCE}}
 = -\log\frac{\exp(q \cdot k_{\scaleto{+}{3pt}} \mathbin{/} \tau )}{\exp(q\cdot k_{\scaleto{+}{3pt}} \mathbin{/} \tau  ) + \sum_{k_{-} } \exp(q\cdot k_{\scaleto{-}{3pt}} \mathbin{/} \tau  ) }
\label{eq:nce},
\end{equation}
with $\tau$ a temperature hyperparameter. 
For vision tasks, positive pairs $q$ and $k_{\scaleto{+}{3pt}}$ can be formed by generating two different views from the same image or different views that pertain to a global image and a local patch~\cite{xie_ding_wang_zhan_xu_sun_li_luo_2021}. 

\textbf{Contrastive learning for unpaired I2I translation.}
Contrastive techniques can be leveraged for the I2I translation task by constraining matching spatial locations (image patches) between the input image and translated output image to have high mutual information. In this case, the query patch $q$ is created by encoding a local region of the output image. The positive key $k_{\scaleto{+}{3pt}}$ refers to the corresponding region of the input, while the set of negative keys $\{k_{\scaleto{-}{3pt}}\}$ are selected by encoding different regions of the input image. As such, the contrastive loss of Eq.~\eqref{eq:nce} ensures content and structure consistency between input and translated patches, while the appearance of the output image is enforced using a discriminator, trained with an adversarial loss~\cite{NIPS2014_5ca3e9b1}.

\subsection{Spatial attention for I2I translation}
\label{sec:method:attention}

We assume a performant pre-trained object detector to be available, trained using images from a source domain $Y$, 
with the aim of applying the detector to a new target \emph{detection} domain $X$. 
Our goal is to learn a function capable of performing the (inverse) image translation task \mbox{$\mathcal{G}: X \rightarrow Y$} such that detection performance is significantly improved for images originally belonging to domain 
$X$. 
In contrast to previous works that extract separate representations to encode global and instance-level information, respectively~\cite{kim2022instaformer,bhattacharjee2020dunit,shen2019towards}, our approach alternatively guides translation to focus on the relevant instance regions, using spatial attention masks. The spatial attention masks are generated by a dedicated trainable module and weight the influence of separate image features in a final translation step.

\begin{figure*}
\centering
\includegraphics[width=10cm]{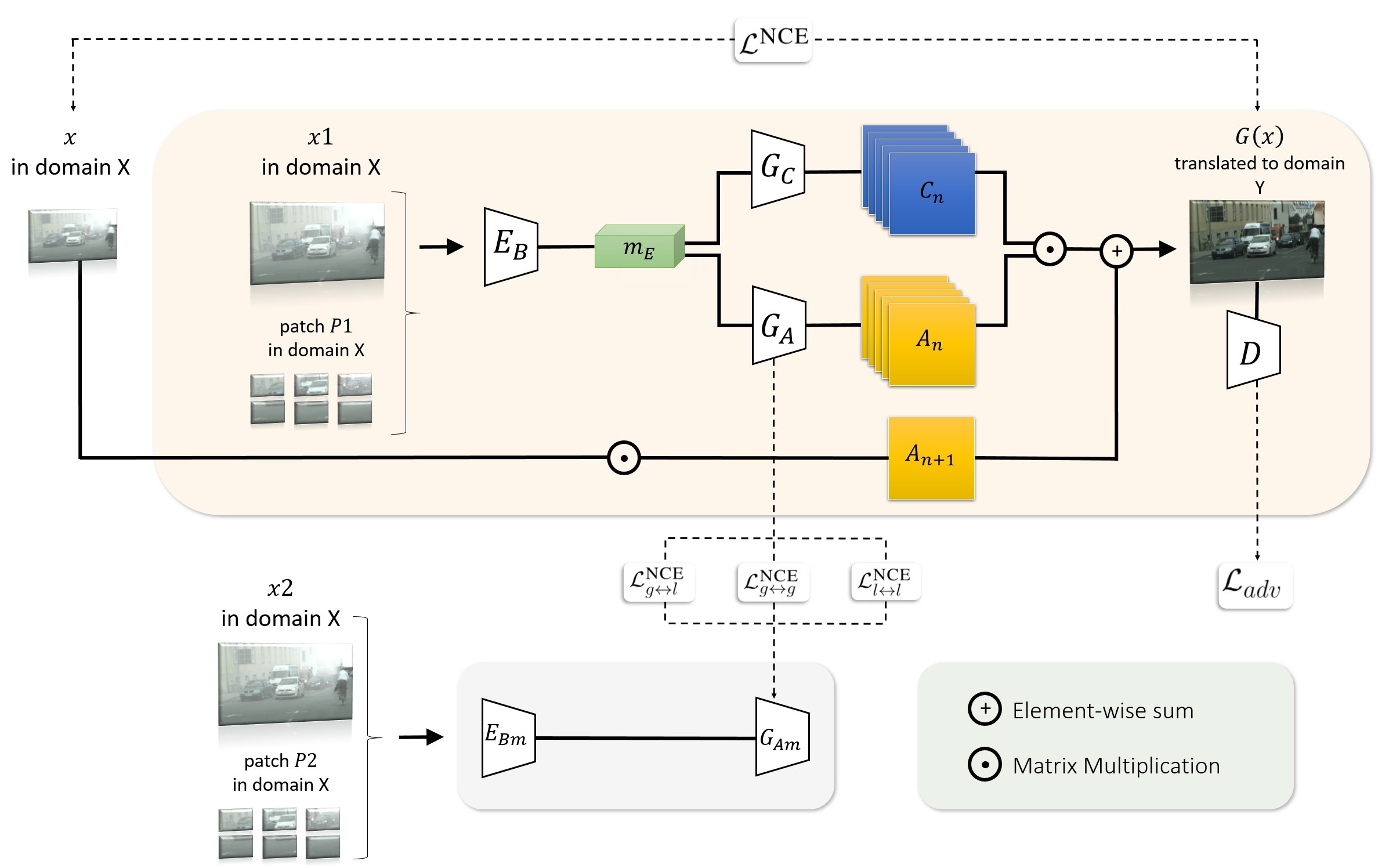}
\caption{Overview of the proposed method - see text for further details.} 
\label{fig:scheme_1}
\end{figure*}

We propose an attention-driven scheme that learns to decompose input image $x$ into foreground and background regions and encourages the translation model to focus on optimising appearance of foreground objects. We adopt an encoder-decoder architecture where the encoder $E_B$ acts as a feature extractor, and generates image representations of lower dimensionality. We decompose our decoder into two components: a content generator $G_C$ that generates multiple image content maps and an attention generator $G_A$ that outputs attention masks. Attention masks enable combination of the generated content maps in a learnable fashion to obtain a final translated image. Fig.~\ref{fig:scheme_1} depicts 
an overview of the proposed method. 

More formally, input image $x$ is first converted into a latent representation via feature extractor \mbox{$E_B$: $m_E =  E_B(x)$}. This representation serves as input to the content and attention generators. The content generator $G_C$ generates a set of $n$ content maps \mbox{$\{C_t \,|\, t \in [0,n{-}1]\}$}. Each layer $l$ of $G_C$ comprises a group of $n$ convolutional filters, such that each filter is associated with a specific content map. Content map $t$ at layer $l$ can be expressed as: \mbox{$C^{t}_l= \sigma(\texttt{Conv}^t(C^{t}_{l-1}))$} with $t=0,\ldots,n{-}1$ and \mbox{$C^t_0 = m_E, \; \forall t$}. 
The activation function $\sigma(\cdot)$ is selected as a 
\texttt{ReLU}~\cite{DBLP:journals/corr/abs-1803-08375} in the intermediate layers, and {$\sigma(\cdot)=\tanh(\cdot)$} in the final layer. Similarly, the attention generator outputs a set of $n{+}1$ attention maps \mbox{$\{A_t \,|\, t \in [0,n]\}$}, using $n{+}1$ convolutional filters per layer, with $\sigma(\cdot)=\texttt{softmax}(\cdot)$ for the last layer. 
Finally, the translated image $G(x)$ is recovered through summation of the generated foreground 
content maps, weighted by the respective attention masks, with an additional explicit component, representing image background content: 
\begin{equation}
G(x)=\sum_{t=1}^{n} \underbrace{(C^{t} \odot A^{t})}_\textrm{\parbox{15mm}{\centering foreground\\[-4pt] }}  
+
\underbrace{(x \odot A^{n+1})}_\textrm{background}.
\label{eq:fb}
\end{equation}

By disentangling the translation task in this manner and explicitly modelling distinct 
spatial regions, we enable our model to actively focus on discriminative 
image locations containing objects. We find that accurate and precise translation of 
such image regions to be of crucial importance for strong detection performance (see Sec.~\ref{sec:experiments:eval:quant} for further details). 

We follow previous contrastive I2I work~\cite{park2020contrastive} and train our generator using the InfoNCE loss found in Eq.~\eqref{eq:nce}, which enforces content and structure consistency at the image patch level. To ensure that translated images match the appearance of source domain $Y$, we further use a discriminator module $D$ trained with a standard adversarial loss:

\begin{equation}
 \Lb_{adv} = -\mathbb{E}_{y \sim Y} \log D(y) 
   -\mathbb{E}_{x \sim X} \log (1- D(G(x)))
\label{eq:adv}.
\end{equation}

The discriminator then minimises the negative log-likelihood for a standard binary classification task 
and this is equivalent to minimising the Jensen-Shannon (JS) divergence between the model output distribution and real source domain distribution $Y$.

Our scheme, thus far, has equipped the generator model with the ability to treat different image regions non-uniformly by decomposing the image into foreground and background content. In order to further guide the translation task to attend to image regions containing semantically meaningful content, we introduce an additional loss on the attention generator $\Lb_{G_A}$, which exploits the relationship between local and global image patches. The full optimisation objective can then be expressed as follows:

\begin{equation}
\Lb_{G} = \Lb_{adv} +  \Lb^{\textsc{NCE}}  + \Lb_{G_A}
\label{eq:total1}.
\end{equation}

\subsection{Local-global contrastive learning}
\label{sec:method:localglobal}
The contrastive loss of Eq.~\eqref{eq:nce} maximizes mutual information between corresponding input and output patches, ensuring structural consistency between the input and the translated image. We further encourage our model to consistently encode local information, 
of benefit to object detection tasks, and global information representing content that spans beyond objects to 
uncountable amorphous regions of similar texture. The goal is to learn representations 
that result in globally consistent translations while also remaining sensitive to accurate 
representation of local details and structure. 
To this end, we design an objective that $(i)$ encourages local representations of an image to be closer to the global representation of the same image, $(ii)$ encourages local representations of an image to be close to one another, $(iii)$ 
pulls the global representations of an image's distinct augmentations close to one another. This provides the model with an ability to discriminate local representations that describe different content while encouraging patch representations of a common scene to cluster in the latent space. 
Similar local, global strategies have previously proved successful in standard unsupervised object detection settings~\cite{xie_ding_wang_zhan_xu_sun_li_luo_2021} and here we alternatively explore the potential benefits when translating content-rich scenes in a cross-domain object detection setup. 

We decompose our input image $x$ into 16 non-overlapping patches: \mbox{$P(x) = \{ x^p \;|\; p \in [1,16]\}$}. We apply two different random augmentations to $x$, yielding transformed images $x_1$ and $x_2$ and correspondingly; two sets of transformed patches $P_1$ and $P_2$. To obtain local and global representations for contrastive learning, we attach two projection heads to our attention generator $G_{A}$, one assigned for local patches: $MLP_{\text{local}}$ and one assigned to the full global image: $MLP_{\text{global}}$. Following a common 
protocol, we additionally introduce momentum copies of $E_B$ and $G_A$, denoted $G_{Am}$, $E_{Bm}$, where their weights are updated using an exponential moving average. Forwarding images $x_1$, $x_2$ and patch sets $P_1$ and $P_2$ through encoders $E_B$, $E_{Bm}$ and then decoders $G_A$ and $G_{Am}$, generates two sets of global and local feature representations $\{ f_{A}x_1,f_{A}x_2, f_{A}x_{1}^p, f_{A}x_{2}^p \} $ and $\{ f_{Am}x_1,f_{Am}x_2, f_{Am}x_{1}^p, f_{Am}x_{2}^p \} $, pertaining to the outputs of $G_{A}$ and $G_{Am}$, respectively. We then compute $\Lb^{NCE}$ between all pairs in these feature sets to optimise model discriminative power, with negative pairs drawn from a memory bank.

Finally, we define multi-scale supervision to improve the model's ability to identify salient regions. We introduce additional local and global MLP layers at the output of each layer in $G_A$ and compute the infoNCE loss for each new set of features. As a result, our unsupervised loss for $G_A$ can be expressed as follows:

\begin{equation}
\Lb_{G_A}=\sum_{i=1}^{L} w_i \Lb^{NCE}_{g \leftrightarrow g} + \sum_{i=1}^{L} w_i \Lb^{NCE}_{g \leftrightarrow l} + 
\sum_{i=1}^{L} w_i \Lb^{NCE}_{l \leftrightarrow l},
\label{eq:detco}
\end{equation}
where $L$ is the number of layers in $G_A$, and $w_i$ is a weight parameter controlling the importance of each layer contribution. 
The first term in our objective defined in Eq.~\eqref{eq:detco}, denoted $g \leftrightarrow g$, computes the loss between \emph{global} representations, while objective terms $g \leftrightarrow l$ and $l \leftrightarrow l$ indicate that the infoNCE loss is considering \emph{local to global} and \emph{local to local} representations, respectively. 
We clarify the number of hyperparameters introduced in Eq.~\eqref{eq:detco} as follows: the three component loss terms, $\Lb^{NCE}_{g \leftrightarrow g}, \Lb^{NCE}_{g \leftrightarrow l}$ and $\Lb^{NCE}_{l \leftrightarrow l}$, are computed using multi-stage features from individual network layers. We use $L{=4}$, leading to a total of four weights $w_i$, for each of the three loss terms. 
In practice, the weighting for each stage $i$ is common across the three losses, resulting in only four additional distinct hyperparameters in total. We follow convention~\cite{xie_ding_wang_zhan_xu_sun_li_luo_2021} and apply smaller weights to shallow layers and larger weights to deeper layers.

By attaching the aforementioned loss to the features of 
the $G_A$ module, we conjecture that we are able to encourage the attention generator to develop an enhanced 
sensitivity to semantic content and attend to translation regions of importance for the object detection task. 
For completeness, we additionally consider a scenario where annotation labels are available and replace the local-global contrastive loss of Eq.~\eqref{eq:detco}   
with a supervised object saliency loss. We add a
simple auxiliary object saliency task that uses a binary object mask $k(x)$, explicitly separating all foreground objects from the
background, as a target ground truth. We introduce a new convolutional layer $\texttt{Conv}^S$ in $G_A$ that receives the predicted attention maps before softmax and outputs a binary object mask prediction $m(x)$. The resulting {supervised} loss can be defined as:
\begin{equation}
\begin{aligned}[b]
& \Lb_{G_{A_{sup}}} = -\mathbb{E}_{x \sim X}\bigr[ k(x) \log m(x) + (1-k(x)) \log (1-m(x)) \bigr].
\end{aligned}
\label{eq:bce}
\end{equation}

\vspace{-3em} 
\begin{center}
\begin{table}[ht]
\centering
\begin{adjustbox}{scale=0.60}
\begin{tabular}{l  | l l l | c c c c c c c c | c } 
\toprule
Method  & D.A. & I2I  &  Backbone &  person & rider & car & truck & bus & train & motor & bike & mAP $\uparrow$   \\

FGRR~\cite{DBLP:journals/pami/ChenLZHHDY23}     \textsc{\textit{{ }     (TPAMI'23) }}          & {\hspace{0.1cm} \checkmark}    &                            &Vgg-16& 34.4 &47.6 & 51.3 &30.0& 46.8 & 42.3 & 35.1 & 38.9 & 40.8 \\

DAF+NLTE~\cite{liu_li_yang_li_yuan_2022}  \textsc{\textit{{ }     (CVPR '22) }} & {\hspace{0.1cm} \checkmark}            &                            &Res-50& 37.0 & 46.9 & 54.8 & 32.1 & 49.9 & 43.5 & 29.9 & 39.6 & 41.8  \\
TIA~\cite{zhao_wang_2022}     \textsc{\textit{{ }     (CVPR '22) }}    & {\hspace{0.1cm} \checkmark}    &          {\hspace{0.1cm} \checkmark}                    &Res-50& 34.8 & 46.3& 49.7 & 31.1 & 52.1  & 48.6 & 37.7 & 38.1 & 42.3  \\
SCAN~\cite{DBLP:conf/aaai/LiLYY22}     \textsc{\textit{{ }     (AAAI '22) }}          & {\hspace{0.1cm} \checkmark}    &                            &Vgg-16& 41.7 & 43.9 & 57.3 & 28.7& 48.6& 48.7& 31.0 &37.3 &42.1\\
SIGMA~\cite{li2022sigma}     \textsc{\textit{{ }     (CVPR '22) }}          & {\hspace{0.1cm} \checkmark}    &                            &Res-50& 46.9 &48.4& {63.7} & 27.1& 50.7& 35.9& 34.7& 41.4& 43.5 \\
SDA~\cite{rezaeianaran2021seeking}  \textsc{\textit{{ }     (CVPR '21) }}   & {\hspace{0.1cm} \checkmark}    &  &Res-50& 38.8 & 45.9 & 57.2 & 29.9 & 50.2 & 51.9 & 31.9 & 40.9 & 43.3  \\

MGA~\cite{Zhou2022MultiGranularityAD}     \textsc{\textit{{ }     (CVPR '22) }}          & {\hspace{0.1cm} \checkmark}    &                            &Vgg-16& 43.9 & 49.6 & 60.6 & 29.6 & 50.7 & 39.0 & 38.3 & 42.8  & 44.3 \\

DA-DETR~\cite{Zhang2021DADETRDA}     \textsc{\textit{{ }     (CVPR '23) }}          & {\hspace{0.1cm} \checkmark}    &                            &Res-50& 49.9 & 50.0 & 63.1 & 24.0 & 45.8 & 37.5 & 31.6 & 46.3 & 43.5 \\

memCLR~\cite{Vibashan2022TowardsOD}     \textsc{\textit{{ }     (WACV'23) }}          & {\hspace{0.1cm} \checkmark}    &                            &Vgg-16& 37.7 & 42.8& 52.4& 24.5& 40.6&31.7 & 29.4 & 42.2  & 37.7 \\

MIC~\cite{Hoyer2022MICMI}     \textsc{\textit{{ }     (CVPR 23) }}          & {\hspace{0.1cm} \checkmark}    &                            &Vgg-16& 

52.4 &47.5 &67.0 &40.6 &50.9 &55.3 &33.7 &33.9 & \textbf{47.6} \\

CDAT~\cite{Cao2023ContrastiveMT}     \textsc{\textit{{ }     (CVPR 23) }}          & {\hspace{0.1cm} \checkmark}    &                            &Vgg-16& 
42.3 &51.7 &64.0 &26.0 &42.7 &37.1 &42.5 &44.0& 43.8 \\

Ours - supervised  ($\Lb_{G_{A_{sup}}}$)        && {\hspace{0.1cm} \checkmark} & Res-50 & 44.4  &49.5& 61.4 & 32.6 & 50.8 & 52.2 & 38.3 & 44.0 & \underline{46.7}  \\


\midrule
\midrule

CUT$^{*\dag}$~\cite{park2020contrastive} \textsc{\textit{{ }     (ECCV '20) }} & &  {\hspace{0.1cm} \checkmark}     &      Res-50& 39.6 & 45.3 & 59.4 & 27.9 &47.4 & 45.4 &35.3 &39.2& 42.4                 \\
FeSeSim$^{*\dag}$~\cite{zheng_cham_cai_2021}  \textsc{\textit{{ }     (CVPR '21) }}       &&  {\hspace{0.1cm} \checkmark}  &   Res-50& 40.9 &47.2 &58.4 &28.4 &48.6 &49.8 &34.3 &42.7 &43.8            \\
Qs-Att.$^{*\dag}$~\cite{hu_zhou_huang_shi_sun_li_2022}  \textsc{\textit{{ }     (CVPR '22) }}       &&      {\hspace{0.1cm} \checkmark}  & Res-50&                     42.2 & 49.0 &60.3 & 23.5 & 50.5  & 52.0  &36.6  & 41.4 & 44.4   \\

NEGCUT$^{*\dag}$~\cite{negcut}  \textsc{\textit{{ }     (CVPR '21) }}       &&      {\hspace{0.1cm} \checkmark}  & Res-50&                     42.2 & 48.2 &58.8 & 27.9 & 47.8  & 50.2  &34.9  & 43.7 & 44.2   \\

Hneg\_SCR$^{*\dag}$~\cite{hneg}  \textsc{\textit{{ }     (CVPR '22) }}       &&      {\hspace{0.1cm} \checkmark}  & Res-50&                     42.8 & 46.9 & 59.7 & 32.3 & 48.4  & 48.9  & 36.8  & 43.4 & 44.9   \\

Santa$^{*\dag}$~\cite{santa}  \textsc{\textit{{ }     (CVPR '23) }}       &&      {\hspace{0.1cm} \checkmark}  & Res-50&                     42.3 & 47.9 & 59.4 & 34.4 & 49.3  & 49.1  &36.4  & 42.3 & \underline{45.1}   \\

\midrule
\emph{Source}            &&&          Res-50                &35.5& 38.7 & 41.5 & 18.4 & 32.8 & 12.5 & 22.3 & 33.6 & 29.4   \\
\emph{Target Oracle}     &&&          Res-50                &47.5& 51.7 & 66.9 & 39.4 & 56.8 & 49.0 & 43.2 & 47.3 & 50.2   \\
    Ours -  local-global $^{\dag}$ \mbox{ } ($\Lb_{G_{A}}$)  && {\hspace{0.1cm} \checkmark} & Res-50 & 43.2 & 50.1 & 61.7 & 33.3 & 48.6  & 47.8 & 35.2 & 42.6 & \textbf{45.3}  \\

\bottomrule

\end{tabular}
\end{adjustbox}
\caption{The Foggy Cityscapes $\rightarrow$ Cityscapes adaptation scenario. We report object detection (mAP) per class. 
Previous works utilises detector adaptation (D.A), image-to-image translation (I2I) components. Locally reproduced methods using publicly available codes are indicated by \textbf{*}. We separate methods that \textbf{do} (upper) and \textbf{do not} (lower) have access to object annotations at training time, with the latter methods denoted $\mathbf{\dag}$. Results are denoted \textbf{best} and \underline{second best} for upper and lower table sections.} 

\label{tab:results:foggy} 
\end{table}
\end{center}

\begin{table}
\centering
\begin{adjustbox}{scale=0.5}
    \begin{tabular}{l | c c |c c c c} 
    \toprule
    Method & D.A. & I2I    & car & person & mAP $\uparrow$  \\
    \hline
    DARL~\cite{Kim2019DiversifyAM} \textsc{\textit{{ }     (CVPR '19) }}  &{\hspace{0.2cm} \checkmark}  &{\hspace{0.05cm} \checkmark}  & 58.7 & 46.4& 52.5 \\
    DAOD~\cite{DBLP:conf/bmvc/RodriguezM19} \textsc{\textit{{ }     (BMVC '19) }}  & {\hspace{0.2cm} \checkmark}  &{\hspace{0.05cm} \checkmark} & 59.1 & 47.3 &62.9  \\
    DUNIT~\cite{bhattacharjee2020dunit} \textsc{\textit{{ }     (CVPR '20) }} & {\hspace{0.2cm} \checkmark} &{\hspace{0.05cm} \checkmark}  & 65.1 & \underline{60.7} &62.9 \\
     MGUIT \cite{9577366} \textsc{\textit{{ }     (CVPR '21) }}  & &{\hspace{0.05cm} \checkmark}  & 68.2 & 58.3 & 63.2\\
     InstaFormer~\cite{kim2022instaformer} \textsc{\textit{{ }     (CVPR '22) }}  & &{\hspace{0.05cm} \checkmark}  & \underline{69.5} & \textbf{61.8} & \textbf{65.6}\\
    DA-DETR~\cite{Zhang2021DADETRDA} \textsc{\textit{{ }     (CVPR '22) }}  & &{\hspace{0.05cm} \checkmark}  &  48.9 & - & - \\
     \hline
     Source &&& 63.4 & 55.0 & 59.2 \\
     Target Oracle  &&& 77.4 & 66.3 & 71.8  \\
    Ours - supervised \hspace{3mm} \mbox{ } ($\Lb_{G_{A_{sup}}}$)  &   &{\hspace{0.05cm} \checkmark} & \textbf{71.0} & 59.5 & \underline{65.2} \\ 
     Ours - local-global $^\dag$ \mbox{ } ($\Lb_{G_{A}}$)   & &{\hspace{0.05cm} \checkmark} & 67.5 & \underline{60.7} & 64.1 \\ 
    
    \bottomrule
     
    \end{tabular}
    \end{adjustbox}
\hspace{0.2cm}
\centering
\begin{adjustbox}{scale=0.5}
\begin{tabular}{l | c c c | c} 
    \toprule
    Method & D.A. & I2I  & Backbone & $AP_{car} \uparrow$  \\
    \hline
      HTCN~\cite{chen2020harmonizing} \textsc{\textit{{ }     (CVPR '20) }} & {\hspace{0.7cm} \checkmark}& {  \checkmark}&Vgg-16& 42.5
     \\
      UMT~\cite{deng_li_chen_duan_2021}   \textsc{\textit{{ }     (CVPR '21) }} & {\hspace{0.7cm} \checkmark} & {  \checkmark}   &Vgg-16&  43.1\\
       FGRR~\cite{DBLP:journals/pami/ChenLZHHDY23} \textsc{\textit{{ }     (PAMI '23) }} & {\hspace{0.7cm} \checkmark}&&Vgg-16& 44.5
     \\ 
     DSS~\cite{wang2021domain} \textsc{\textit{{ }     (CVPR '21) }}  &  {\hspace{0.7cm} \checkmark}  & &Res-50 & 44.5 \\
    SWDA~\cite{saito_ushiku_harada_saenko_2019} \textsc{\textit{{ }     (CVPR '19) }}  & {\hspace{0.7cm} \checkmark}  & &Res-50  & 44.6\\
    SCDA~\cite{zhu_pang_yang_shi_lin_2019} \textsc{\textit{{ }     (CVPR '19) }}   & {\hspace{0.7cm} \checkmark}  & &Res-50&   45.1 \\
    AFAN~\cite{wang_liao_shao_2021} \textsc{\textit{{ }     (TIP '21) }} & {\hspace{0.7cm} \checkmark}  &   {  \checkmark}& Res-50 & 45.5\\
     GPA~\cite{xu2020cross} $(CVPR '20)$  &  {\hspace{0.7cm} \checkmark}   &    & Res-50 & 47.6 \\
     SDA~\cite{rezaeianaran2021seeking} \textsc{\textit{{ }     (CVPR '21) }} &{\hspace{0.7cm} \checkmark}&    & Vgg-16& 49.3  \\
      MGA~\cite{DBLP:journals/pami/ChenLZHHDY23} \textsc{\textit{{ }     (CVPR '22) }} & {\hspace{0.7cm} \checkmark}&&Vgg-16& 49.8
     \\
     KTNet~\cite{9710025} \textsc{\textit{{ }     (ICCV '21) }}&  {\hspace{0.7cm} \checkmark} &&Vgg-16& 50.7 \\
     SSAL~\cite{Munir2021SynergizingBS} \textsc{\textit{{ }     (NeurIPS '21) }} & {\hspace{0.7cm} \checkmark}&&Vgg-16& 51.8
     \\
       SCAN~\cite{DBLP:conf/aaai/LiLYY22} \textsc{\textit{{ }     (AAAI '22) }} & {\hspace{0.7cm} \checkmark}&&Vgg-16& 52.6
     \\
      SIGMA~\cite{li2022sigma} \textsc{\textit{{ }     (CVPR '22) }} & {\hspace{0.7cm} \checkmark}&&Vgg-16& 53.4
     \\
    
    DA-DETR~\cite{Zhang2021DADETRDA} \textsc{\textit{{ }     (CVPR '23) }} & {\hspace{0.7cm} \checkmark}&&Vgg-16& \textbf{54.7}
     \\
    \hline
    \emph{Source} & && Res-50 & 41.7 \\
    Ours - supervised \hspace{3mm} \mbox{ } ($\Lb_{G_{A_{sup}}}$) & &{  \checkmark}& Res-50 &  \underline{53.6} \\ 
    Ours -  local-global $^{\dag}$ \mbox{ } ($\Lb_{G_{A}}$)   & &{  \checkmark}& Res-50 & 52.1 \\ 
        \bottomrule
\end{tabular} 
\end{adjustbox}
\caption{Adaptation results for KITTI $\rightarrow$ Cityscapes (left) and Sim10K $\rightarrow$ Cityscapes (right). }
\label{tab:results:sim_kitt_combined}
\end{table}

\begin{figure}
\centering
\includegraphics[width=2.5cm]{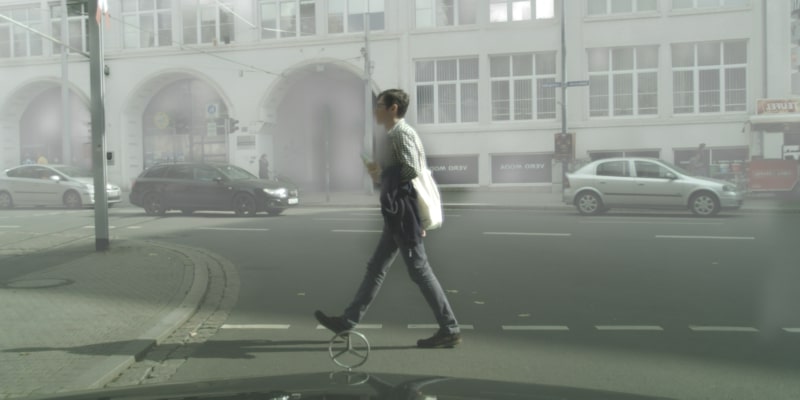}%
\hfill
\includegraphics[width=2.5cm]{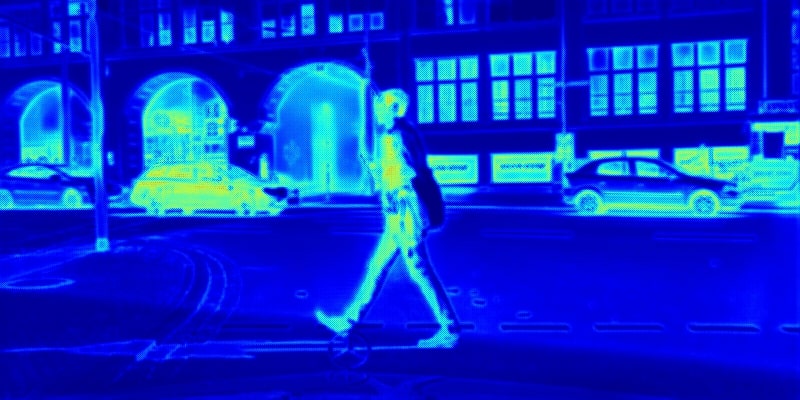}%
\hfill
\includegraphics[width=2.5cm]{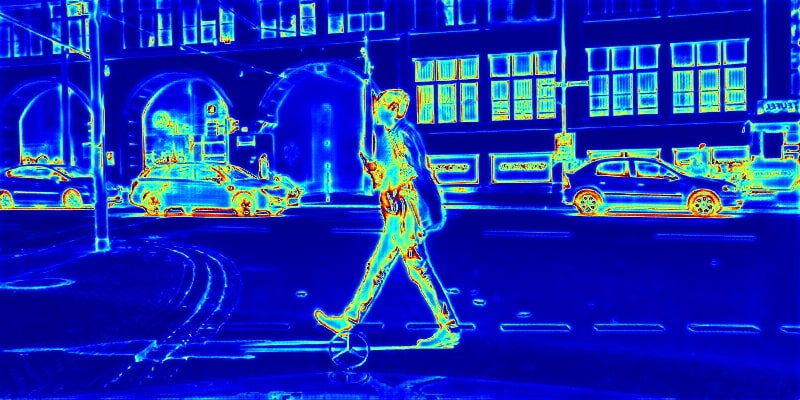}%
\hfill
\includegraphics[width=2.5cm]{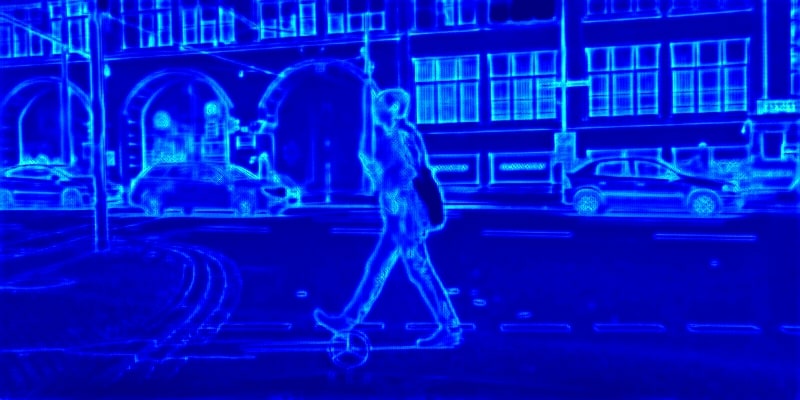}%
\hfill
\includegraphics[width=2.5cm]{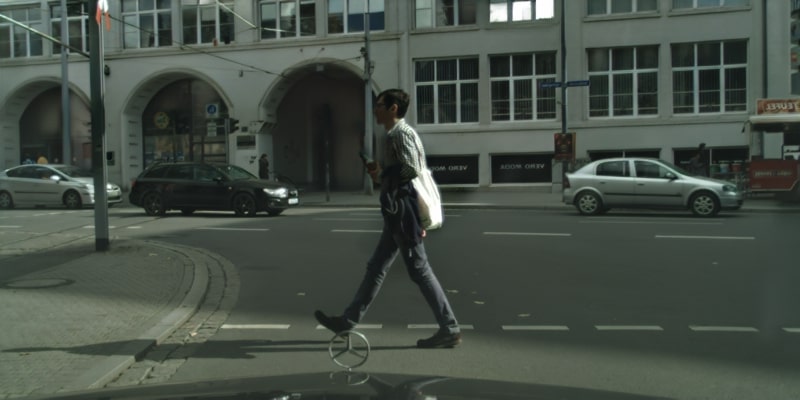}%
\hfill
\includegraphics[width=2.5cm]{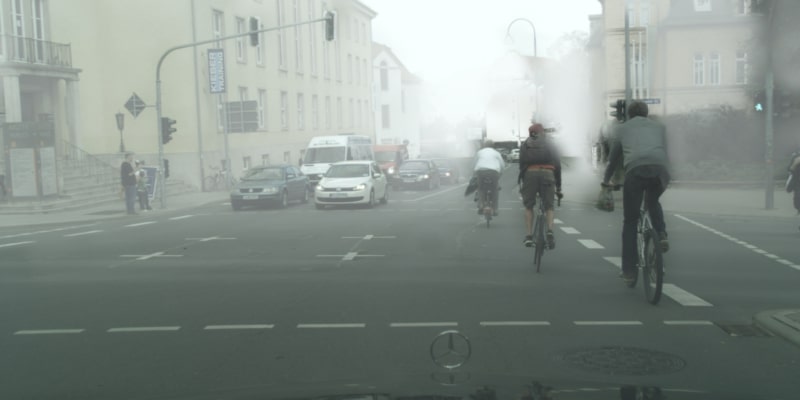}%
\hfill
\includegraphics[width=2.5cm]{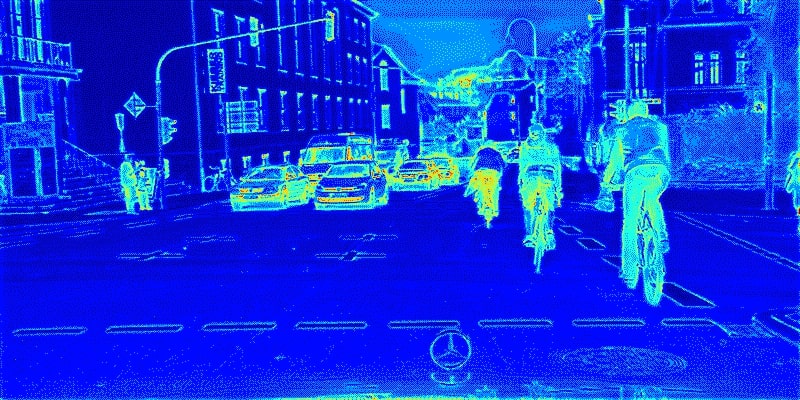}%
\hfill
\includegraphics[width=2.5cm]{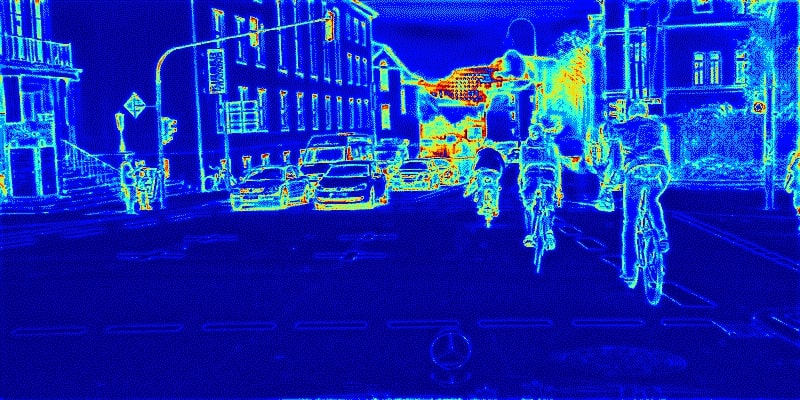}%
\hfill
\includegraphics[width=2.5cm]{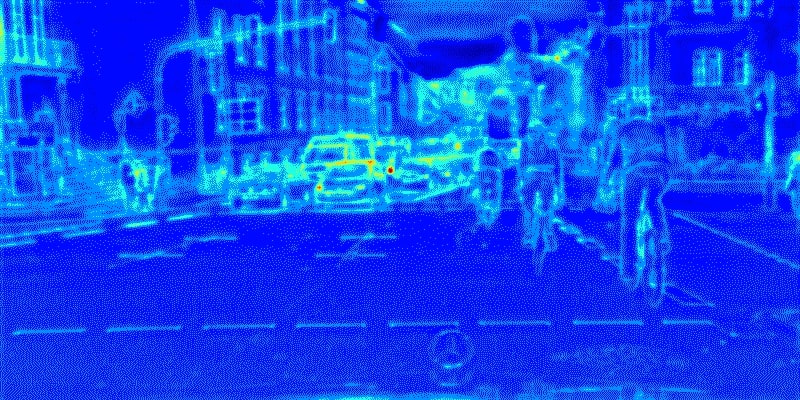}%
\hfill
\includegraphics[width=2.5cm]{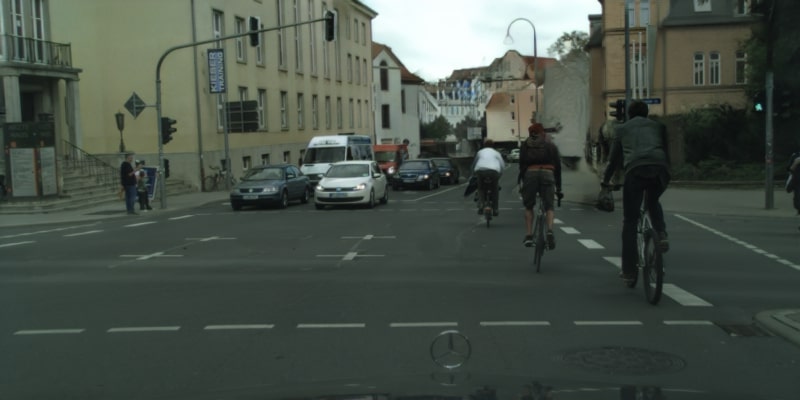}%
\caption{Col $1$: input images. Cols $2-4$: learned foreground attention masks. 
Local-global self-supervision accentuates semantic object content regions and improves translation in areas critical for object detection (e.g.~people, cars). Col $5$: translated output images.}
\label{fig:attention}
\end{figure}

\section{Experiments}
\label{sec:experiments}

\subsection{Datasets}
\label{sec:experiments:data}

\textbf{Foggy Cityscapes $\rightarrow$ Cityscapes.} Cityscapes~\cite{cordts_omran_ramos_rehfeld_enzweiler_benenson_franke_roth_schiele_2016} was collected by capturing images from outdoor urban street scenes, containing $2,975$ images for training and $500$ images for testing with eight annotated object categories, namely: \textit{person, rider, car, truck, bus, train, motorcycle
 and bicycle}. Foggy Cityscapes~\cite{sakaridis_dai_van_gool_2018}, analogously, is a synthetic foggy dataset rendered using Cityscapes, using aligned depth information to simulate synthetic fog on the original clear weather scenes. We firstly evaluate our method under this adversarial weather scenario. 

\textbf{KITTI $\rightarrow$ Cityscapes.}  
KITTI~\cite{geiger_lenz_urtasun_2012} is a widely used autonomous driving dataset containing videos of traffic scenarios recorded with different sensors. The dataset consists of $7,481$ training images and $7,518$ test images, with a total of $80,256$ annotated objects which span eight different categories: \textit{car, van, truck, pedestrian, person sitting, cyclist, tram, misc}. In this challenging real-to-real translation scenario, we study cross-camera adaptation by performing translation from Cityscapes~\cite{cordts_omran_ramos_rehfeld_enzweiler_benenson_franke_roth_schiele_2016} imagery and evaluate on classes \textit{car} and \textit{person}.

\textbf{Sim10$k$ $\rightarrow$ Cityscapes.} 
Sim10$k$~\cite{johnson-roberson_barto_mehta_sridhar_rosaen_vasudevan_2017} is a simulated dataset generated using the GTA-V game engine. It consists of $10,000$ images of synthetic driving scenes and $58,701$ annotated object instances. We perform domain adaptation between the synthesized imagery and the real-world images of the Cityscapes~\cite{cordts_omran_ramos_rehfeld_enzweiler_benenson_franke_roth_schiele_2016} dataset. Here we evaluate our proposed approach by considering detection performance using the \emph{car} class, following a common protocol. 

\subsection{Comparison with State-of-the-Art}
\label{sec:experiments:eval:quant}

\noindent \textbf{Foggy Cityscapes $\rightarrow$ Cityscapes.}
We report object detection results under our initial adaptation scenario in Tab.~\ref{tab:results:foggy}. We present detection performance in terms of per-class average precision (AP) and mean average precision (mAP). With respect to the subset of methods that do not have annotations, our self-supervised local-global configuration achieves state-of-the-art performance of \mbox{$45.3\%$} mAP with the supervised counterpart offering additional further improvement of \mbox{$46.7\%$}. We additionally probe framework efficacy by replacing our specific I2I translation model with three alternative state-of-the-art I2I approaches, whilst keeping the object detector component fixed. Namely we consider I2I approaches CUT~\cite{park2020contrastive}, FeSeSim~\cite{zheng_cham_cai_2021} and Qs-Attn~\cite{hu_zhou_huang_shi_sun_li_2022} (see Tab.~\ref{tab:results:foggy}, lower). Our translation model can show detection gains c.f.~these recent I2I translation models, in each case. We attribute improvements to our local-global framework, capable of accurate object region translation. 
We provide visualisations of the learned attention masks in Fig.~\ref{fig:attention}. These intend to highlight the ability to delineate semantically meaningful regions and attend to relevant discriminative areas in local regions. Specifically, we observe that generated attention masks focus on semantic regions that contain objects, enhancing the discriminative ability of the model. 

\textbf{KITTI $\rightarrow$ Cityscapes.} 
In Tab.~\ref{tab:results:sim_kitt_combined} (left) \mbox{we report results} for this adaptation scenario in comparison with several instance-aware translation methods. 
Following~\cite{bhattacharjee2020dunit}, 
we present the per-class (AP) in addition to the mAP for classes \emph{car} and \emph{person}. Our supervised model achieves \mbox{$65.2\%$} mAP, 
while our self-supervised local-global strategy again exhibits 
competitive performance. 
Our local-global approach stands out as the only method in Tab.~\ref{tab:results:sim_kitt_combined} that does not rely on object annotations during training for this challenging real-to-real scenario.

\textbf{Sim10k $\rightarrow$ Cityscapes.} 
In Tab.~\ref{tab:results:sim_kitt_combined} (right) we report detection results for a further adaptation scenario. In this setting our approach is able to achieve \mbox{$52.1\%$} and \mbox{$53.6\%$} $AP_{50}$ in self-supervised and supervised settings, respectively. 
Compared with recent strategies that specifically employ an image translation module and use an identical detector backbone (namely 
AFAN~\cite{wang_liao_shao_2021}) 
our supervised model achieves gains of 
over $8\%$. 
Our self-supervised variant trained with local-global contrastive learning can also achieve performance competitive with detector adaptation based methods, yet without access to object label information. 

\subsection{Ablative study}
We examine the impact of the proposed components in detail and report results in Tab.~\ref{tab:ablation}. We select the \mbox{Foggy Cityscapes $\rightarrow$ Cityscapes} adaptation scenario and train the proposed model architecture under the following ablations: $(i)$ \textit{without} the $G_A$ network and \textit{without} the proposed attention module; $(ii)$ \textit{with} the $G_A$ network, \textit{with} the proposed attention module and \textit{without} loss $\Lb_{G_A}$; $(iii)$ \textit{with} the $G_A$ network, \textit{with} the proposed attention module and \textit{with} an unsupervised $\Lb_{G_A}$ loss; and finally $(iv)$ \textit{with} the $G_A$ network, \textit{with} the proposed attention module and \textit{with} a supervised $\Lb_{G_{A_{sup}}}$ loss. All models are trained under identical settings which are reported in our supplementary materials. In all cases we use the adversarial loss $\Lb_{adv}$ found in Eq.~\eqref{eq:adv} and the InfoNCE loss found in Eq.~\eqref{eq:nce}. We observe that detection performance is lower in case $(i)$ where all method components under consideration are absent. In case $(ii)$ performance is improved by 0.5--1.7\% mAP@.5 which we attribute to the addition of the proposed attention module, trained without any guidance (i.e. $\Lb_{G_A}=0$). Inclusion of all components results in 2.3--2.6\% mAP@.5 gains for the unsupervised model (case $(iii)$) and 2.3--4\% mAP@.5 gains for the supervised model (case $(iv)$). We provide further ablative comparisons in our supp.~materials.

\begin{table}[ht]

\footnotesize
\begin{adjustbox}{scale=0.67}
\begin{tabular}{c | c c c c c c c c c c c} 
\toprule

Det. backbone & $G_A$   &  $\Lb_{G_A}$ & Supervision  & Attention &  mAP@[.5:.95] & mAP@.5 &  mAP@.75  & mAP@[.5:.95] & mAP@[.5:.95] & mAP@[.5:.95]  \\
 
 &   &   &  &  &   & &   &  \scriptsize{small} &  \scriptsize{medium}  &  \scriptsize{large}  \\
\hline
 & & & - & &    23.0  & 42.7  & 21.8 & 2.2  &20.8 & 47.4 \\

 & {\hspace{0.2cm} \checkmark} &  & - &  {\hspace{0.2cm} \checkmark}  & 23.5 & 44.4 & 20.8  & 2.5 & 22.3 & 46.3  \\

\hspace{0.0cm} Res-50 &{\hspace{0.2cm} \checkmark}  & {\hspace{0.2cm} \checkmark}  &   local--global $\Lb_{G_{A}}$ &  {\hspace{0.2cm} \checkmark}   & 24.1 & 45.3 & 23.2 & 2.6 & 23.3 & 47.1    \\

&{\hspace{0.2cm} \checkmark}  & {\hspace{0.2cm} \checkmark}  &  supervised $\Lb_{G_{A_{sup}}}$ &  {\hspace{0.2cm} \checkmark}   & 24.5 & 46.7  & 22.9 & 2.7 & 23.4 & 47.1  \\

\bottomrule
\end{tabular}

\end{adjustbox}

\caption{Ablation on method components (Foggy Cityscapes $\rightarrow$ Cityscapes).}


\label{tab:ablation} 

\end{table}

\begin{wrapfigure}[10]{r}{0.5\textwidth}
\begin{minipage}{\linewidth}
\centering 
\subfigure[\tiny{\mbox{Baseline without $G_A$}}]{\includegraphics[width=0.3\linewidth]{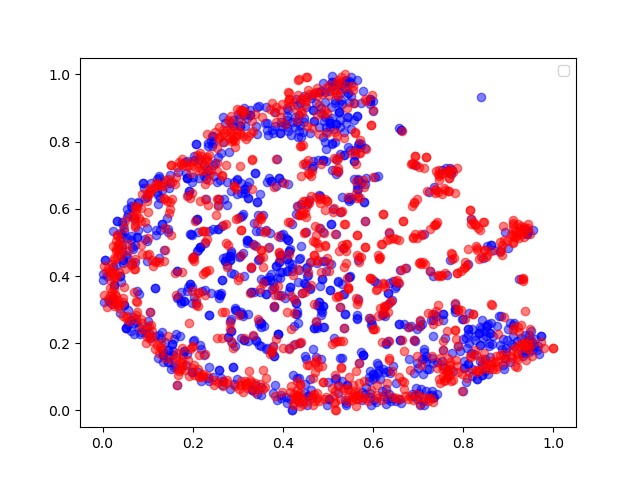}}\label{fig:tsne:a}
\subfigure[\tiny{\mbox{supervised $G_{A_{\scalebox{.75}{\emph{sup}}}}$}}]{\includegraphics[width=0.3\linewidth]{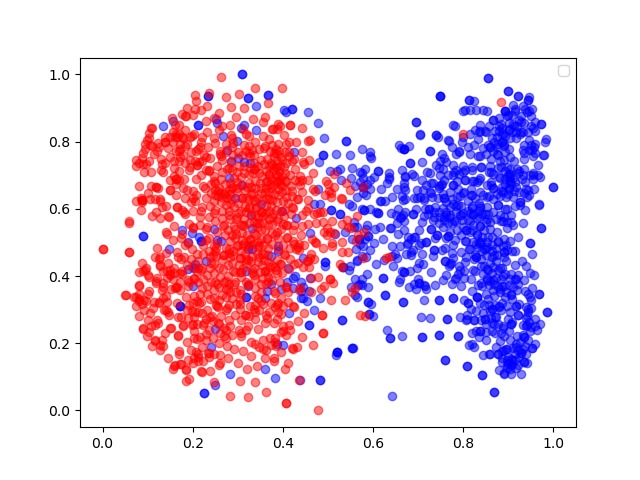}}\label{fig:tsne:b}
\subfigure[\tiny{\mbox{self-supervised $G_A$ 
}} ]{\includegraphics[width=0.3\linewidth]{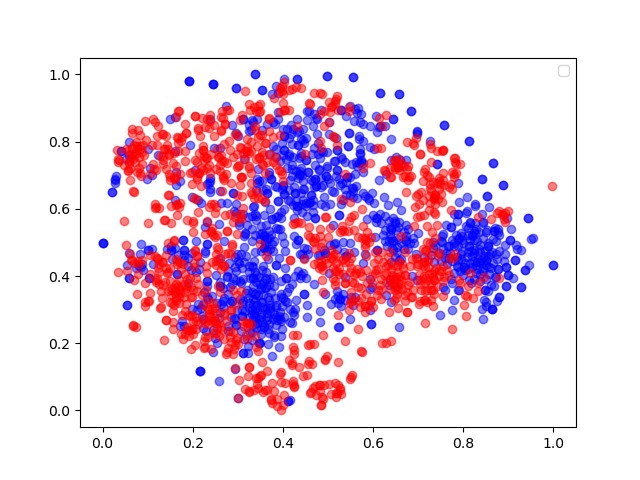}}\label{fig:tsne:c}
\end{minipage}
\caption{t-SNE feature visuliazation; we randomly sample object features corresponding to salient objects (red) and image background regions (blue).}
\label{fig:tsne}
\end{wrapfigure}

We additionally present a feature-level visualization via t-SNE~\cite{tsne} in Fig.~\ref{fig:tsne}, towards evidencing method effectiveness in terms of identifying relevant salient object regions. 
%
We visualise two classes, defined using object bounding box labels, as $\{$object, background$\}$ and randomly sample $1000$ feature points. We observe that in case (a) the learned representations do not afford discriminability between 
these two classes. Adding the supervised object saliency signal in case (b) results in a clearly separable learned feature embedding. Finally, in case (c) we evidence that our self-supervised local-global model, using Eq.~\eqref{eq:detco}, 
can enhance separability c.f.~case (a), which concurs with our empirical 
observations that 
manifest as object, background disentanglement behaviour.

\section{Conclusion}
\label{sec:conc}

We propose a novel approach for cross-domain object detection using unpaired image-to-image translation. 
Our contrastive-learning based attention mechanism endows the model with object awareness and steers feature representations to be discriminative in terms of benefit to downstream detection tasks, post image translation between source and target domains. We explore generation of attention masks in 
fully unsupervised regimes and evidence competitive detection results in comparison with numerous state-of-the-art methods, whilst requiring neither domain-paired image data nor access to object labels.

\bibliographystyle{plain}
\bibliography{refs}

\clearpage

\section{Improving Object Detection via Local-global Contrastive Learning: Supplementary materials}

\label{sec:supplementary}

\setcounter{equation}{0}
\setcounter{figure}{0}
\setcounter{table}{0}
\setcounter{page}{1}
\renewcommand{\theequation}{S\arabic{equation}}
\renewcommand{\thefigure}{S\arabic{figure}}
\renewcommand{\thetable}{S\arabic{table}}

We provide additional materials to supplement our main paper. 
In Sec.~\ref{sec:supplementary:qual} we report additional qualitative results to complement those found in the main paper. 
Sec.~\ref{sec:supplementary:ablation} provides a sensitivity analysis and further extended ablative study on method components for various detector backbone architectures. 
In Sec.~\ref{sec:supplementary:compare} we provide a comparison with contrastive learning based I2I translation methods. 
Finally, Sec.~\ref{sec:supplementary:imp} gives supplementary information on learning hyperparameters and further implementation details. 

\subsection{Additional qualitative results}
\label{sec:supplementary:qual}

\begin{figure}[]
    \centering
    \begin{subfigure}
        \centering
        \includegraphics[width=2.5cm]{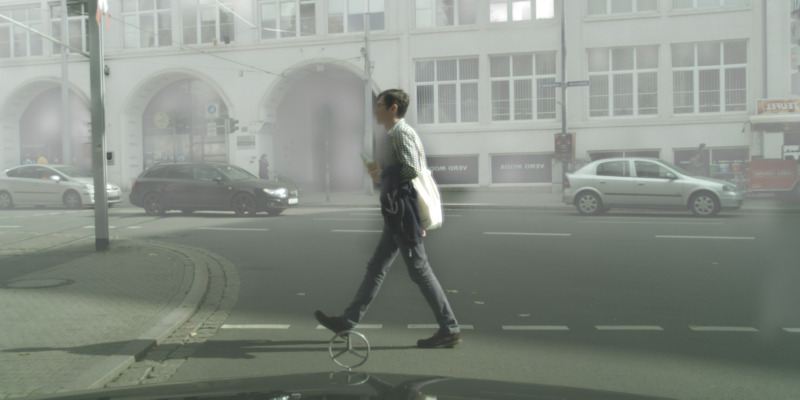}
        \includegraphics[width=2.5cm]{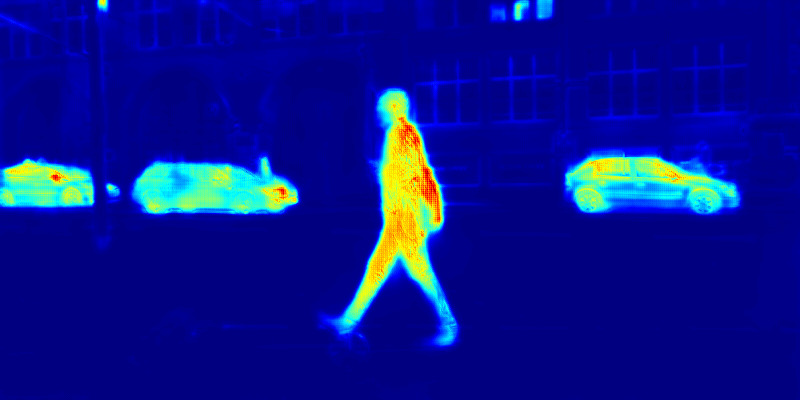}
        \includegraphics[width=2.5cm]{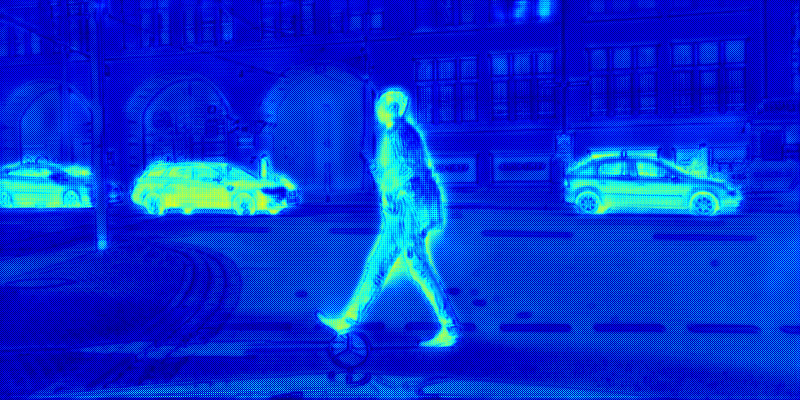}
        \includegraphics[width=2.5cm]{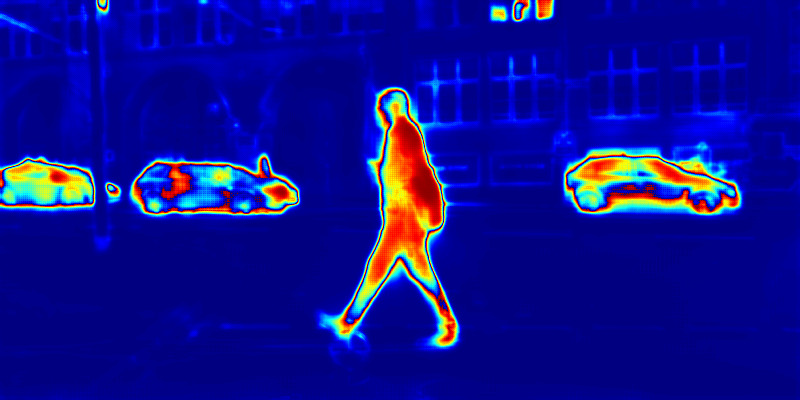}
        \includegraphics[width=2.5cm]{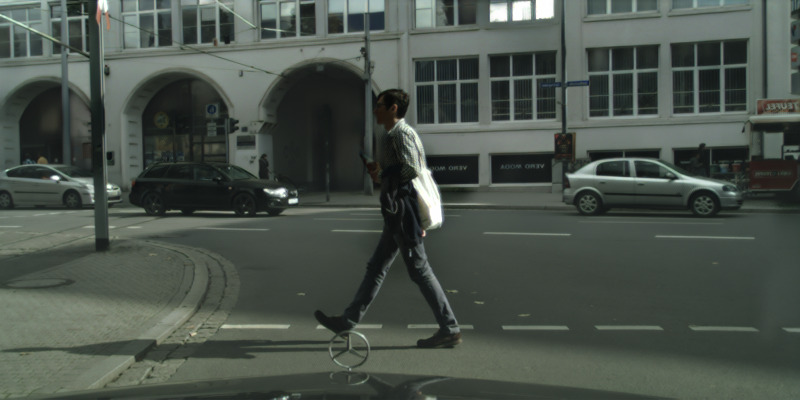}
    \end{subfigure}
    \begin{subfigure}
        \centering
        \includegraphics[width=2.5cm]{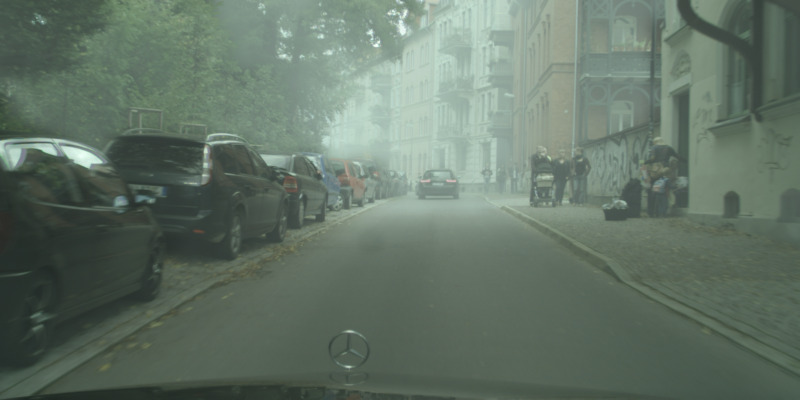}
        \includegraphics[width=2.5cm]{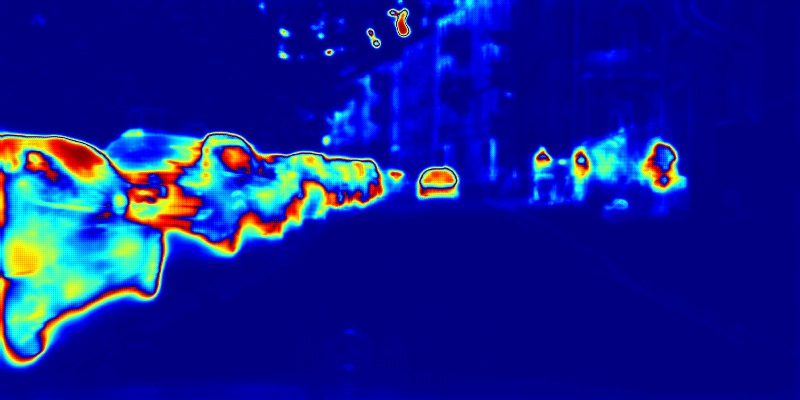}
        \includegraphics[width=2.5cm]{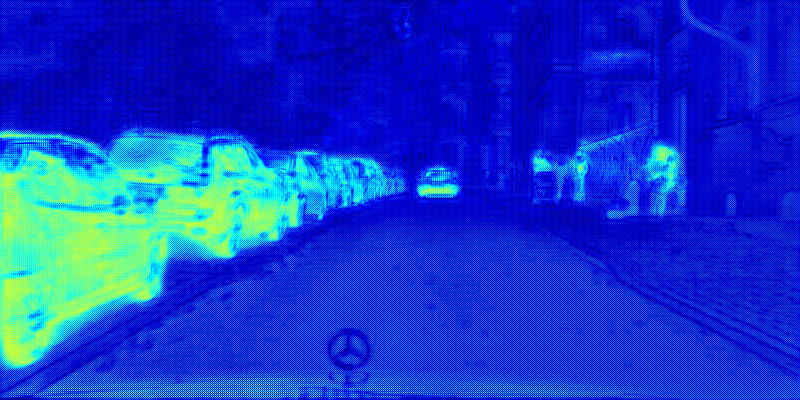}
        \includegraphics[width=2.5cm]{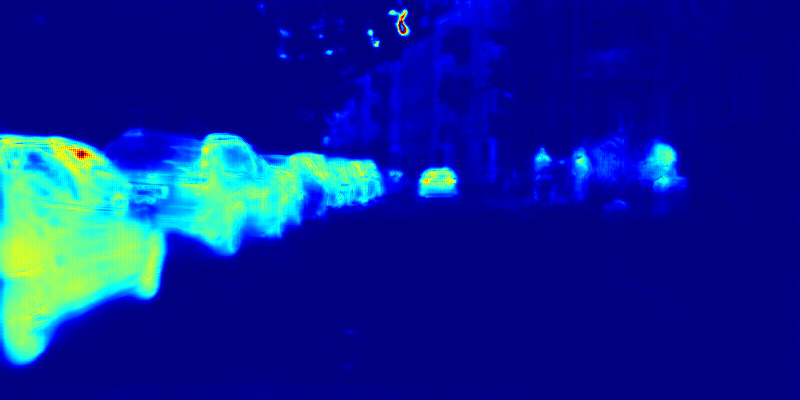}
        \includegraphics[width=2.5cm]{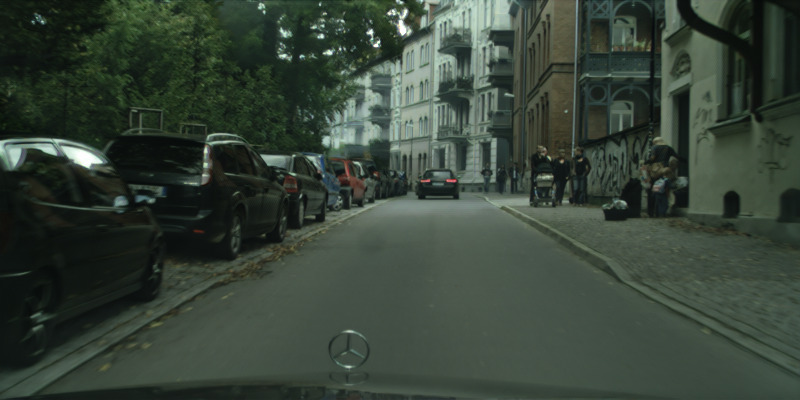}
    \end{subfigure}
    \begin{subfigure}
        \centering
        \includegraphics[width=2.5cm]{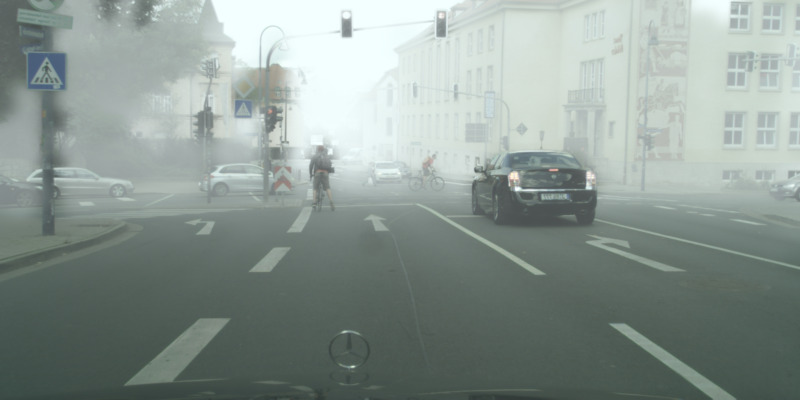}
        \includegraphics[width=2.5cm]{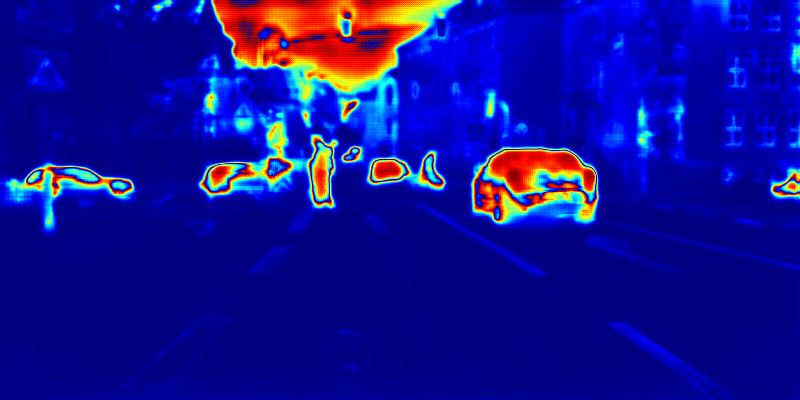}
        \includegraphics[width=2.5cm]{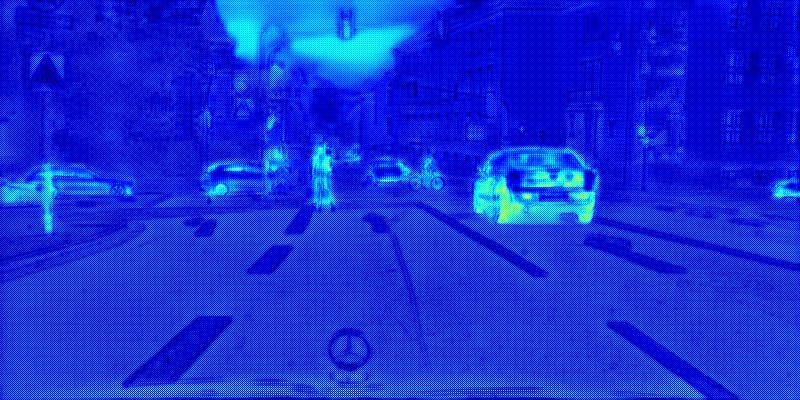}
        \includegraphics[width=2.5cm]{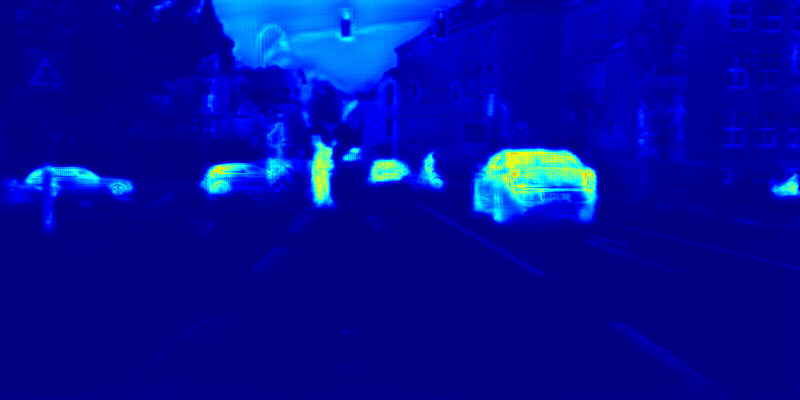}
        \includegraphics[width=2.5cm]{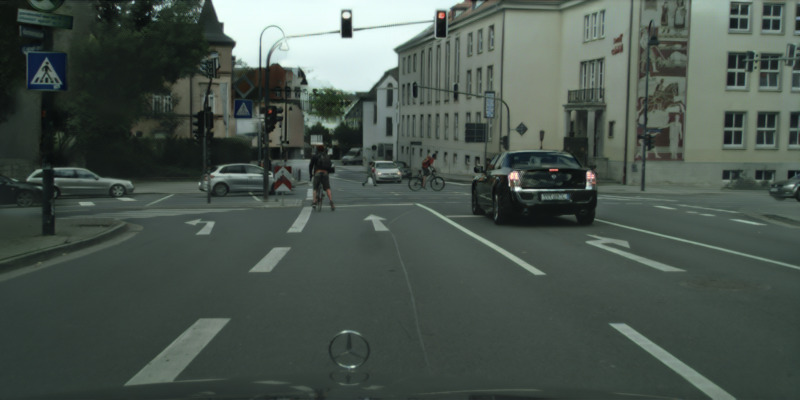}
    \end{subfigure}
    \begin{subfigure}
        \centering
        \includegraphics[width=2.5cm]{images/supplementary/attention_masks/epoch001_real_A_58.jpg}
        \includegraphics[width=2.5cm]{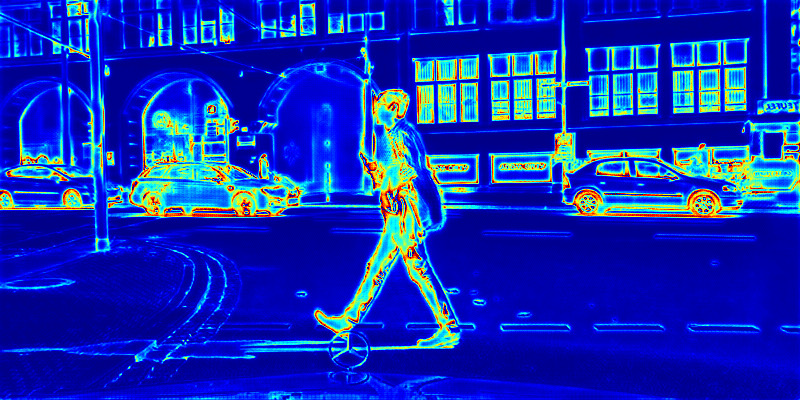}
        \includegraphics[width=2.5cm]{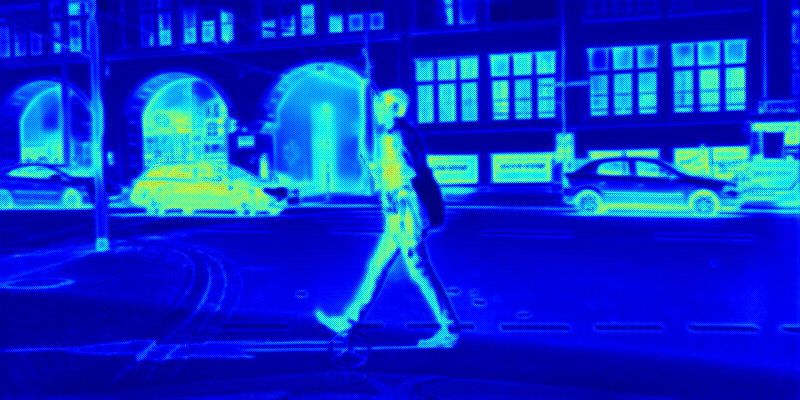}
        \includegraphics[width=2.5cm]{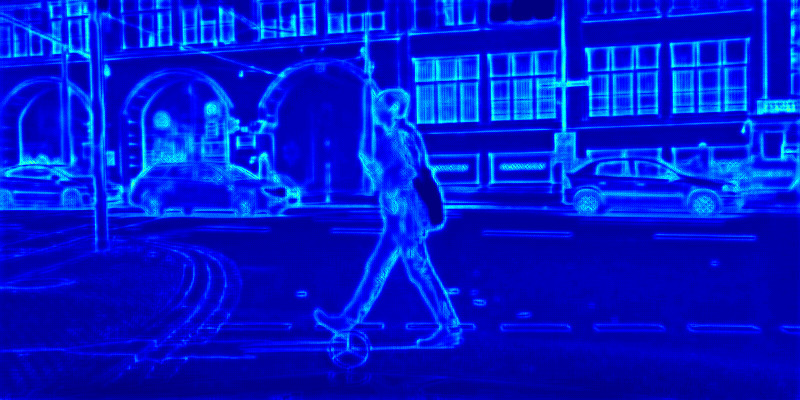}
        \includegraphics[width=2.5cm]{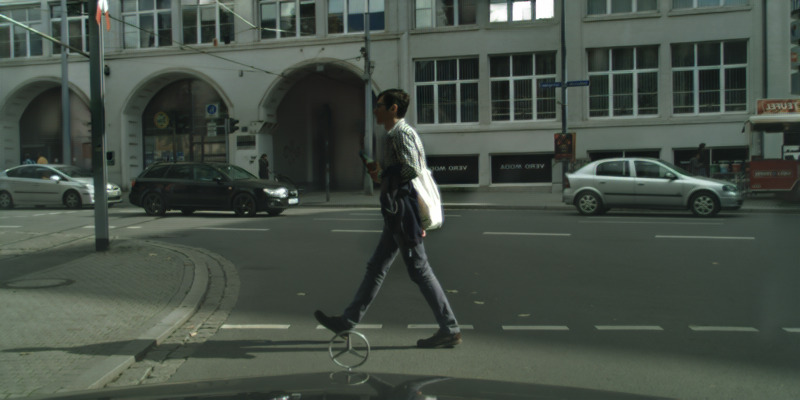}
    \end{subfigure}
    \begin{subfigure}
        \centering
        \centering
        \includegraphics[width=2.5cm]{images/supplementary/attention_masks/epoch001_real_A_66.jpg}
        \includegraphics[width=2.5cm]{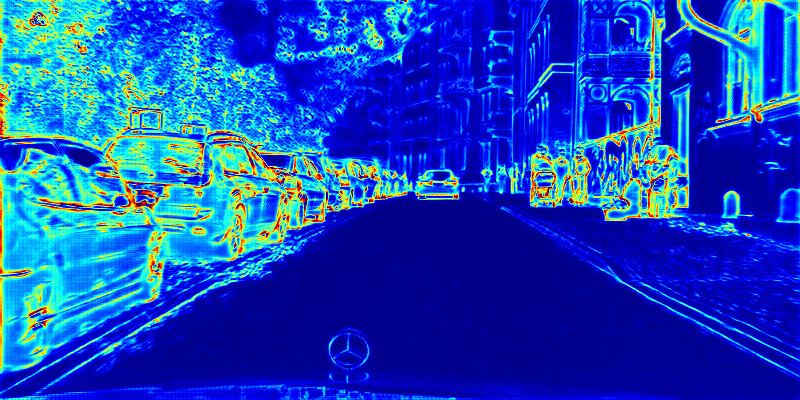}
        \includegraphics[width=2.5cm]{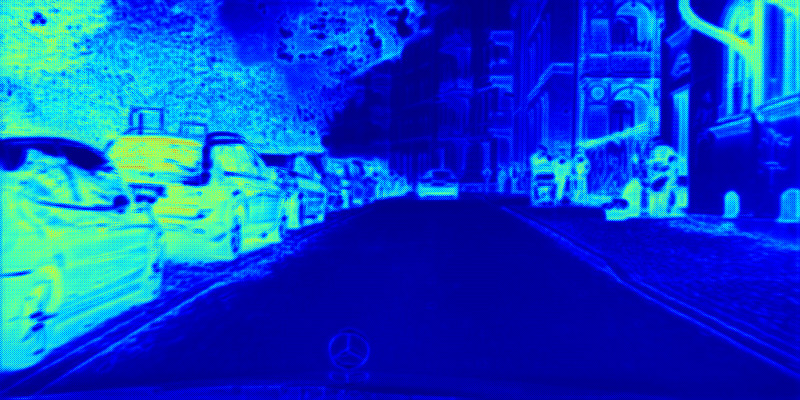}
        \includegraphics[width=2.5cm]{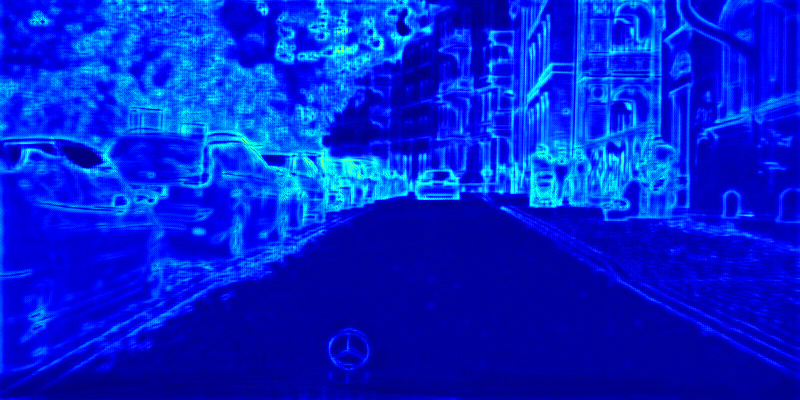}
        \includegraphics[width=2.5cm]{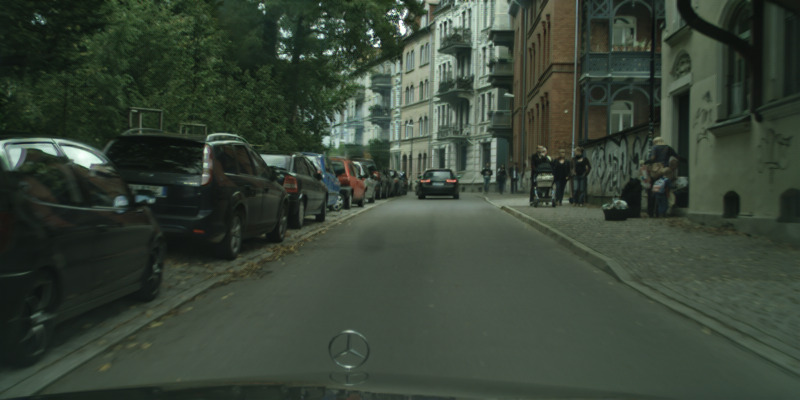}
        \vspace{0.07cm}
    \end{subfigure}

    \begin{subfigure}
        \centering
    \begin{subfigure}
        \centering
        \centering
        \includegraphics[width=2.5cm]{images/supplementary/attention_masks/epoch001_real_A_79.jpg}
        \includegraphics[width=2.5cm]{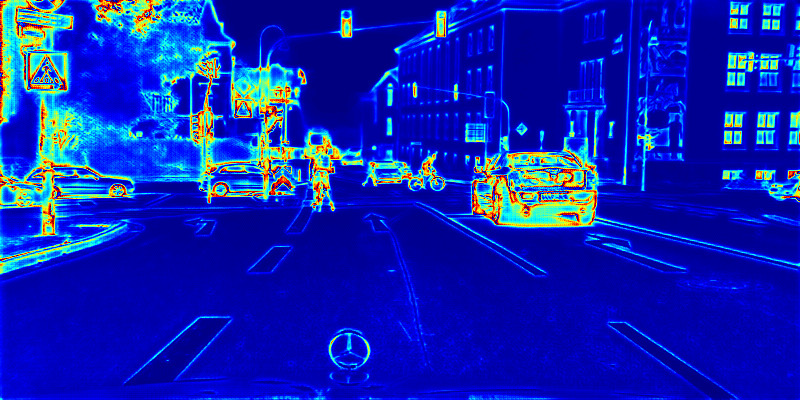}
        \includegraphics[width=2.5cm]{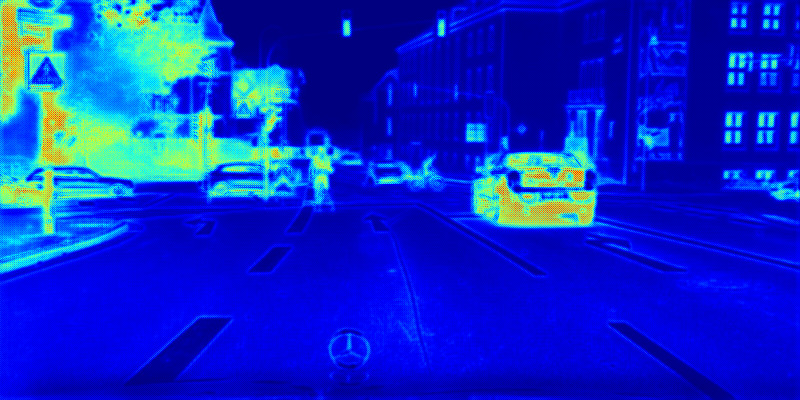}
        \includegraphics[width=2.5cm]{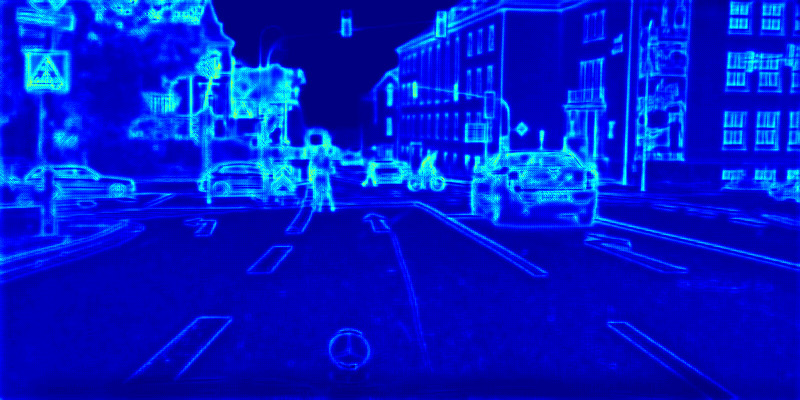}
        \includegraphics[width=2.5cm]{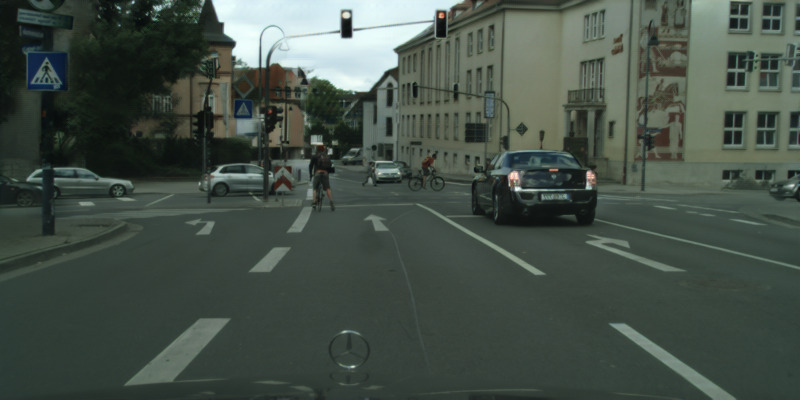}
        \vspace{0.07cm}
    \end{subfigure}
    \end{subfigure}

\caption{Visualization of the learned foreground attention masks of the proposed supervised model (rows 1-4) and self-supervised local-global model (rows 5-8). Column 1 shows the input (foggy weather) image, columns 2--4 visualize attention masks from 3 different channels $A_k$ and column 5 shows the translated (clean weather) result. }
\label{fig:supp:attention}
\end{figure}

\begin{figure}
    \centering
    \begin{subfigure}
        \centering
        \includegraphics[width=4cm,height=2cm]{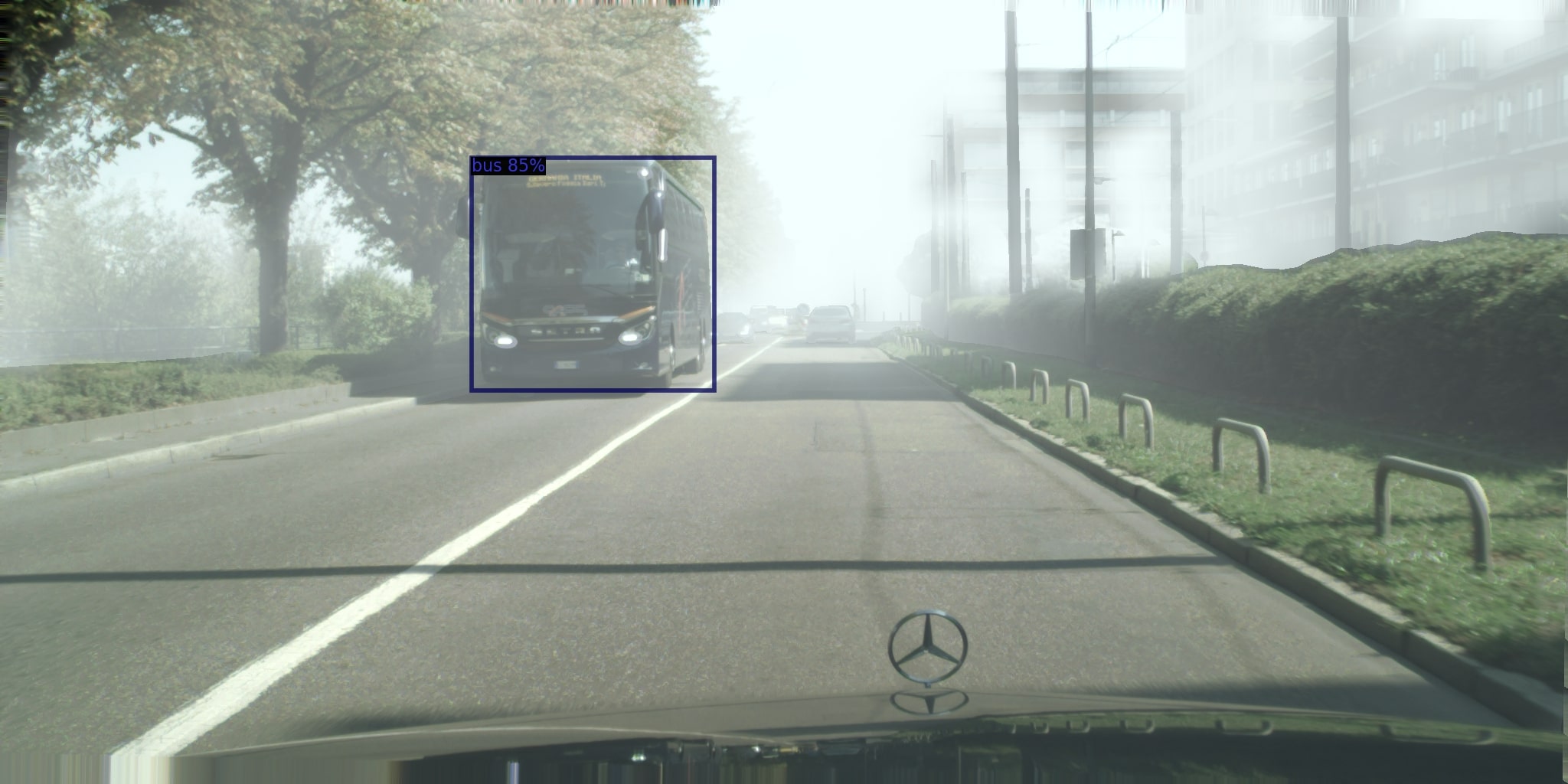}
        \includegraphics[width=4cm,height=2cm]{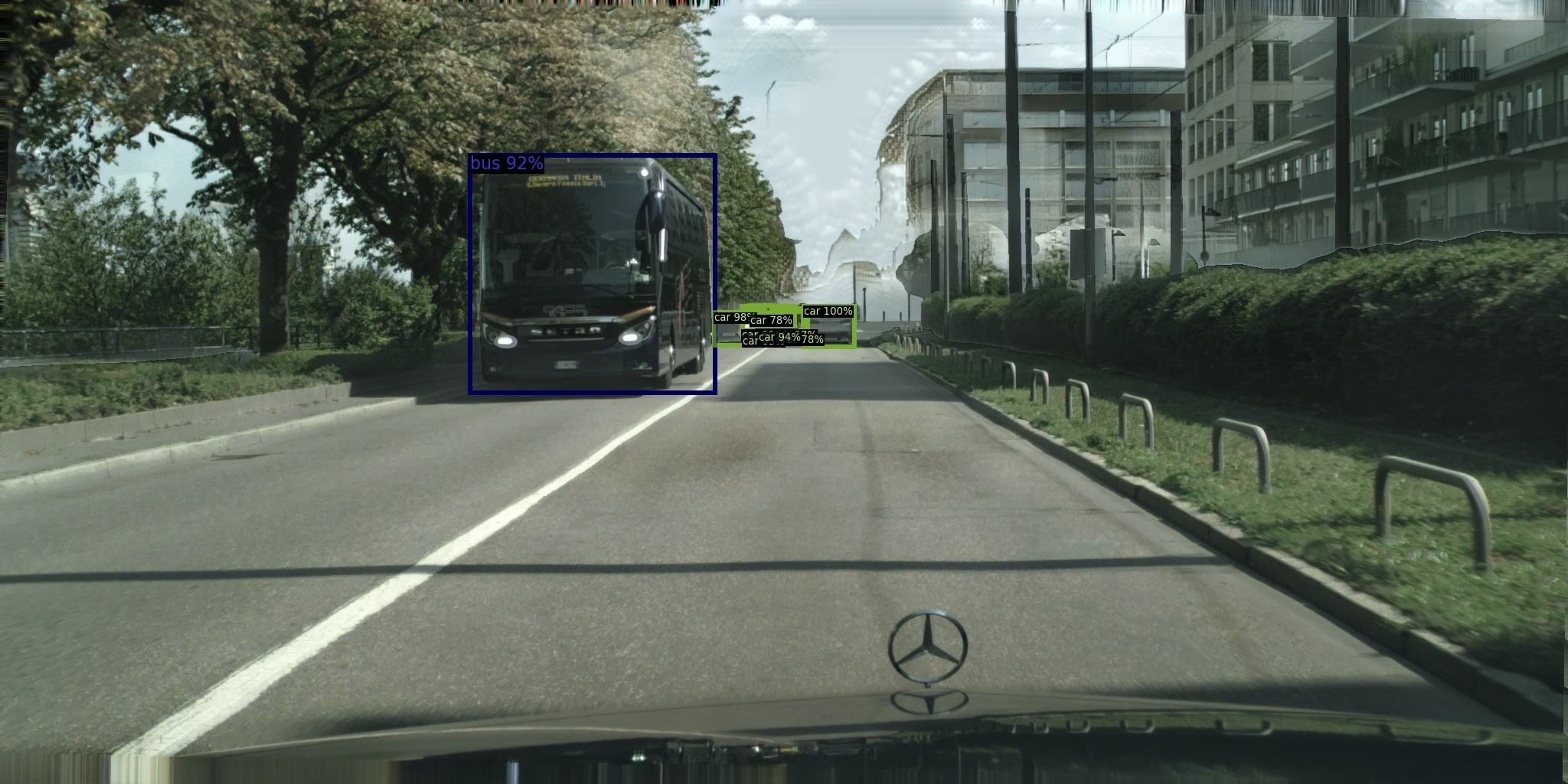}
        \includegraphics[width=4cm,height=2cm]{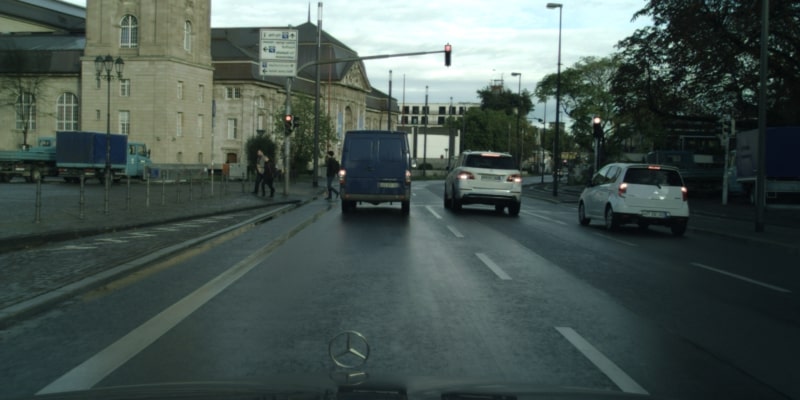}
    \end{subfigure}
    \begin{subfigure}
        \centering
        \includegraphics[width=4cm,height=2cm]{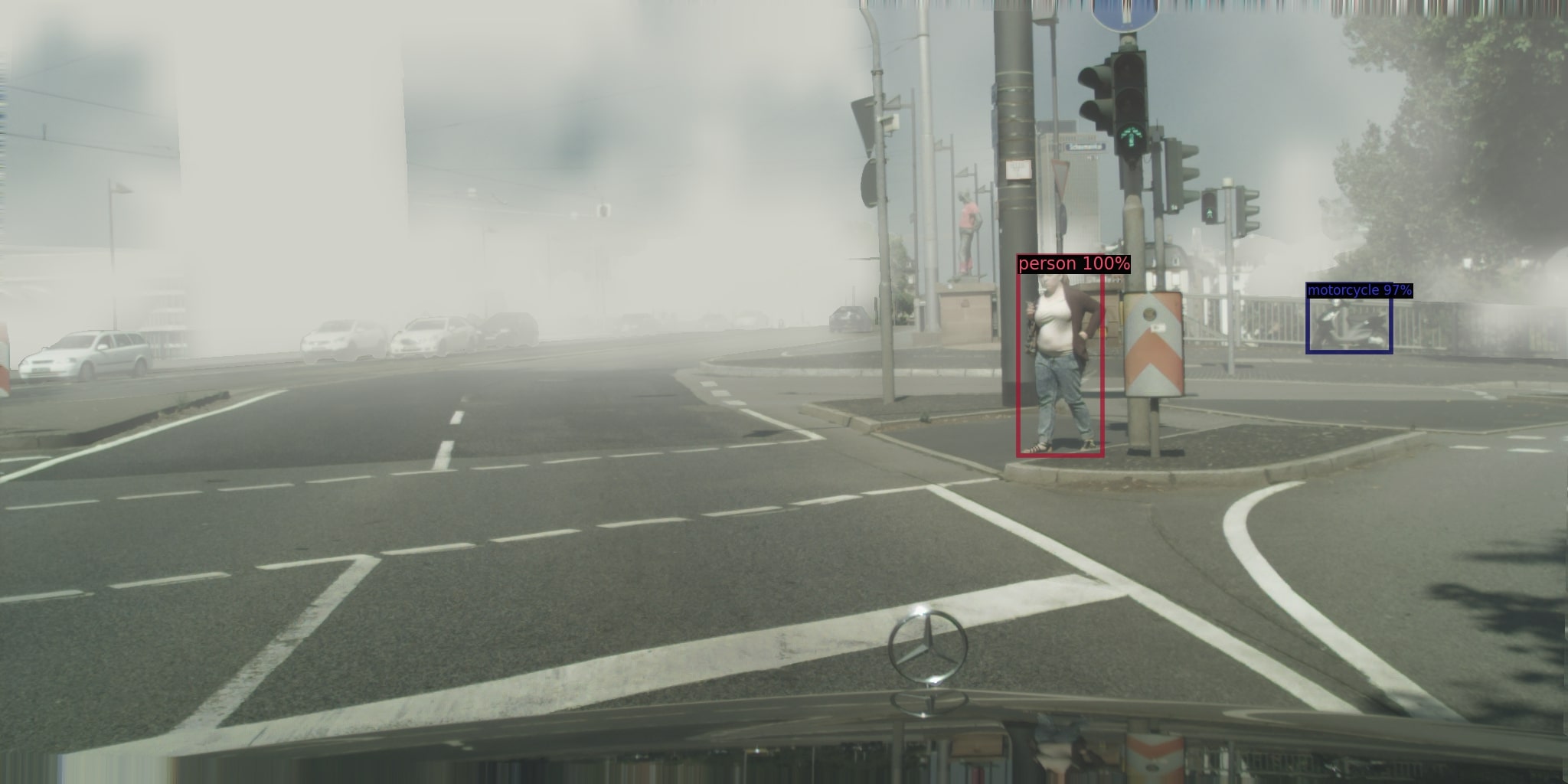}%
        \includegraphics[width=4cm,height=2cm]{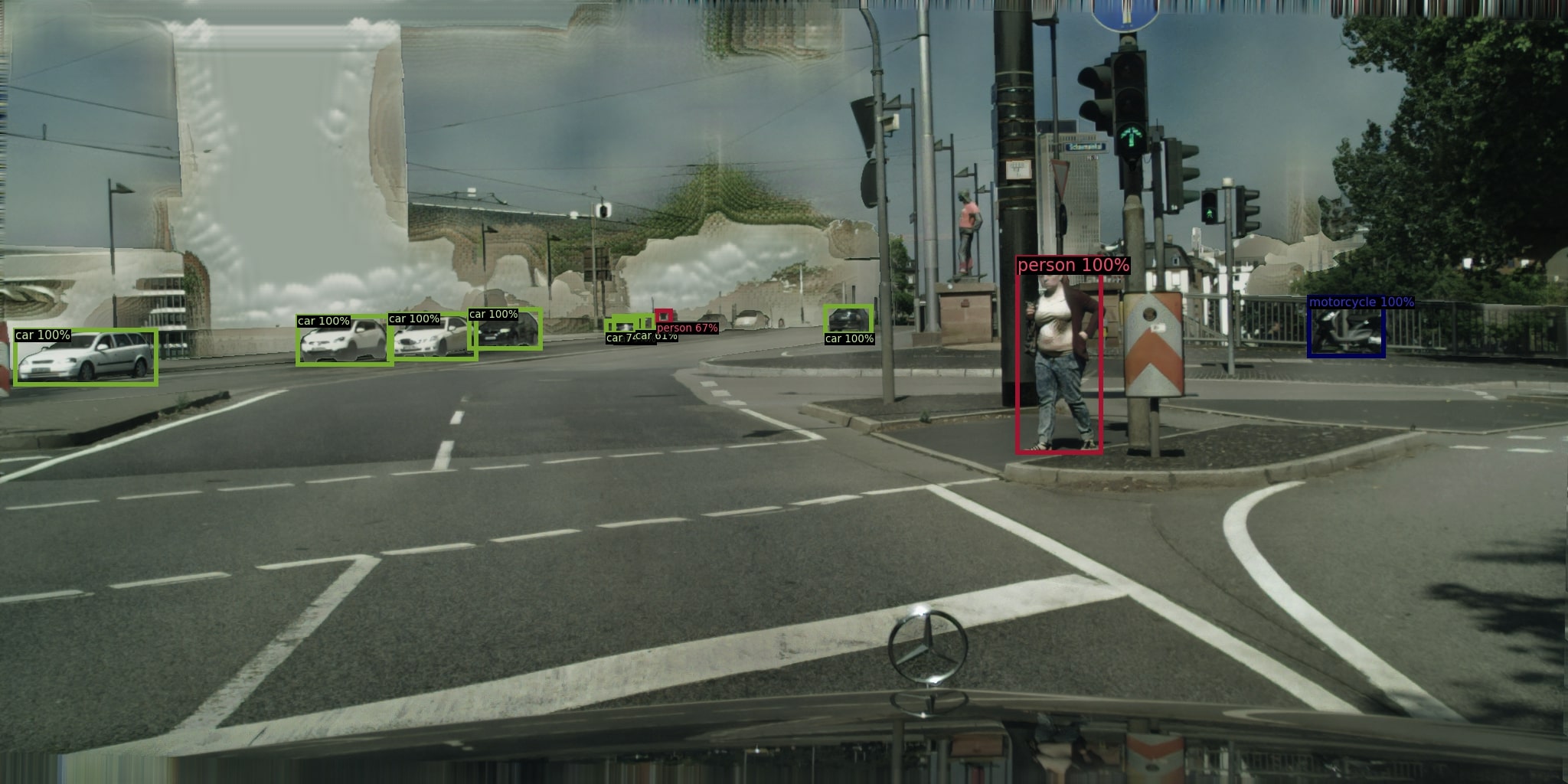}
        \includegraphics[width=4cm,height=2cm]{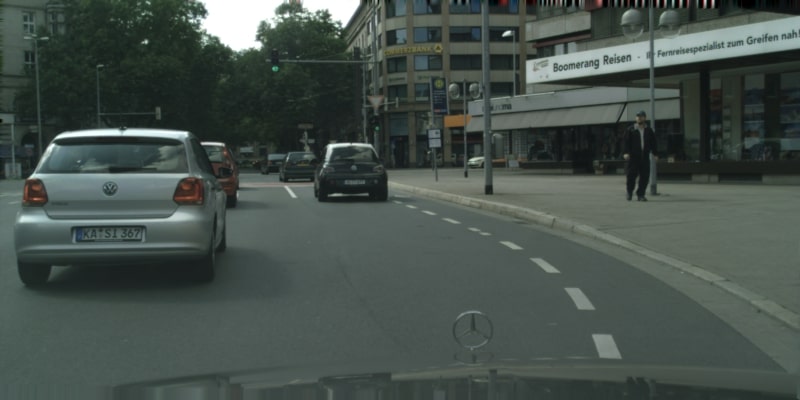}
    \end{subfigure}
    \begin{subfigure}
        \centering
        \includegraphics[width=4cm,height=2cm]{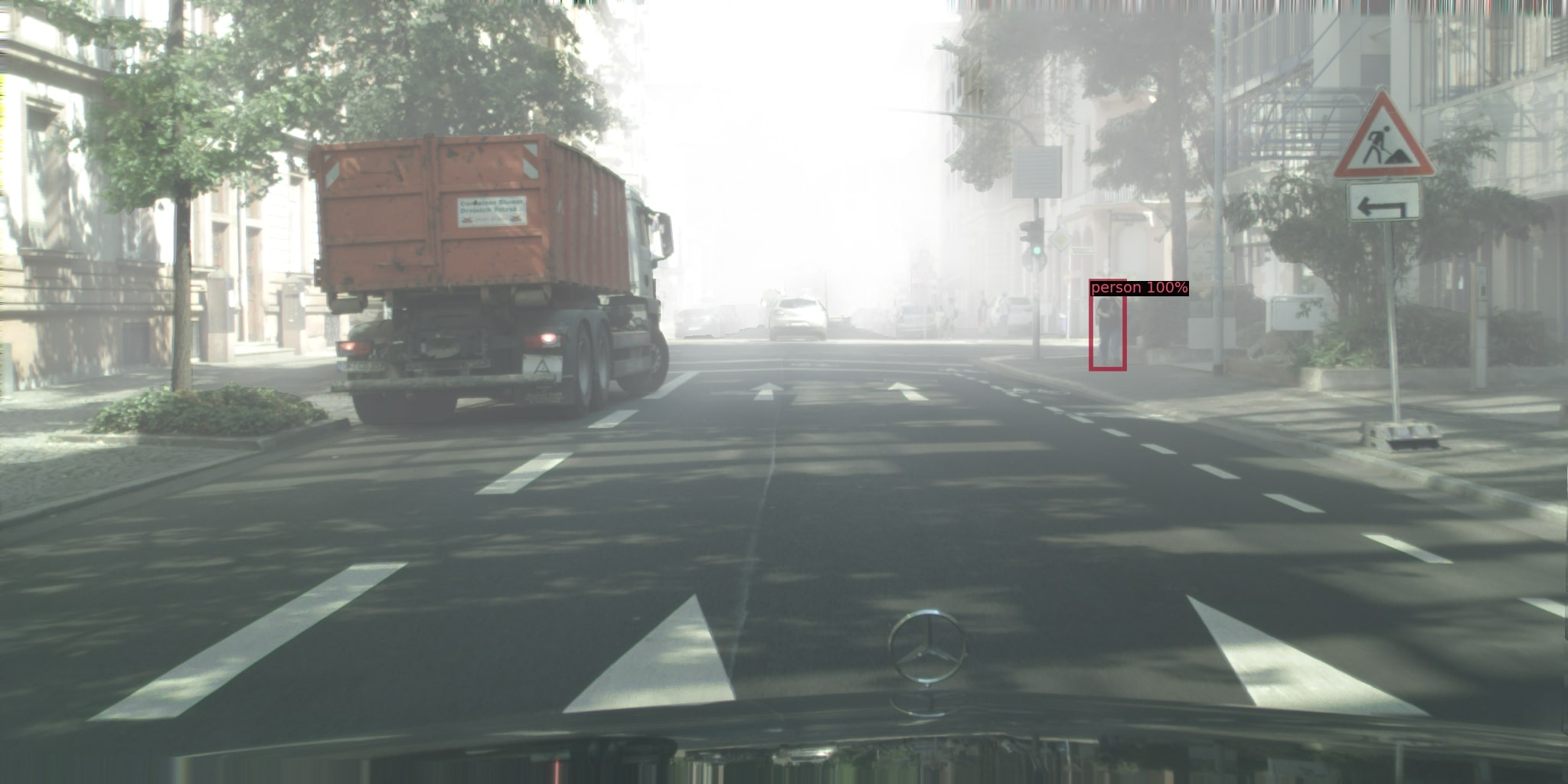}
        \includegraphics[width=4cm,height=2cm]{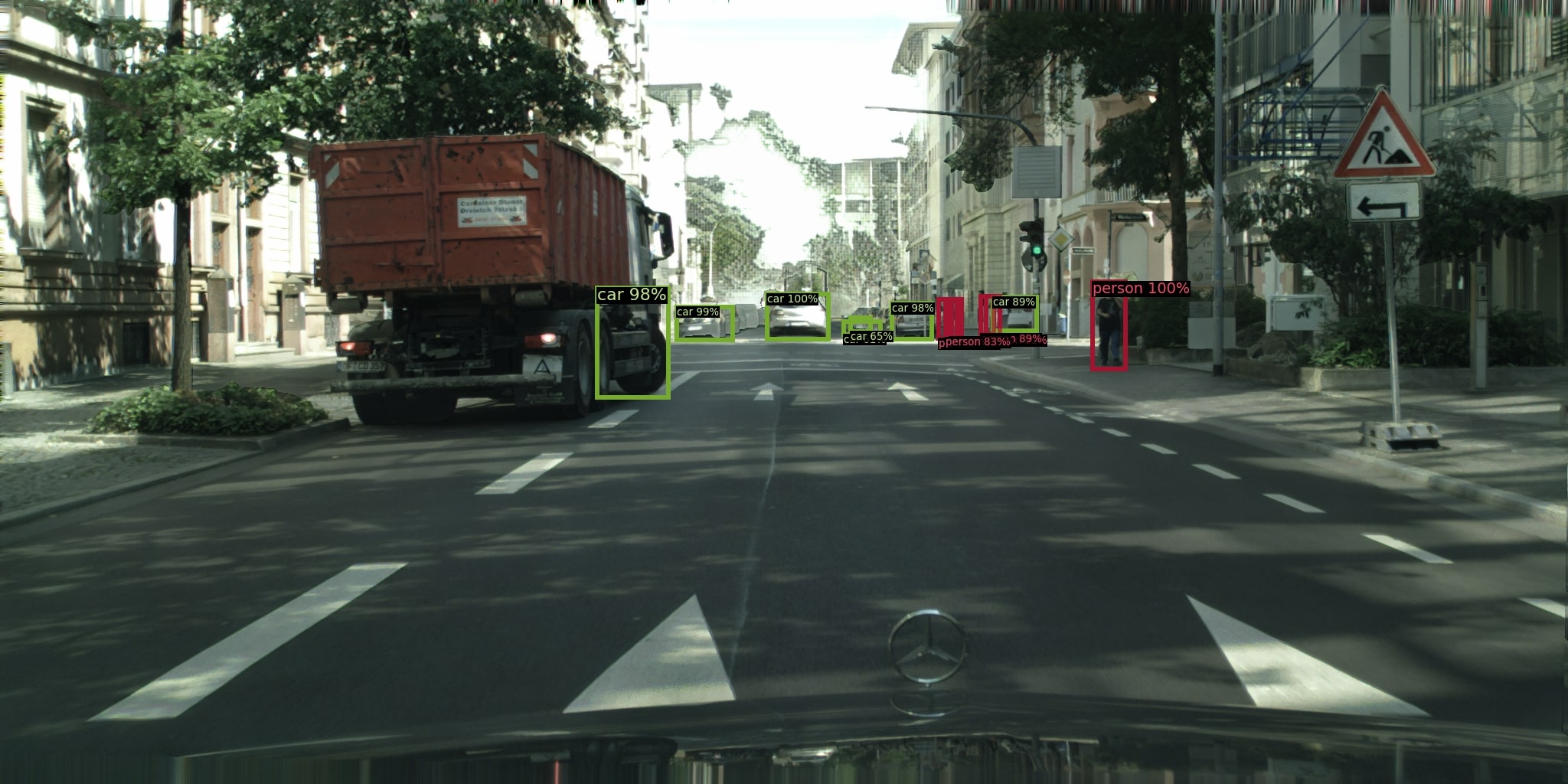}
        \includegraphics[width=4cm,height=2cm]{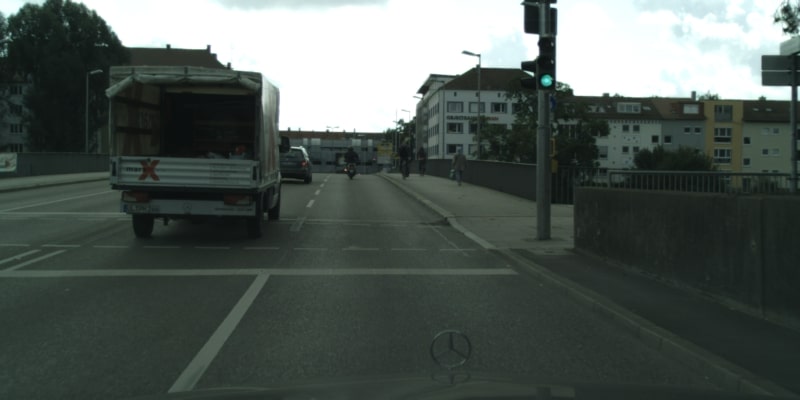}
    \end{subfigure}

\caption{ The Foggy Cityscapes $\rightarrow$ Cityscapes adaptation scenario. In all cases the detector model is trained on the Cityscapes dataset and evaluated on Foggy Cityscapes. We visualise detection inference results on Foggy Cityscapes imagery (column 1), detection inference results on translated imagery (column 2) and an (unpaired) real image from the Cityscapes dataset, to aid effective translation assessment (column 3).  }
\label{fig:supp:foggy_to_cityscapes}
\end{figure}

\begin{figure}
    \centering
    \begin{subfigure}
        \centering
        \includegraphics[width=4cm,height=2cm]{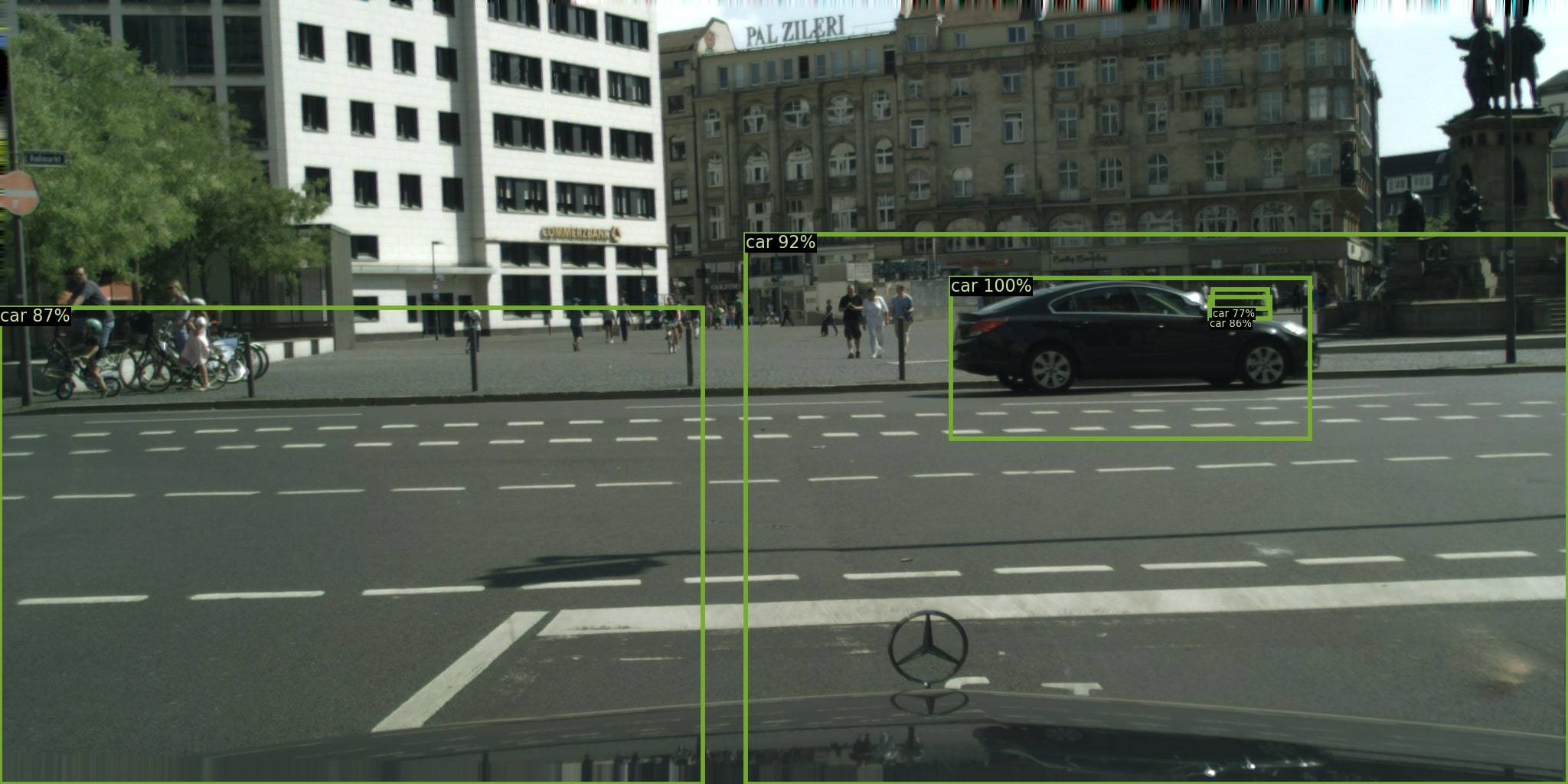}
        \includegraphics[width=4cm,height=2cm]{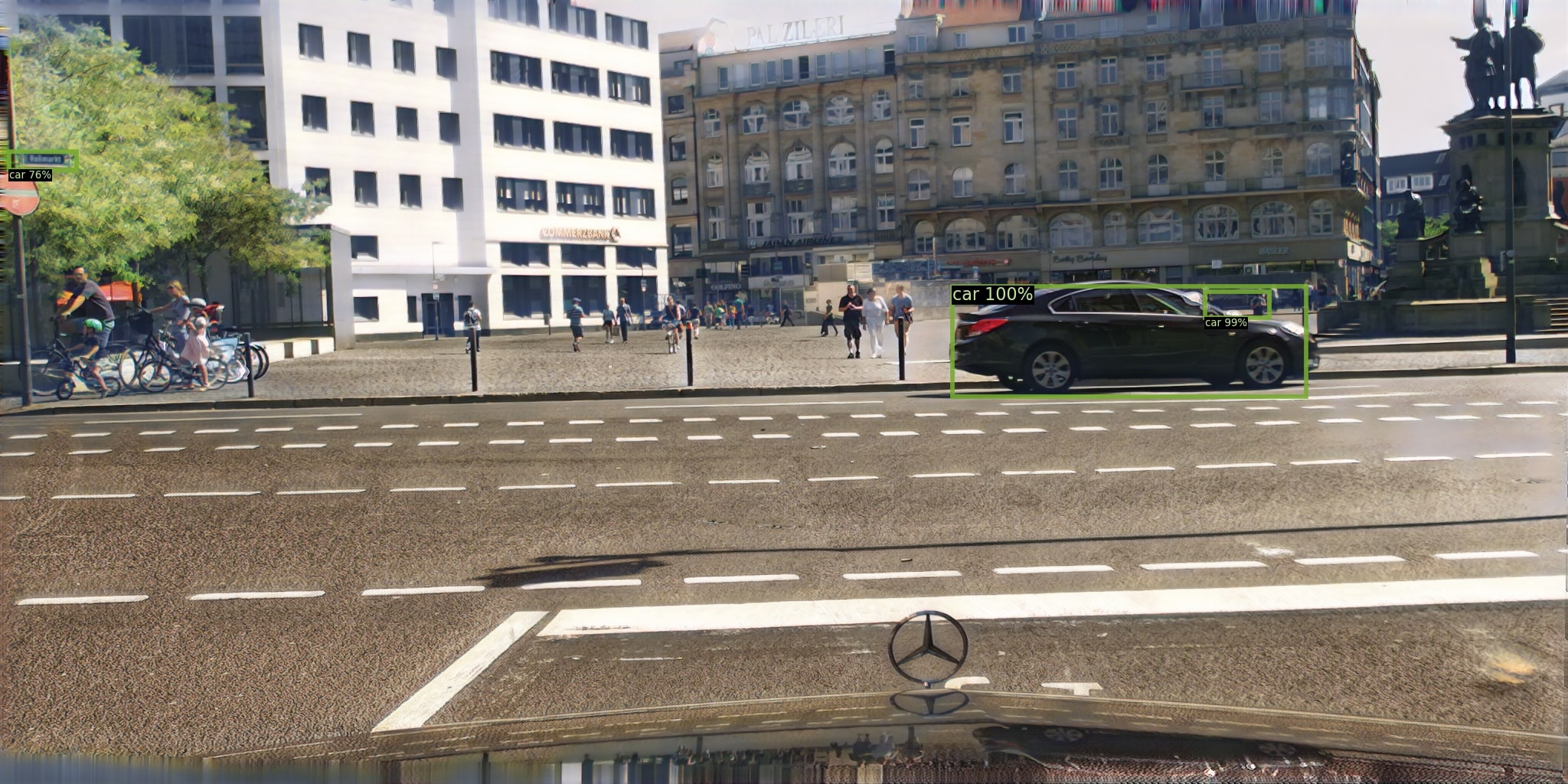}
        \includegraphics[width=4cm,height=2cm]{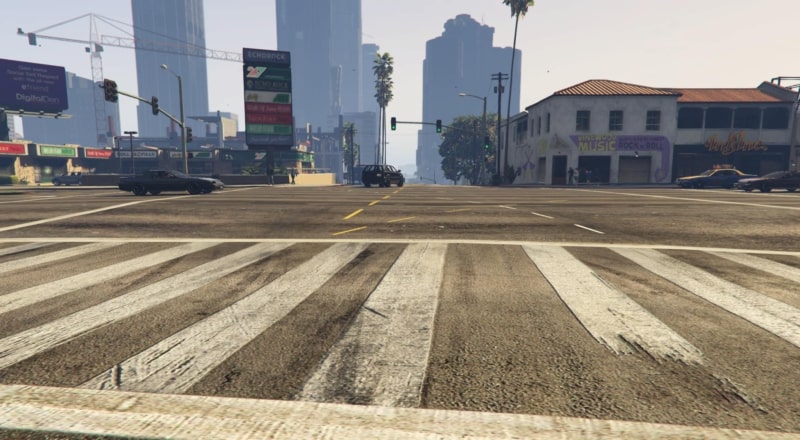}
    \end{subfigure}
    \begin{subfigure}
        \centering
        \includegraphics[width=4cm,height=2cm]{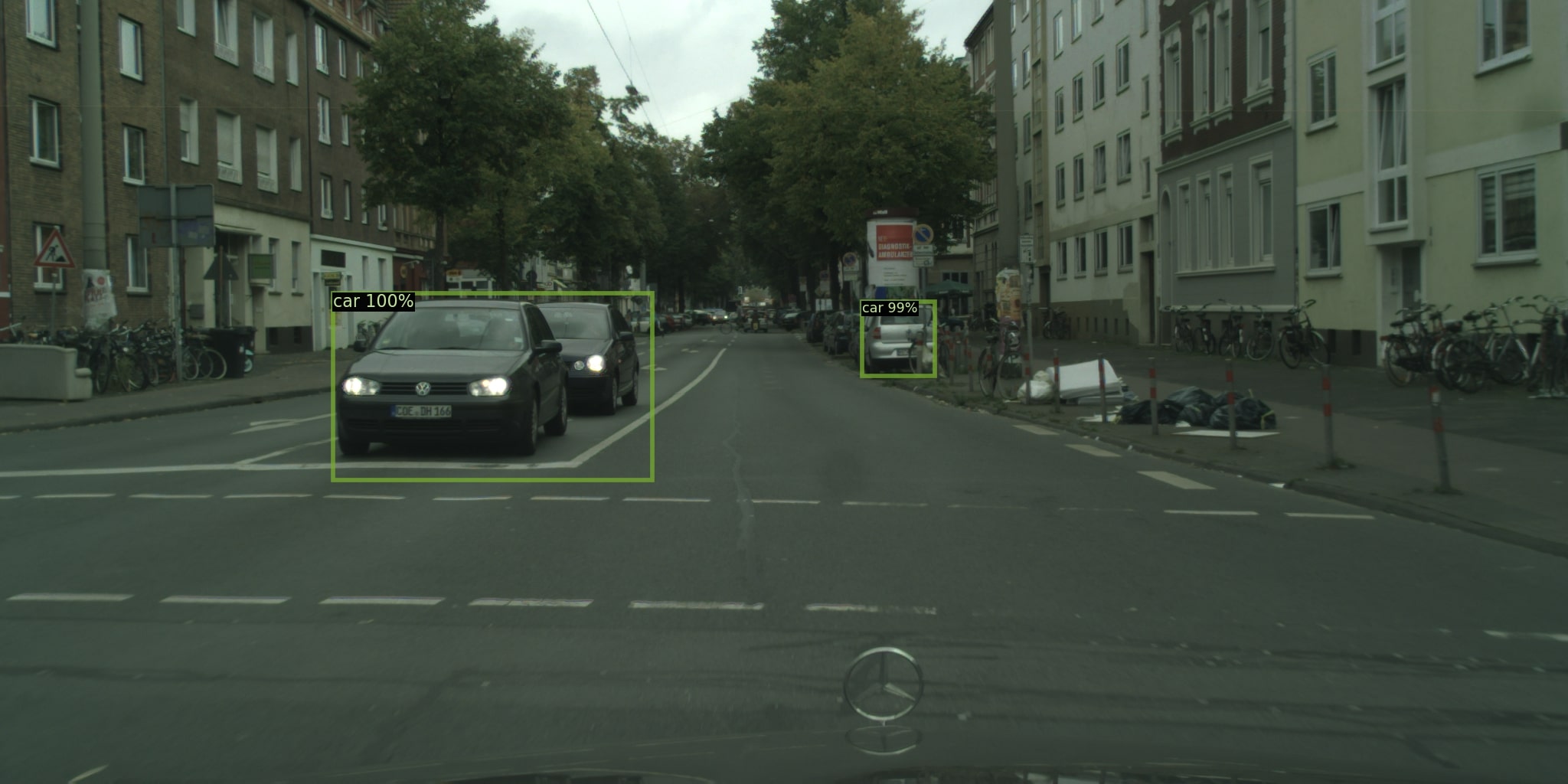}
        \includegraphics[width=4cm,height=2cm]{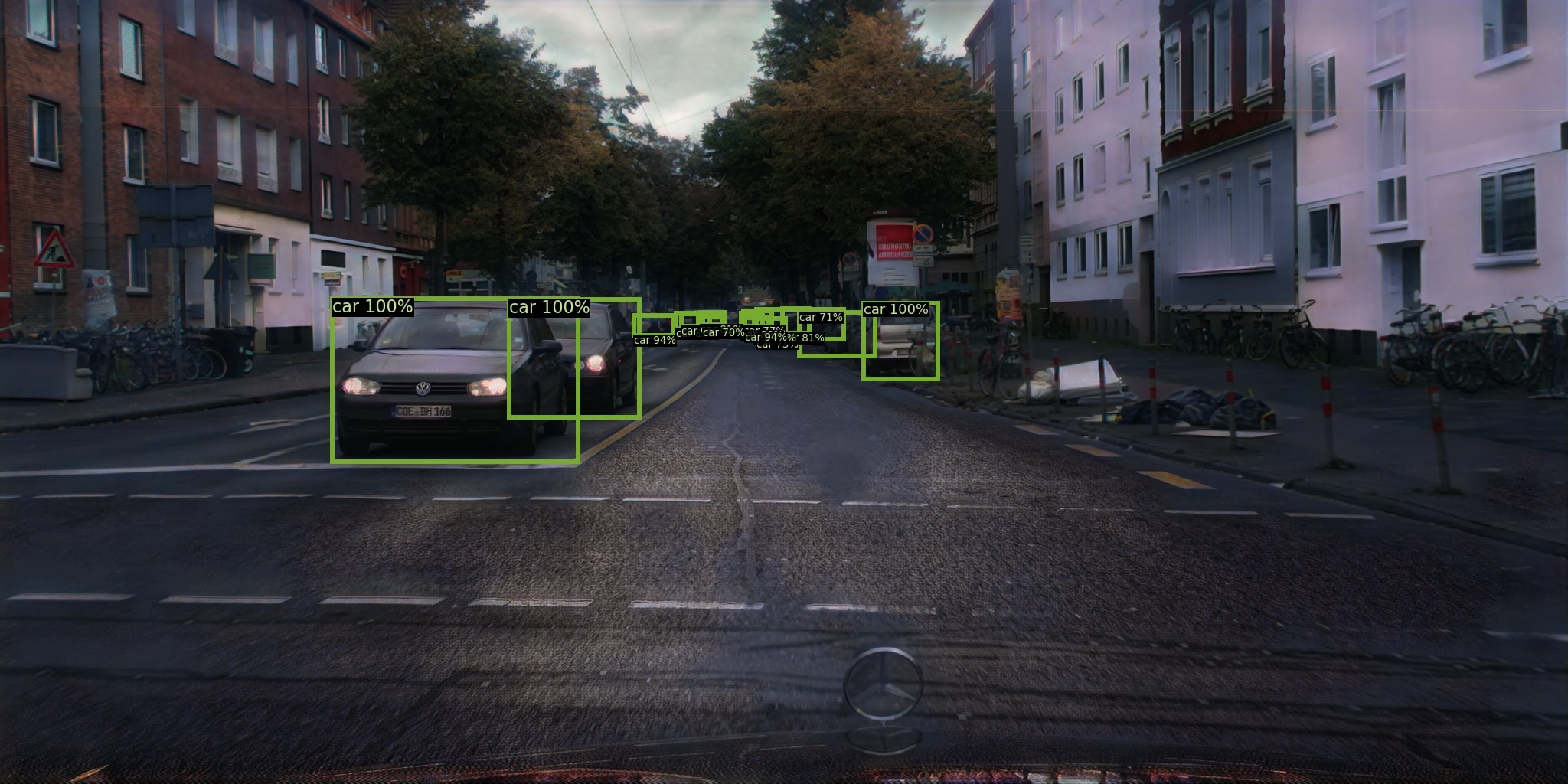}
        \includegraphics[width=4cm,height=2cm]{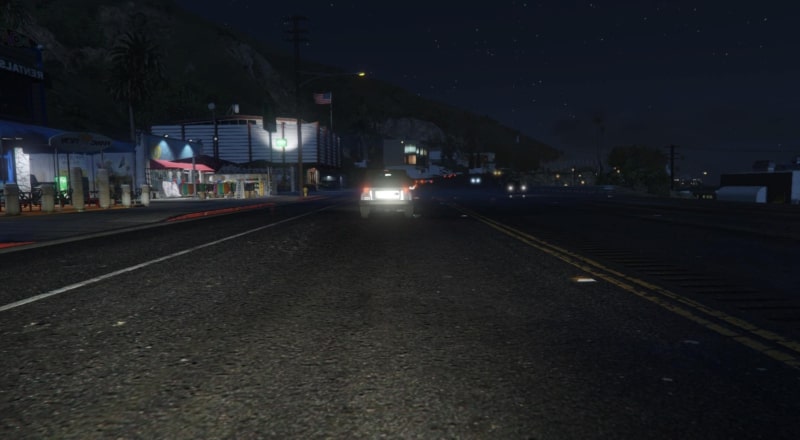}
    \end{subfigure}
    \begin{subfigure}
        \centering
        \includegraphics[width=4cm,height=2cm]{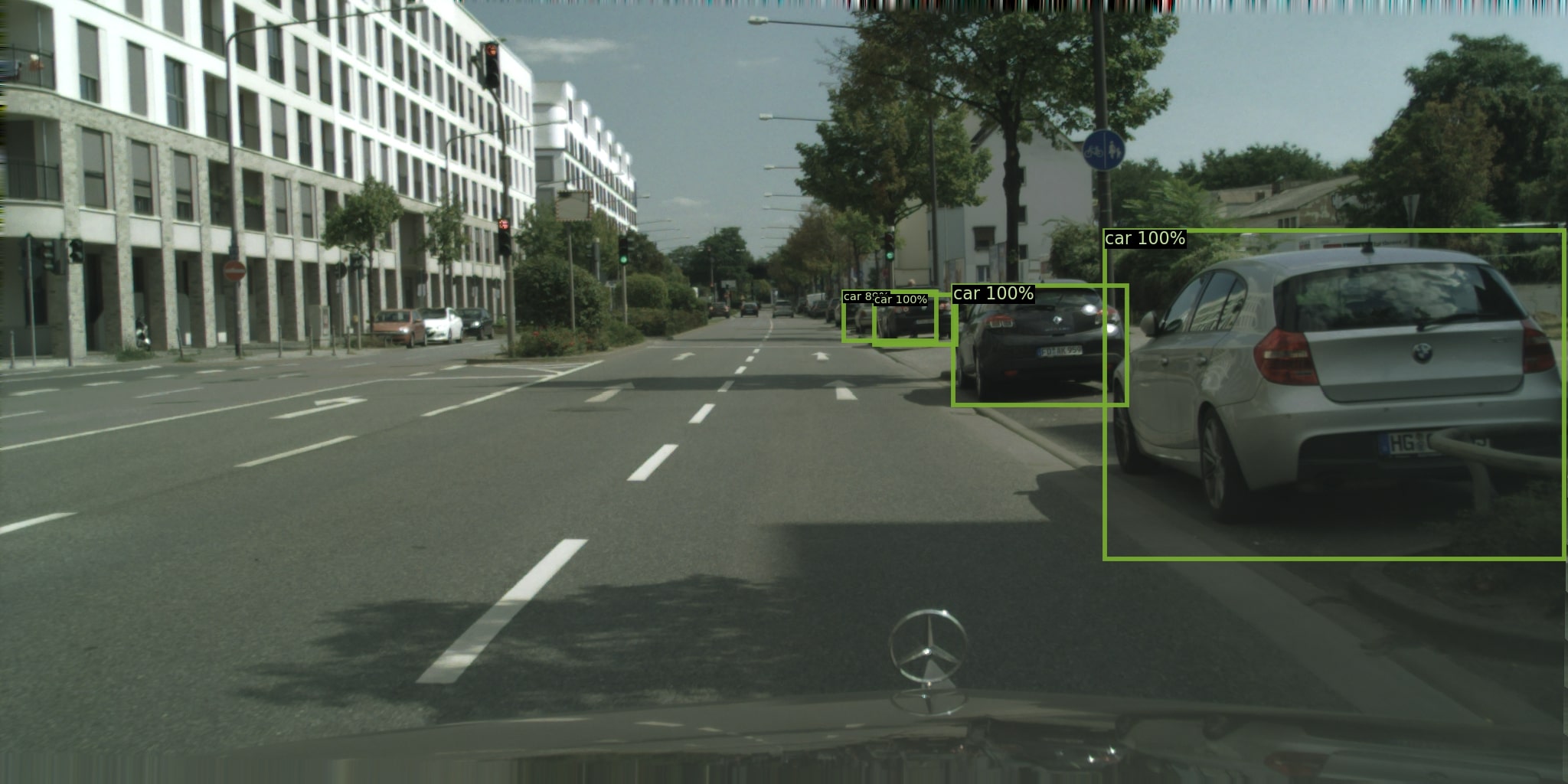}
   \includegraphics[width=4cm,height=2cm]{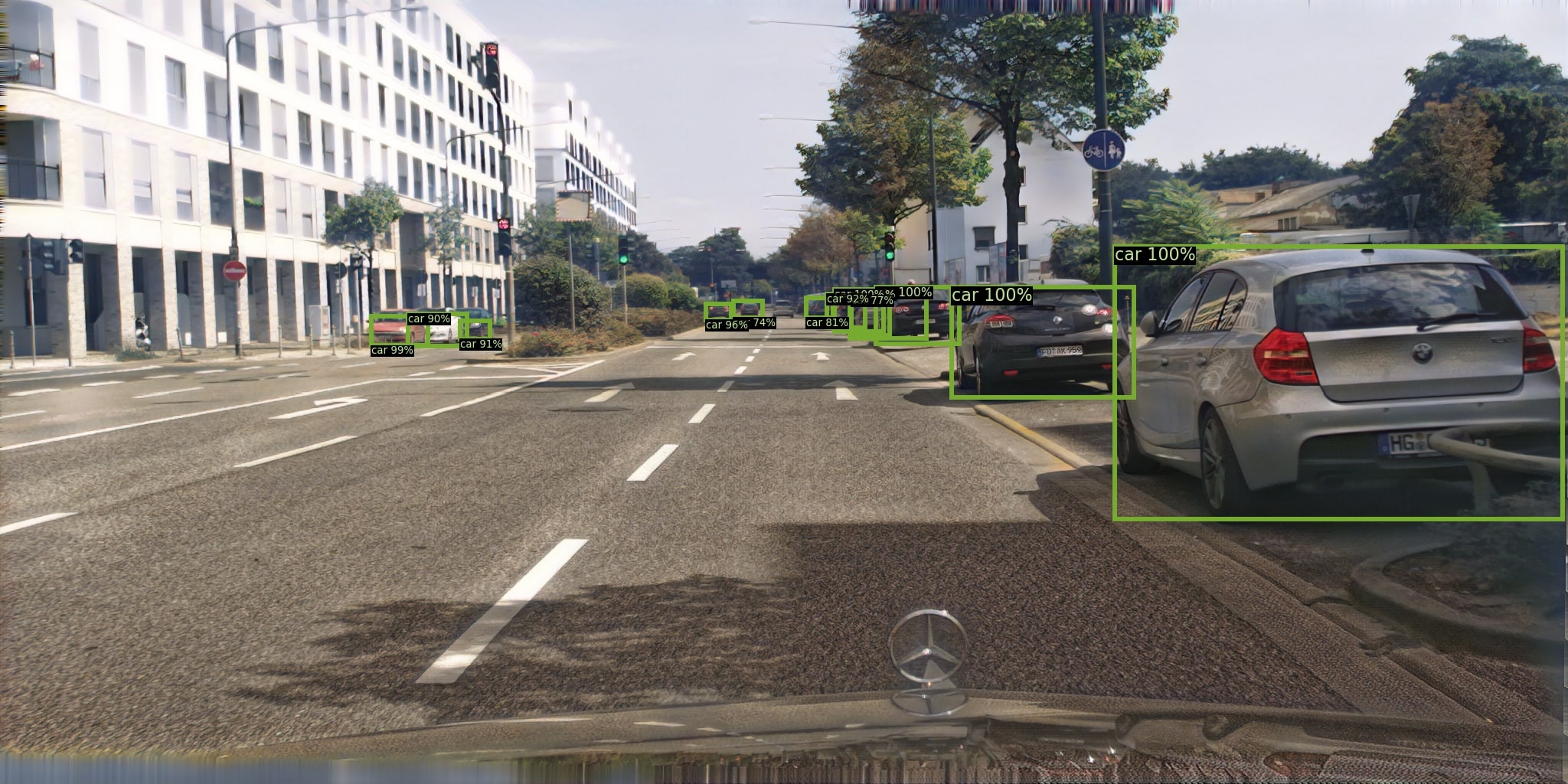}
        \includegraphics[width=4cm,height=2cm]{images/supplementary/sim10k_to_cityscapes/epoch009_real_B.jpg}
    \end{subfigure}

\caption{ The Sim10k $\rightarrow$ Cityscapes adaptation scenario. In all cases the detector model is trained on the Sim10k dataset and evaluated on Cityscapes. We visualise detection inference results on Cityscapes (column 1), inference result on the translated image (column 2) and an (unpaired) real image from the Sim10k dataset, to aid effective translation assessment (column 3). }
\label{fig:supp:sim10k_to_cityscapes}
\end{figure}

\begin{figure}
    \centering
    \begin{subfigure}
        \centering
        \includegraphics[width=4cm,height=2cm]{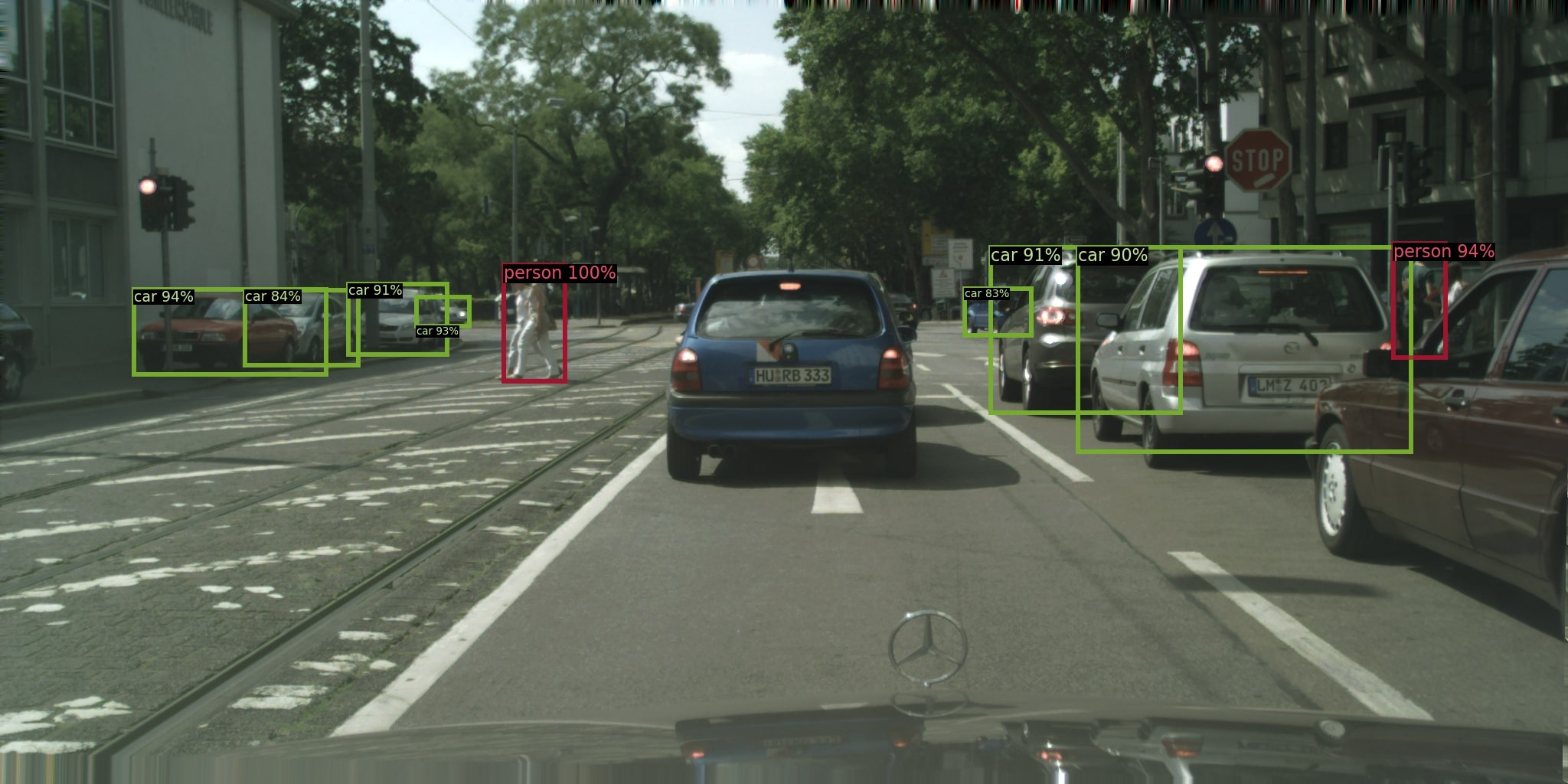}
        \includegraphics[width=4cm,height=2cm]{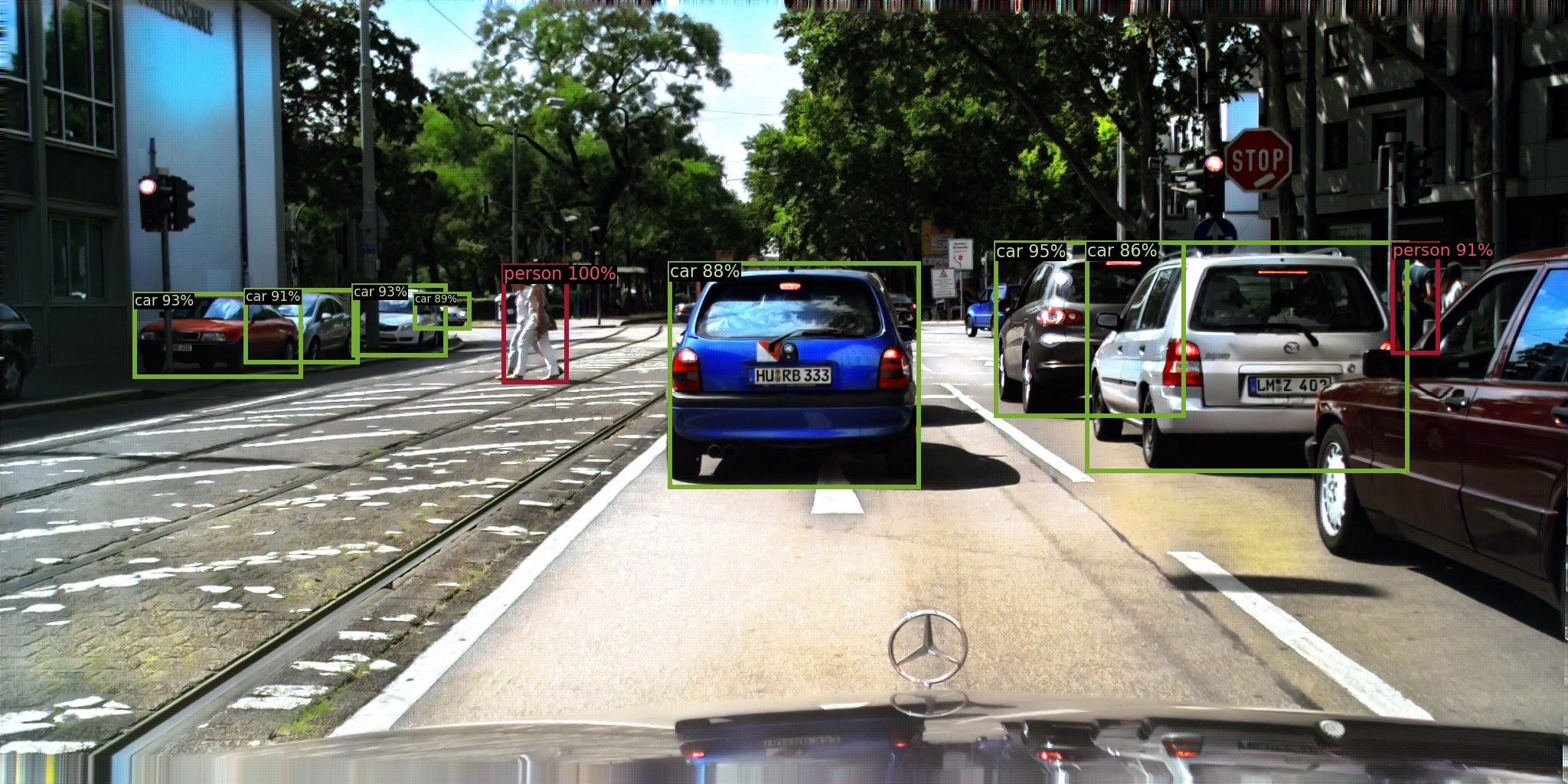}
        \includegraphics[width=4cm,height=2cm]{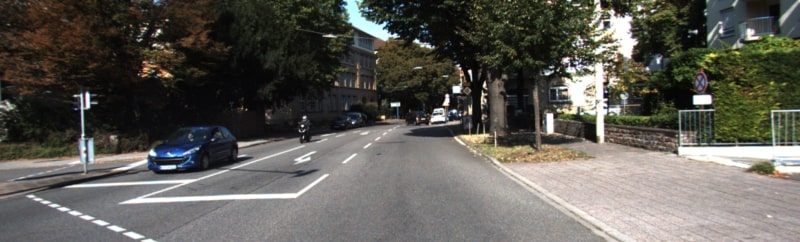}
    \end{subfigure}
    \begin{subfigure}
        \centering
        \includegraphics[width=4cm,height=2cm]{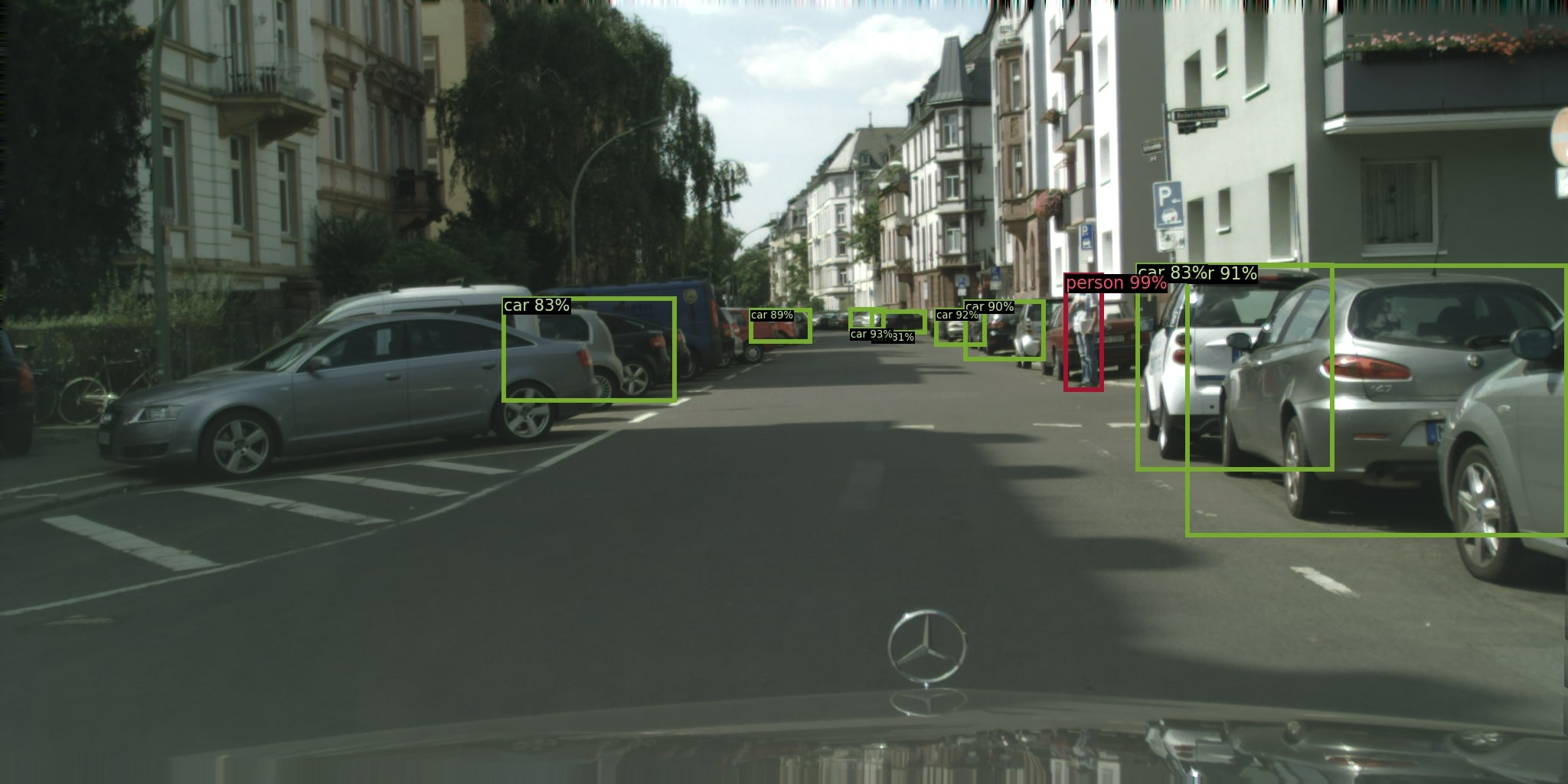}
        \includegraphics[width=4cm,height=2cm]{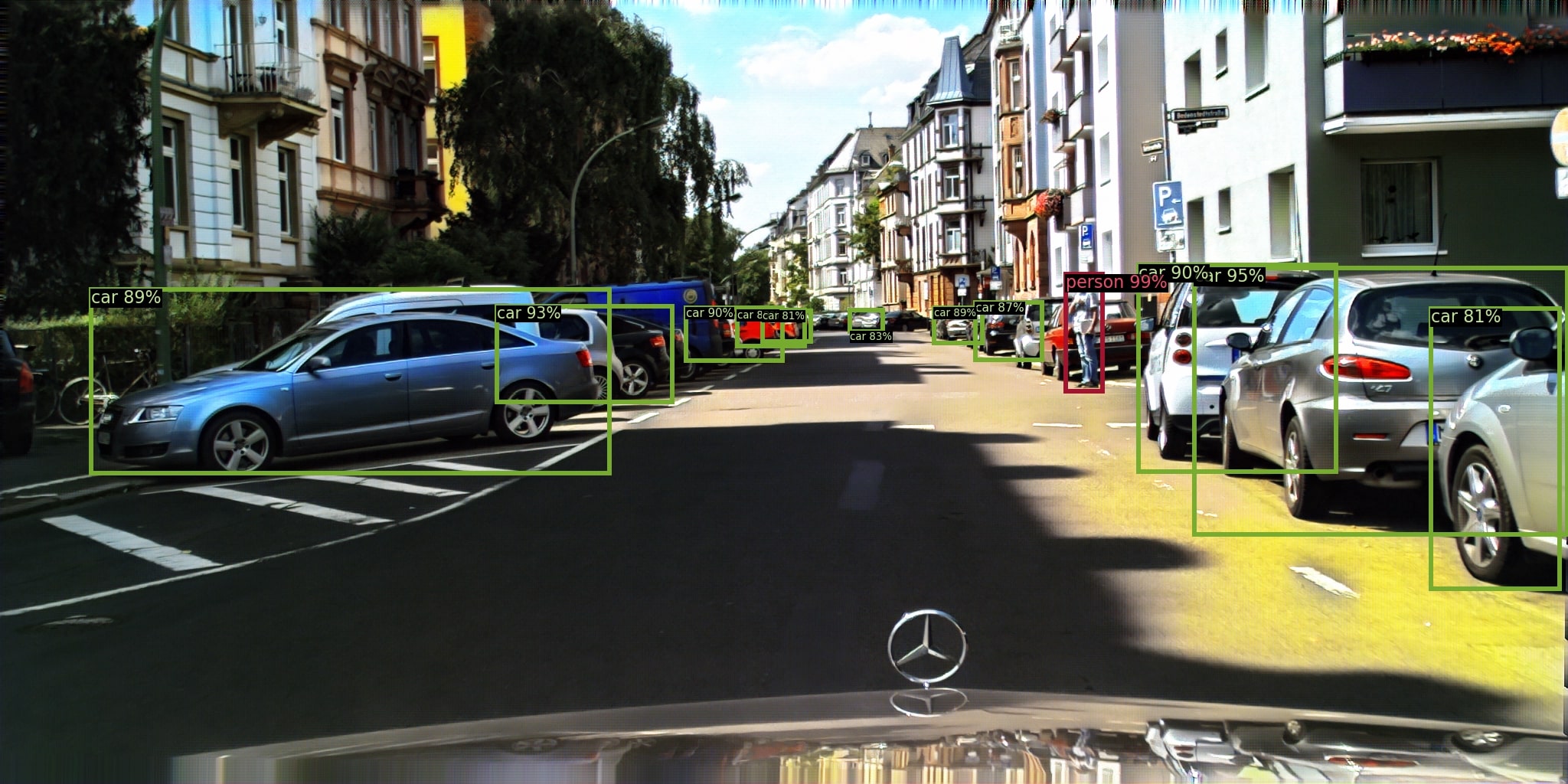}
        \includegraphics[width=4cm,height=2cm]{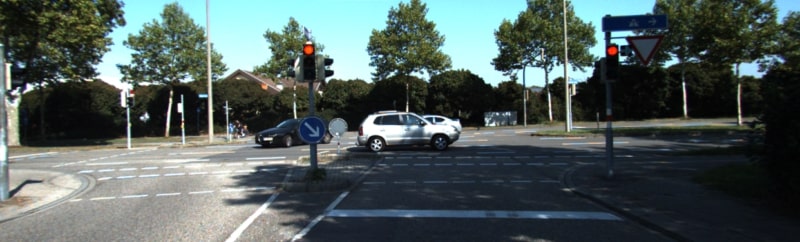}
    \end{subfigure}
    \begin{subfigure}
        \centering
        \includegraphics[width=4cm,height=2cm]{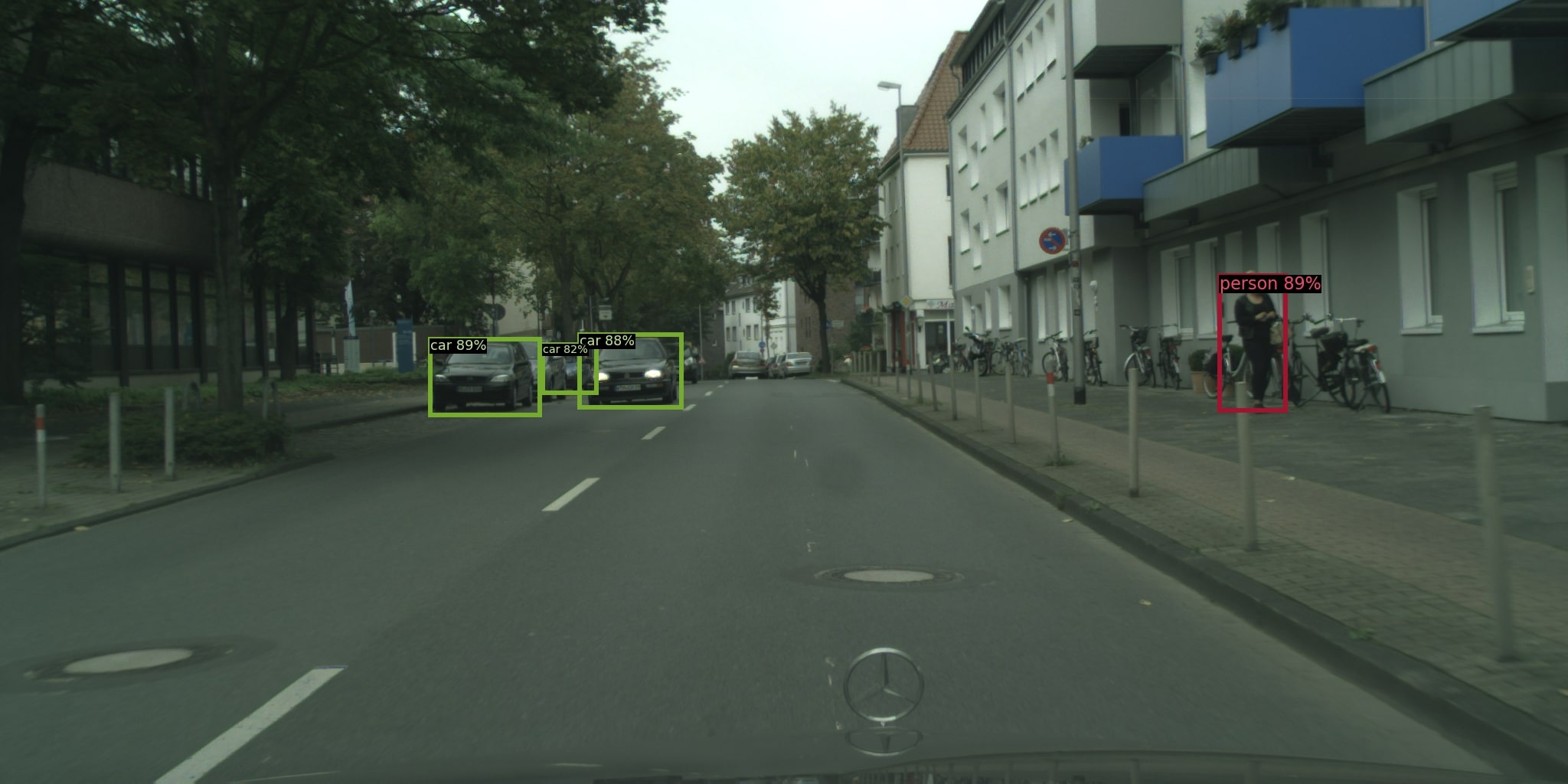}
       \includegraphics[width=4cm,height=2cm]{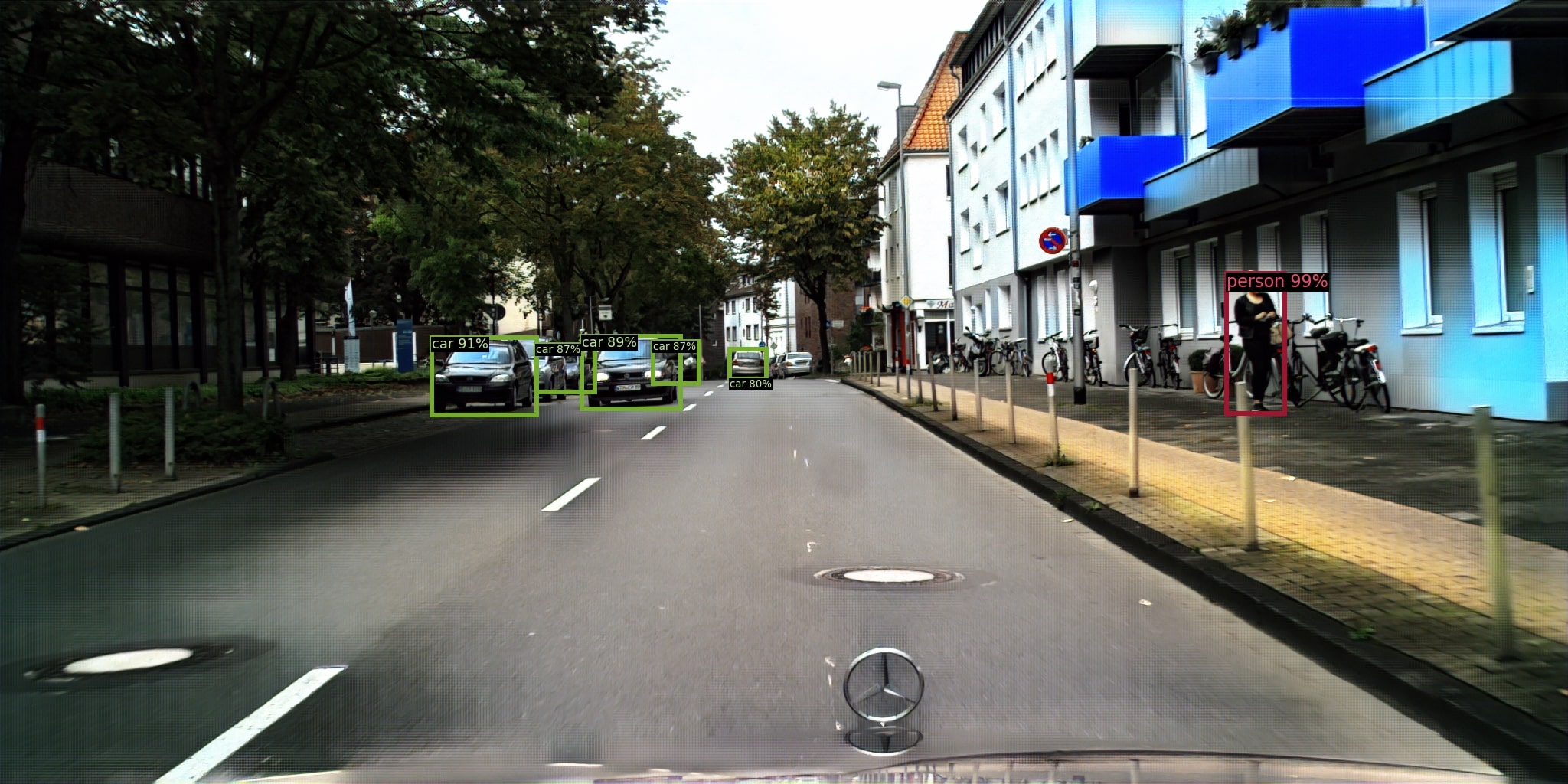}
        \includegraphics[width=4cm,height=2cm]{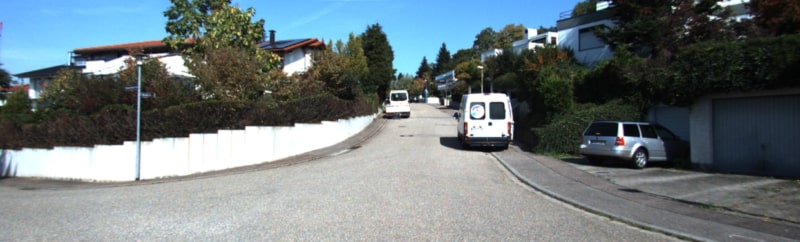}
    \end{subfigure}

\caption{ KITTI $\rightarrow$ Cityscapes adaptation scenario. In all cases the detector model is trained on KITTI dataset and evaluated on Cityscapes. We visualise detection inference results on Cityscapes (column 1), detection inference results on a translated image (column 2) and an (unpaired) real image from the KITTI dataset, to aid effective translation assessment (column 3). }
\label{fig:supp:KITTI_to_cityscapes}
\end{figure}

We show additional visualisations of the learned attention masks, for our two instantiations of $G_A$, in Fig.~\ref{fig:supp:attention}. We observe that in the supervised case the attention masks learn to explicitly focus on detection target instances. In the self-supervised local-global case, even if foreground / background separation in relation to detection targets is less clear, we find the model is still able to learn to disentangle the semantic content and focus on regions that contain objects. We illustrate further qualitative detection performance results, under the three adaptation scenarios studied in the main paper, in Figs.~\ref{fig:supp:sim10k_to_cityscapes}, \ref{fig:supp:KITTI_to_cityscapes} and \ref{fig:supp:foggy_to_cityscapes}. The provided examples further illustrate adaptation gains and the ability of the proposed method to improve cross domain detection performance.


\subsection{Sensitivity analysis}
\label{sec:supplementary:ablation}

We examine the impact of the proposed components in detail and report results in Tab.~\ref{tab:ablation:method}. We select the \mbox{Foggy Cityscapes $\rightarrow$ Cityscapes} adaptation scenario and train the proposed model architecture under the following ablations; $(i)$ \textit{without} the $G_A$ network and \textit{without} the proposed attention module, $(ii)$ \textit{with} the $G_A$ network, \textit{with} the proposed attention module and \textit{without} loss $\Lb_{G_A}$, $(iii)$ \textit{with} the $G_A$ network, \textit{with} the proposed attention module and \textit{with} an unsupervised $\Lb_{G_A}$ loss and finally $(iv)$ \textit{with} the $G_A$ network, \textit{with} the proposed attention module and \textit{with} a supervised $\Lb_{G_A}$ loss. All models are trained under identical settings which are reported in Sec~\ref{sec:supplementary:imp}. In all cases we use the adversarial loss $\Lb_{adv}$ found in Eq.~\eqref{eq:adv} and the InfoNCE loss found in Eq.~\eqref{eq:nce} of the main paper.

We observe that detection performance is lower in case $(i)$ where all method components under consideration are absent. In case $(ii)$ performance is improved by 0.5--1.7\% mAP@.5 which we attribute to the addition of the proposed attention module, trained without any guidance (i.e. $\Lb_{G_A}=0$). Inclusion of all components results in 2.3--2.6\% mAP@.5 gains for the unsupervised model (case $(iii)$) and 2.3--4\% mAP@.5 gains for the supervised model (case $(iv)$). We note that when using the Res-Net-101-FPN backbone, our unsupervised model (mAP@.5 49.1) outperforms its supervised counterpart (mAP@.5 48.0), highlighting the potential of our unsupervised proposal. Our unsupervised loss not only guides the attention generator towards disentangling background and foreground, but additionally improves representation learning at the level of the encoder network. We conjecture that this is more effective on higher capacity networks such as Res-Net-101. Finally, the ablation study provides quantitative evidence towards verifying the efficacy of the individual components under the proposed method.

\begin{table*}[ht]
\footnotesize
\begin{adjustbox}{scale=0.68}
\begin{tabular}{c | c | c c c c c c c c | c c} 

\toprule

Det. backbone & $G_A$   &  $\Lb_{G_A}$ & $\Lb_{G_A}$ type  & Attention &  mAP@[.5:.95] & mAP@.5 &  mAP@.75  & mAP@[.5:.95] & mAP@[.5:.95] & mAP@[.5:.95]  \\
 
 &   &   &  &  &   & &   &  \scriptsize{small} &  \scriptsize{medium}  &  \scriptsize{large}  \\
\hline
 & & & - & &    23.0  & 42.7  & 21.8 & 2.2  &20.8 & 47.4 \\

 & {\hspace{0.2cm} \checkmark} &  & - &  {\hspace{0.2cm} \checkmark}  & 23.5 & 44.4 & 20.8  & 2.5 & 22.3 & 46.3  \\

\hspace{0.4cm} R-50-C4**&{\hspace{0.2cm} \checkmark}  & {\hspace{0.2cm} \checkmark}  &  unsupervised &  {\hspace{0.2cm} \checkmark}   & 24.1 & 45.3 & 23.2 & 2.6 & 23.3 & 47.1    \\

&{\hspace{0.2cm} \checkmark}  & {\hspace{0.2cm} \checkmark}  &  supervised &  {\hspace{0.2cm} \checkmark}   & 24.5 & 46.7  & 22.9 & 2.7 & 23.4 & 47.1  \\

\hline
\hline

 & & & - & &  24.3 & 45.5  & 21.9  & 3.8  &22.4 & 46.1 \\

 & {\hspace{0.2cm} \checkmark} &  & - &  {\hspace{0.2cm} \checkmark}  & 25.2 & 46.2 & 23.2 & 4.0 & 24.2 & 46.8  \\

\hspace{0.4cm} R-101-FPN &{\hspace{0.2cm} \checkmark}  & {\hspace{0.2cm} \checkmark}  &  unsupervised &  {\hspace{0.2cm} \checkmark}   &  26.2 & 49.1 & 24.4 & 5.1 & 25.3 & 47.1    \\

&{\hspace{0.2cm} \checkmark}  & {\hspace{0.2cm} \checkmark}  &  supervised &  {\hspace{0.2cm} \checkmark}   & 26.0 & 48.0  & 24.1 & 4.7 & 25.0 & 46.3  \\

\hline
\hline

 & & & - & & 24.3 & 44.6 &22.7 & 2.5 &21.7 &50.4  \\

 & {\hspace{0.2cm} \checkmark} &  & - &  {\hspace{0.2cm} \checkmark}  & 24.8 & 45.1 & 23.4 &2.5 &23.0 &50.1 \\

\hspace{0.4cm} R-50-DC5&{\hspace{0.2cm} \checkmark}   &{\hspace{0.2cm} \checkmark}  & unsupervised &  {\hspace{0.2cm} \checkmark}   & 25.9 & 47.2 & 24.0 & 2.5 & 24.1 & 50.2 \\

&{\hspace{0.2cm} \checkmark}  & {\hspace{0.2cm} \checkmark}  &  supervised &  {\hspace{0.2cm} \checkmark}   &  25.7 & 46.9 & 24.5 & 2.5 & 23.7 & 51.6   \\

\bottomrule
\end{tabular}
\end{adjustbox}
\caption{Ablation on method components (Foggy Cityscapes $\rightarrow$ Cityscapes). We report the effect of ablating method components in columns 2--5, across multiple detector backbone networks. ** denotes R-50-C4 experimental results, also reported in the main paper. } 
\label{tab:ablation:method} 
\end{table*}

\begin{table*}[ht]
\footnotesize
\setlength\tabcolsep{4pt}

\begin{adjustbox}{scale=0.77}
\begin{tabular}{ l   l l l  l |   c    c c c c c c c  c c  }
\toprule
Model  & backbone & Weight init. &   FPN & Eval. scenario    &  person & rider & car & truck & bus & train & motor & bike &  mAP@.5 $\uparrow$   \\
 \hline
 &  &      &             & source  &  49.1 & 40.3 & 46.7 & 30.3 & 36.8 & 24.0 & 29.1 & 42.8 & 37.4    \\
R-50-FPN   &ResNet-50    & COCO & {\hspace{0.2cm} \checkmark}           & target oracle    &  74.9 & 59.7& 60.5& 44.0 & 69.0& 58.6 & 50.1 & 52.2 & 58.6  \\

 &  &  &             & ours unsupervised   & 67.0 & 54.5 & 58.6  & 39.7  & 57.1 
 & 44.3 & 41.3 & 49.2 & 51.4    \\





\hline
   &  &       &          & source & 46.6& 38.6& 45.1& 20.6& 35.6& 10.5& 29.7 & 40.3 & 33.4 \\

R-50-FPN  &ResNet-50 & ImageNet  &  {\hspace{0.2cm} \checkmark} &target oracle                 &74.8& 57.8& 61.0& 42.2& 67.0& 50.0& 50.0& 52.1 & 56.9 \\

   & &  &               & ours unsupervised  & 69.1& 52.9& 59.3& 36.5& 57.3& 43.8& 46.1& 48.6 & 51.7\\

\hline

   &  &       &             & source  & 35.5 & 38.7 & 41.5 & 18.4 & 32.8 & 12.5 & 22.3 & 33.6 & 29.4 \\

R-50-C4**    &ResNet-50 &    ImageNet             && target oracle    &47.5& 51.7 & 66.9 & 39.4 & 56.8  & 49.0 & 43.2 & 47.3 & 50.2 \\

    &  &  &      &       ours unsupervised  & 43.2 & 50.1 & 61.7 & 33.3 & 48.6  & 47.8 & 35.2 & 42.6 & 45.3 \\

\hline
\hline 
\hline 

 & &     &             &source & 35.4 & 39.3 & 43.8 & 22.3 & 34.9  & 8.9  & 23.3 & 34.1  &  30.2 \\

R-101-C4   &ResNet-101 &    ImageNet&             & target oracle  & 46.8 & 49.6 & 67.4 & 40.5 & 60.6 & 52.7  &42.7  &45.4    &  50.7 \\

  &   &   &      & ours unsupervised & 43.1    & 48.2 & 62.4 & 35.9 & 51.7 & 46.0 & 36.3  &44.3  & 46.0   \\

\hline

  &  &     &             &   source & 38.1 & 43.0 & 45.2 & 24.9 & 37.3  & 27.4 & 24.7 & 37.9 & 34.8  \\

R-101-FPN   &ResNet-101 &   ImageNet  &   {\hspace{0.2cm} \checkmark}         &   source   & 54.2 & 57.7 & 72.7 & 44.8  & 59.8  & 47.3 & 45.9 &48.0  & 53.8  \\

  &  &   &      &    ours unsupervised       & 49.8 & 54.1 & 66.7 & 41.4 & 52.4   & 46.3  & 37.4  & 37.4 & 49.1   \\

\hline

   &&    &             &source   & 36.5 & 41.3 & 46.6 & 26.8 & 37.1 & 16.0 & 27.2 & 37.5 &  33.6  \\

R-101-DC5  &ResNet-101  &   ImageNet  &           &    target oracle  & 46.3& 51.0  &67.5  &43.8  &62.0   &52.3 & 43.7 & 47.1 & 51.7  \\

 &   &    &      & ours unsupervised & 44.0 & 48.6  & 62.6 & 40.2& 55.4 & 42,7 &  37.3  & 44.3 & 46.9    \\

\hline

 & &    &             &   source & 35.9 & 38.1 & 45.5 & 32.8 & 29.3 & 25.0 & 22.5 &  29.7 & 32.4  \\

Retina-101-FPN  &ResNet-101     &ImageNet             &{\hspace{0.2cm} \checkmark}    & target oracle &   42.7 & 48.1 & 64.5 & 38.5 & 50.5 & 37.6 & 35.1 & 39.2 & 44.7  \\

 &  &   &      &     ours unsupervised    & 41.1 & 46.3 & 60.5 & 37.3 &47.4   &36.4  & 31.5 &  37.8&42.3  \\
\bottomrule

\\

\end{tabular}
\end{adjustbox}
\caption{Sensitivity analysis considering backbone model architecture and detector training settings (Foggy Cityscapes $\rightarrow$ Cityscapes). ** indicates experimental results, reported in the main paper. } 
\label{tab:ablation:detector} 
\end{table*}

We conduct a further sensitivity study on detector training settings and backbone model architectures. Results are found in Tab.~\ref{tab:ablation:detector}. We select the \mbox{Foggy Cityscapes $\rightarrow$ Cityscapes} adaptation scenario for our analysis. Towards fair comparison with existing 
work, all results reported in the main paper follow the common experimental setup as described in multiple previous works~\cite{li2022sigma, zhao_wang_2022, deng_li_chen_duan_2021, Zhou2022MultiGranularityAD, DBLP:conf/aaai/LiLYY22, rezaeianaran2021seeking}; i.e.~making use of a Faster-RCNN model with a Res-Net-50-C4 backbone. Experimentally, we additionally consider and evaluate a total of six backbones: Res-Net-50-C4, Res-Net-50-FPN, Res-Net-101-C4, Res-Net-101-FPN, Res-Net-101-DC5 and Retina-101-FPN. Futher details regarding the aforementioned architectures are found in~\cite{wu2019detectron2}. All models are trained using the hyperparameters and settings reported in Sec.~\ref{sec:supplementary:imp}.


For every experiment we report detection performance on imagery pertaining to \textit{source}, \textit{target oracle}, and \textit{local-global}; which refer to images obtained from Foggy Cityscapes, Cityscapes datasets and images generated by our self-supervised local-global model, respectively. We observe consistent gains, over the baseline \textit{source} evaluation scenario, that range from 9.9--18.3\% mAP@.5 under all considered backbones architectures. Models that incorporate the FPN module~\cite{8099589} (denoted as *-FPN) consistently outperform the baseline backbones (denoted as *-C4), with Retina-101-FPN being an exception. We additionally observe that pre-training the detector on the COCO dataset~\cite{Lin2014MicrosoftCC} improves both source and target domain performance which may be attributed to the fact that the model initialisation has then been optimised for an object detection task. 

Our exploration of additional model backbones allows us to evidence scenarios in which our approach, promisingly, brings us to within $2.5\%$ of target oracle performance (\eg Retina-101-FPN). Finally, we note that mAP accuracy increases cf.~our main paper results are possible, however we opt to retain the R-50-C4 setting in our manuscript, towards highlighting fair comparisons.

\subsection{Comparison with contrastive learning I2I translation methods}
\label{sec:supplementary:compare}

We further evaluate the quality of the translated images using standard I2I translation metrics. Tab.~\ref{tab:results:foggy_ours} reports Frechet Inception Distance~\cite{heusel2017gans} (FID) and Kernel Inception Distance~\cite{Binkowski2018DemystifyingMG} (KID), comparing our approach with images generated using CUT~\cite{park2020contrastive}, FeSeSim~\cite{zheng_cham_cai_2021} and Qs-Attn~\cite{hu_zhou_huang_shi_sun_li_2022}. We use object labels to explicitly evaluate image quality in regions that contain object instances; by computing the aforementioned metrics exclusively only in those regions. These derivative metrics are denoted $\text{FID}_{\textsc{inst}}$ and $\text{KID}_{\textsc{inst}}$, respectively. Interestingly, our method shows improvement in standard I2I translation metrics. Large improvements can be found in our object-region specific metrics, in the case when object labels are available.

\begin{figure}
\centering
\label{}\includegraphics[width=2.3cm]{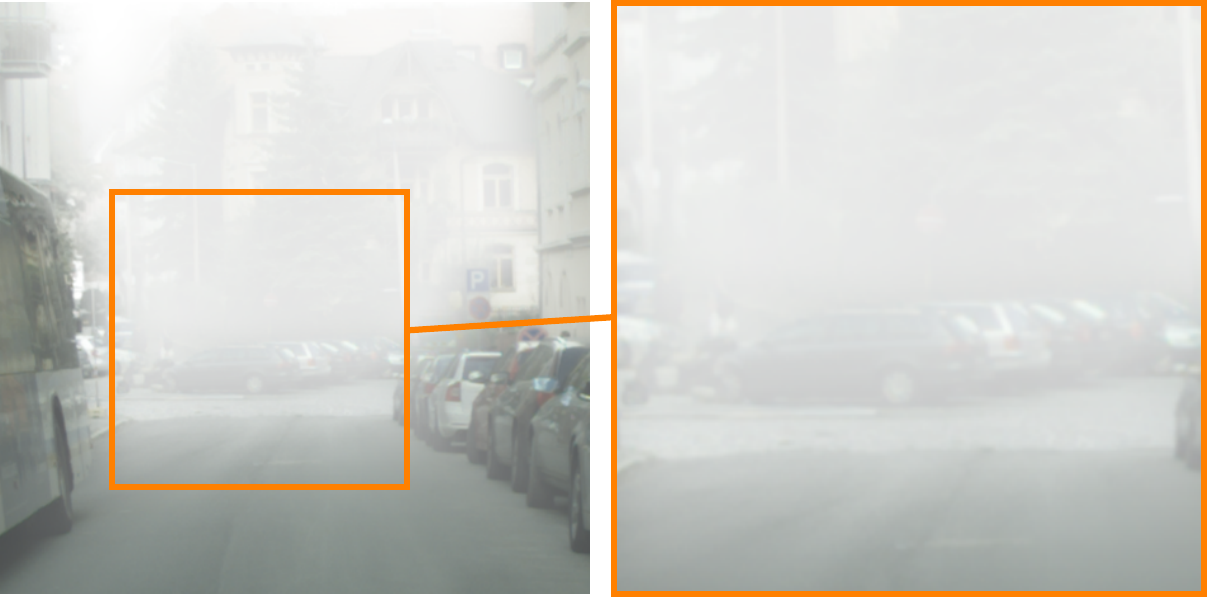}\hfill
\label{}\includegraphics[width=2.3cm]{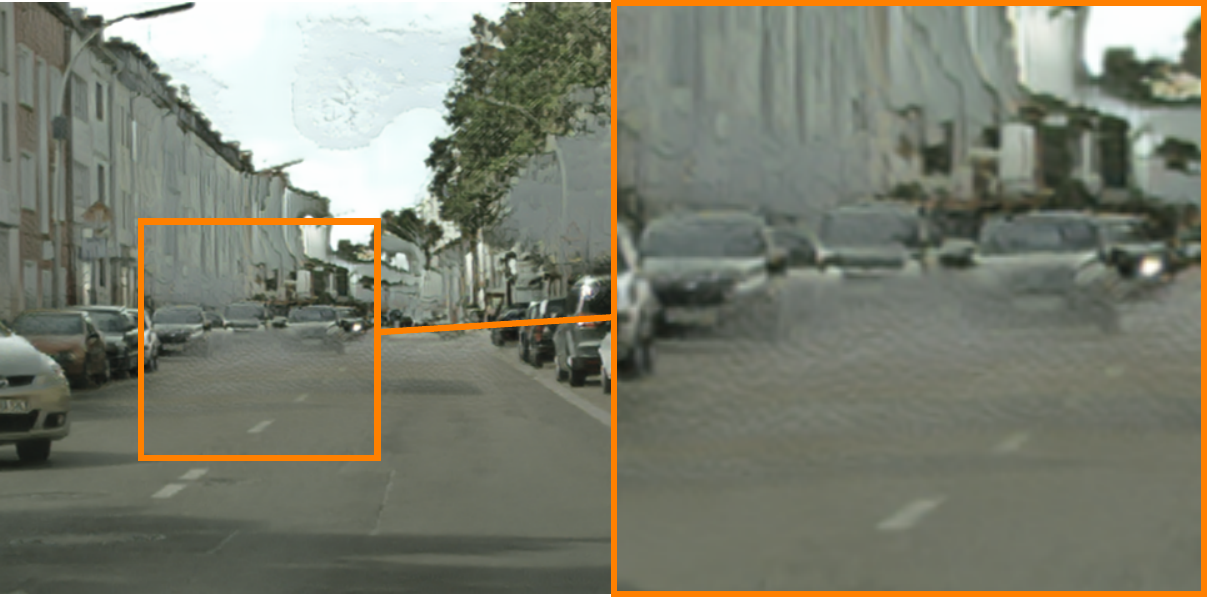}\hfill
\label{}\includegraphics[width=2.3cm]{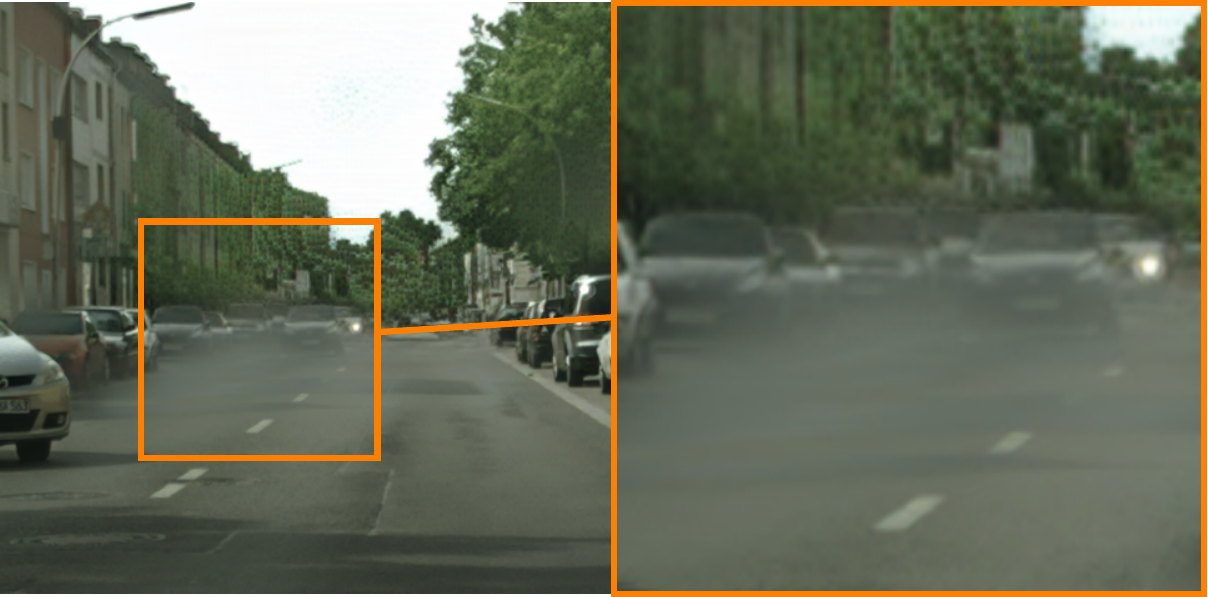}\hfill
\label{}\includegraphics[width=2.3cm]{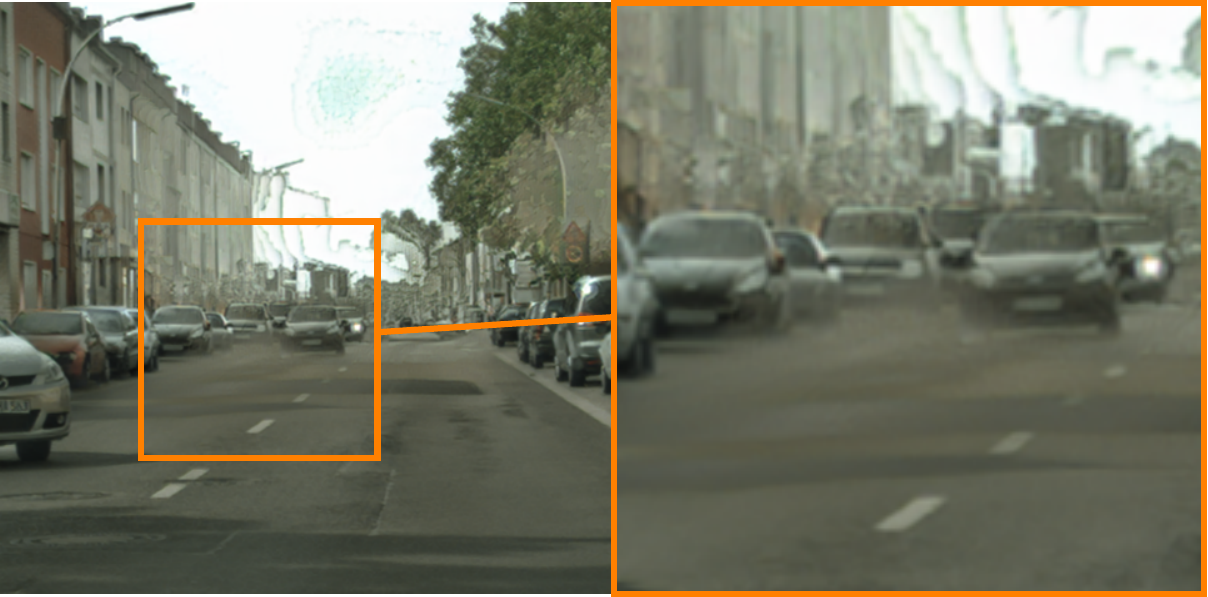}\hfill
\label{}\includegraphics[width=2.3cm]{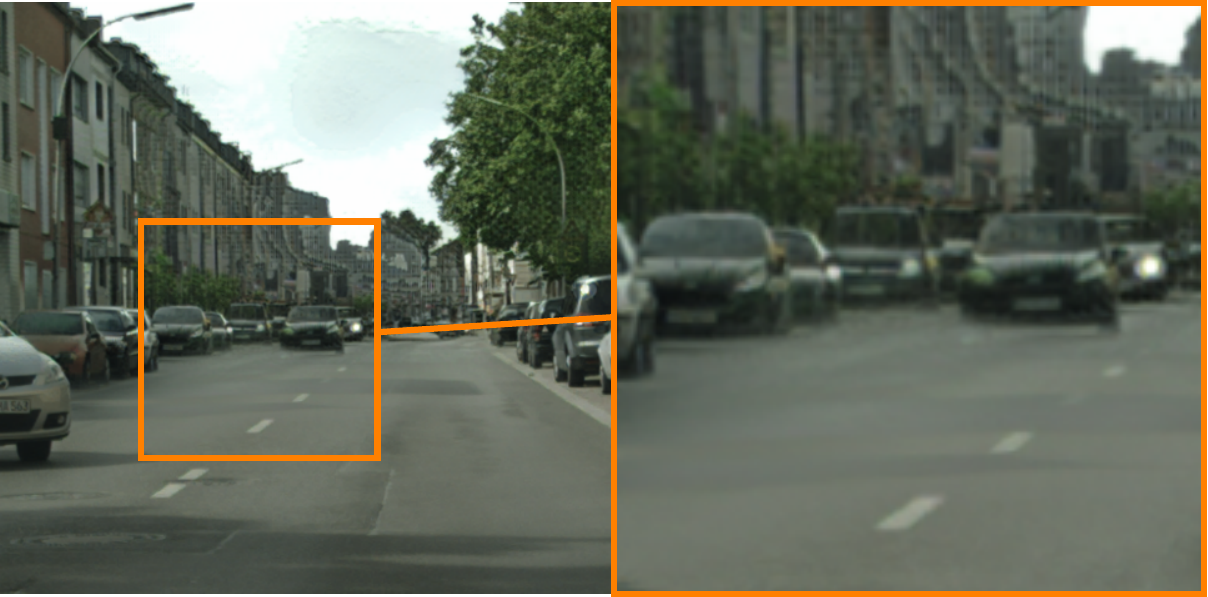} \\


\label{}\includegraphics[width=2.3cm]{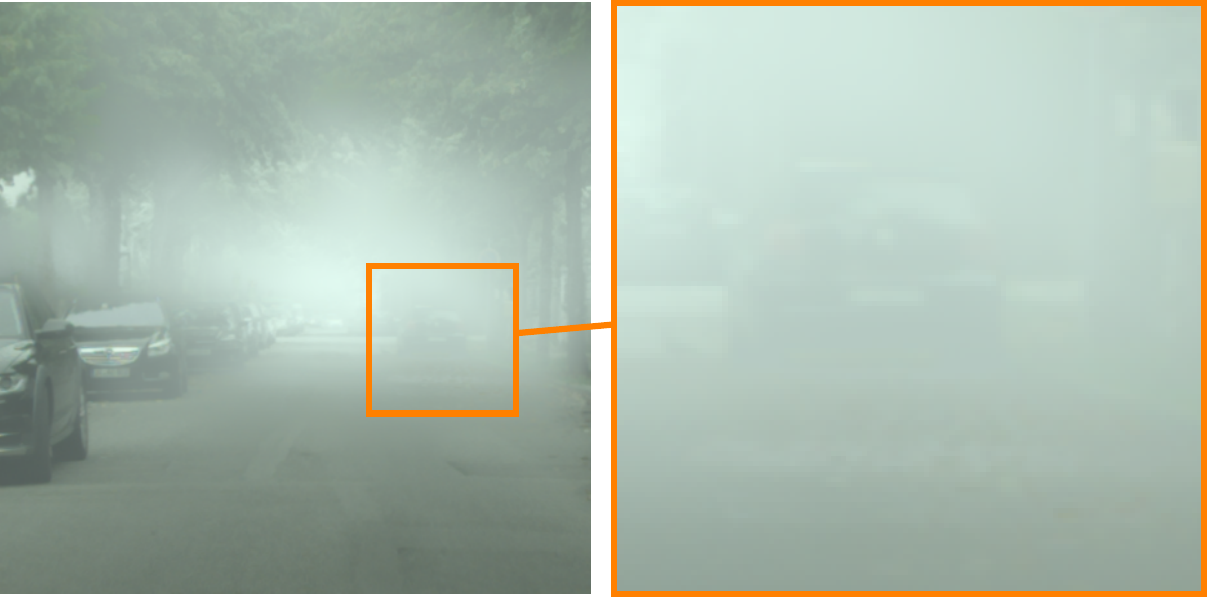}\hfill
\label{}\includegraphics[width=2.3cm]{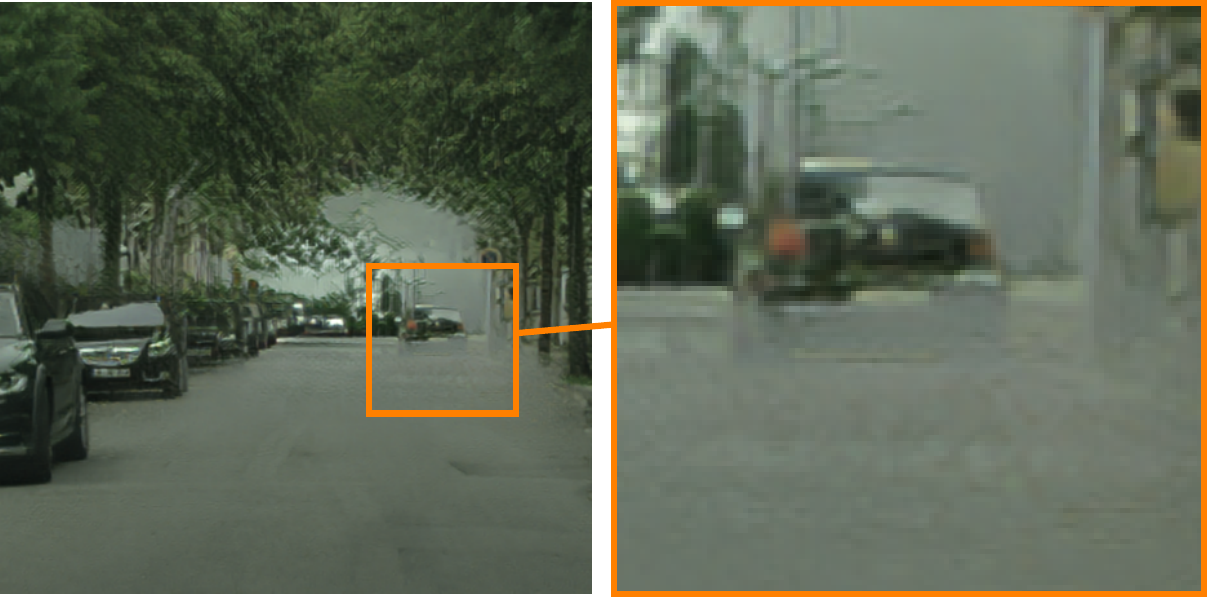}\hfill
\label{}\includegraphics[width=2.3cm]{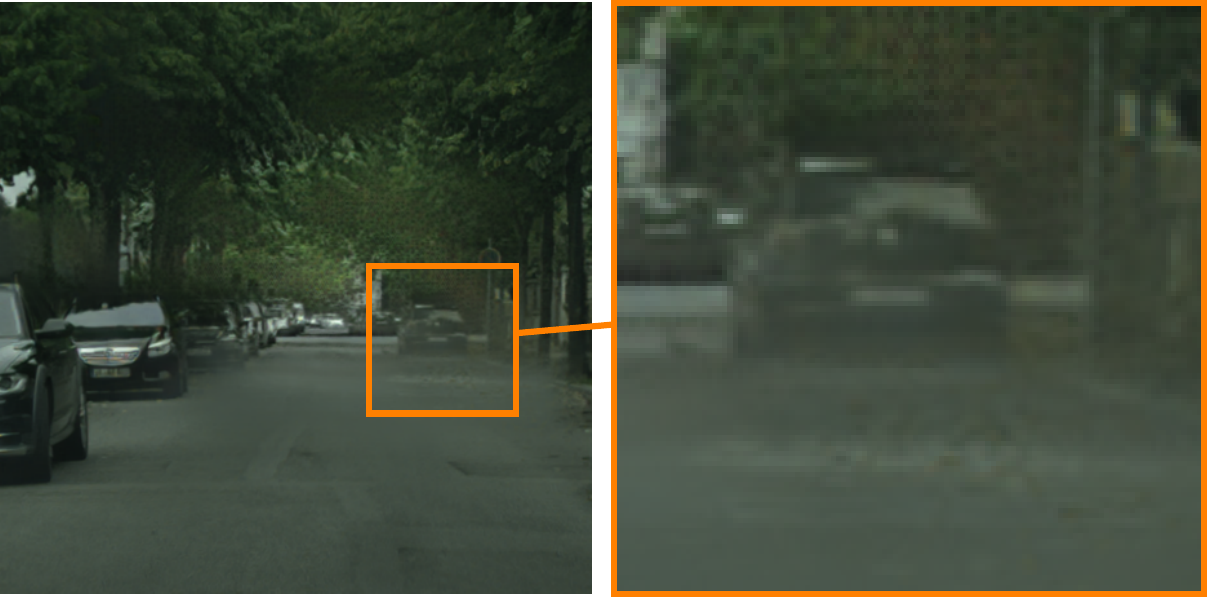}\hfill
\label{}\includegraphics[width=2.3cm]{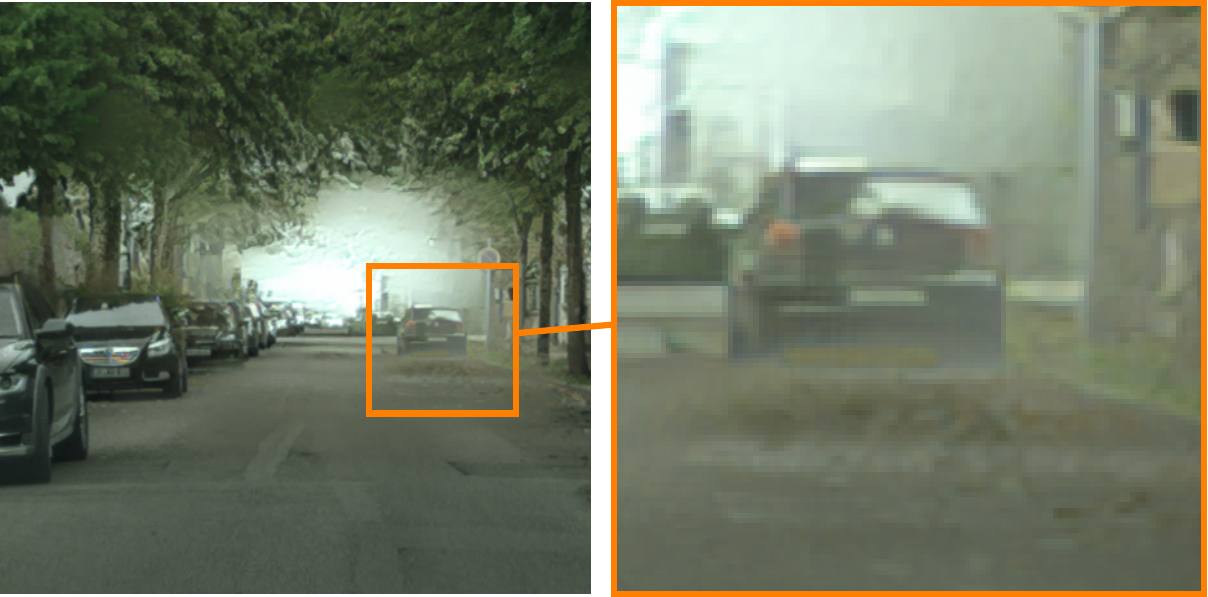}\hfill
\label{}\includegraphics[width=2.3cm]
{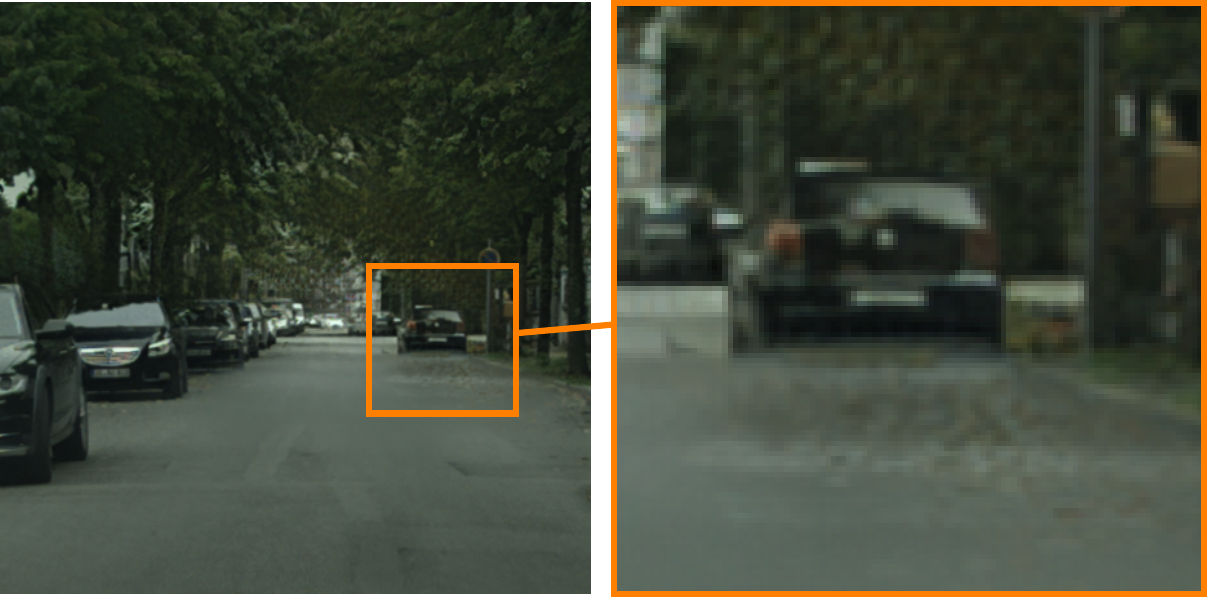} \\

\label{}\includegraphics[width=2.3cm]{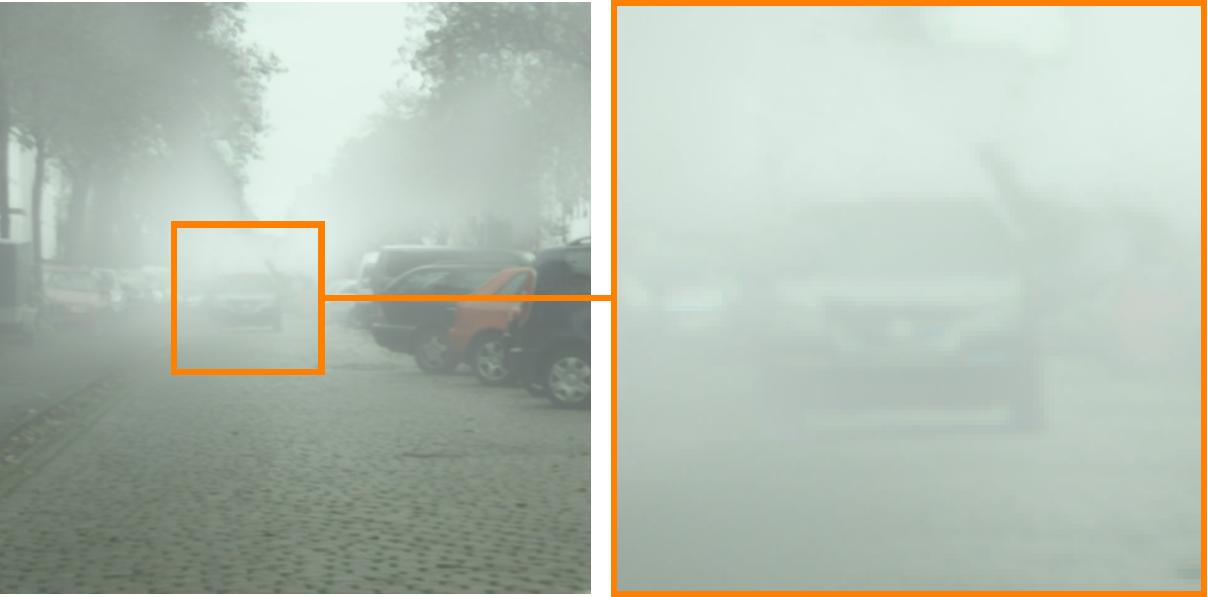}\hfill
\label{}\includegraphics[width=2.3cm]{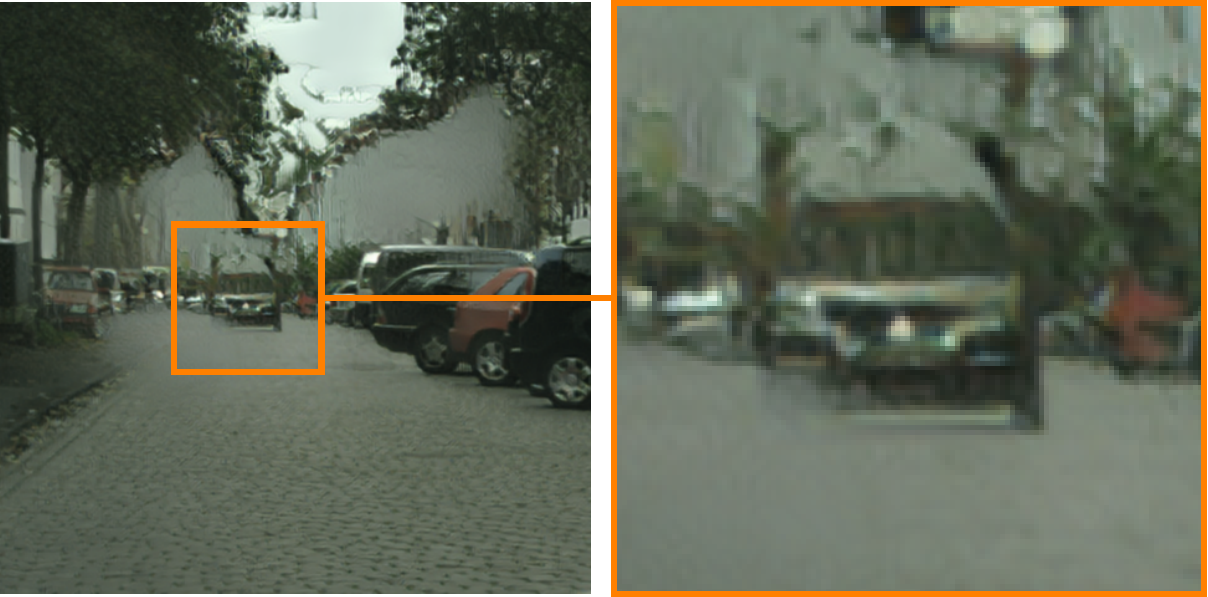}\hfill
\label{}\includegraphics[width=2.3cm]{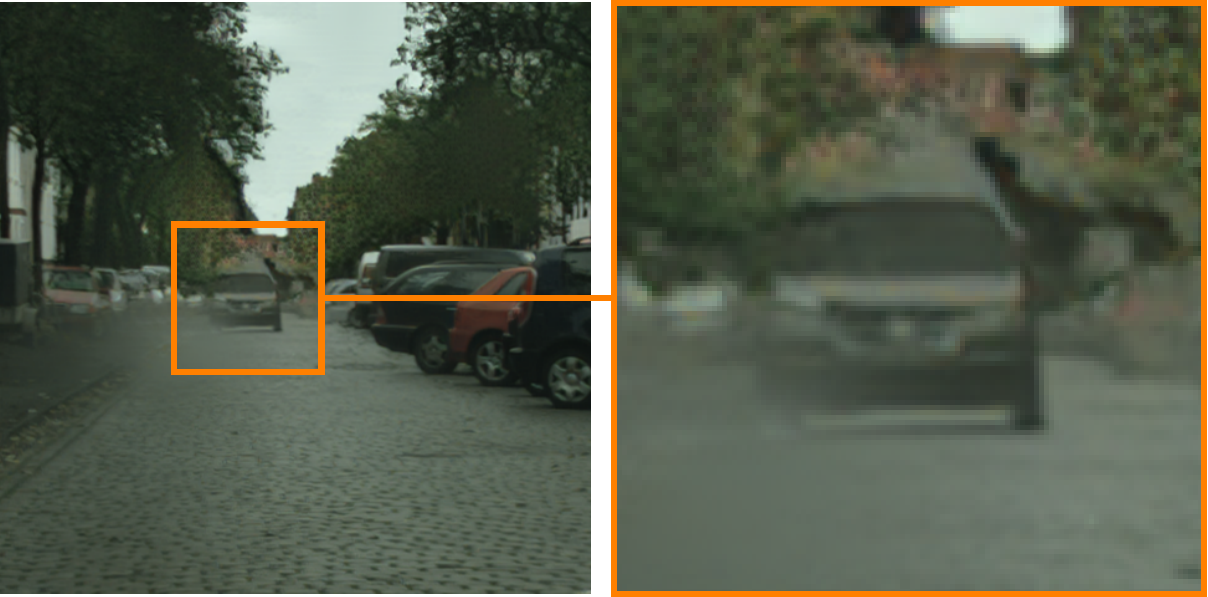}\hfill
\label{}\includegraphics[width=2.3cm]{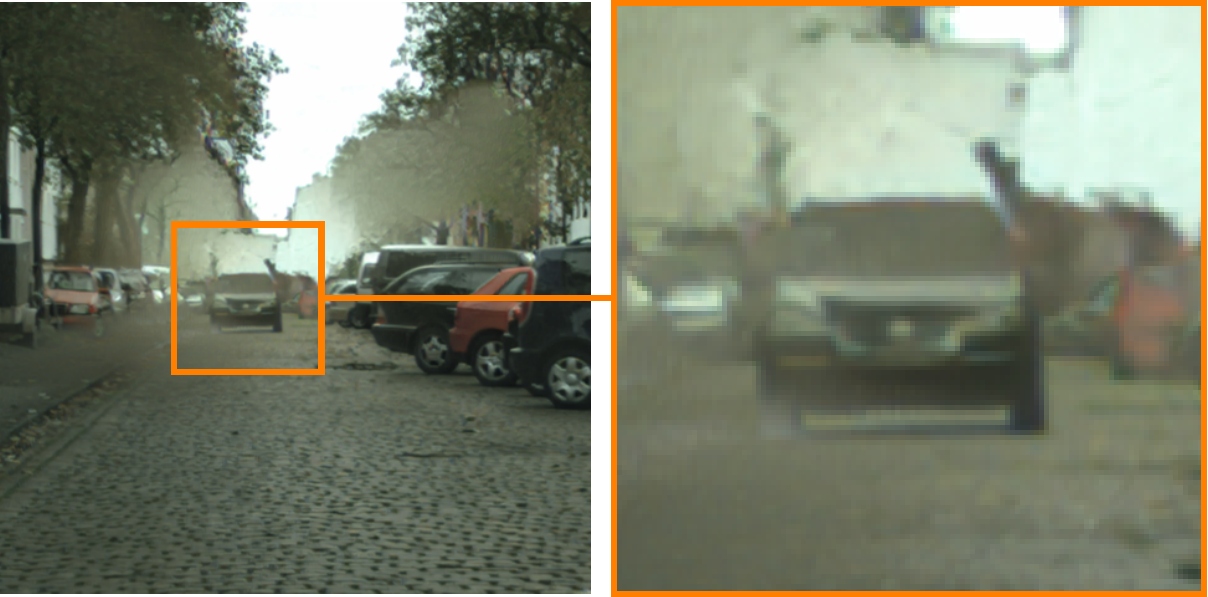}\hfill
\label{}\includegraphics[width=2.3cm]
{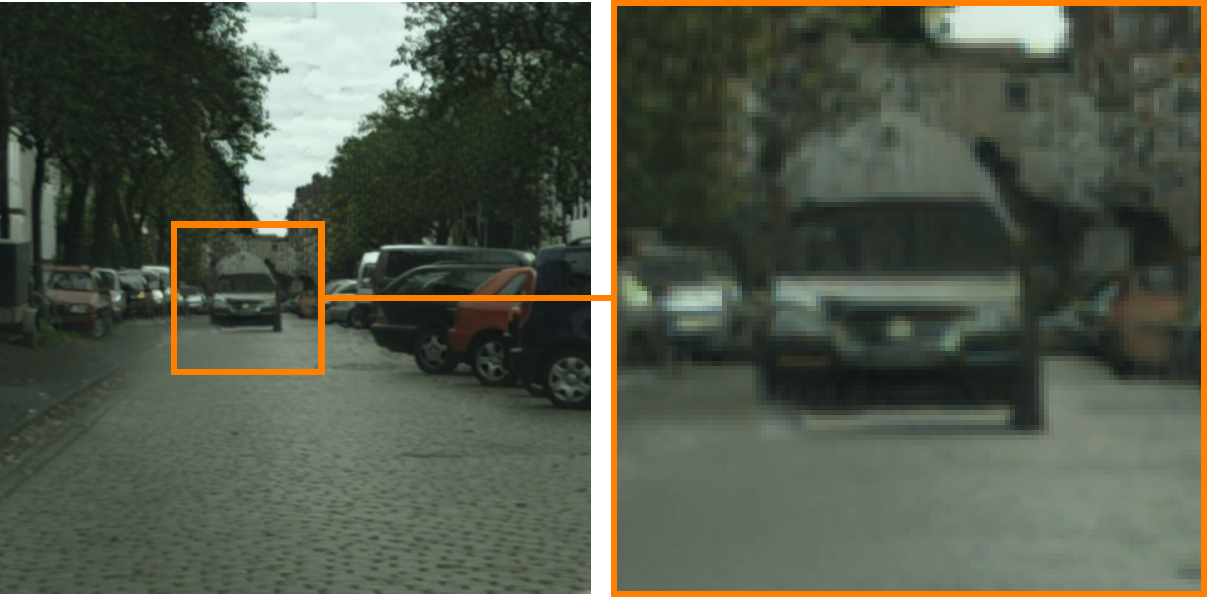} \\

\label{}\includegraphics[width=2.3cm]{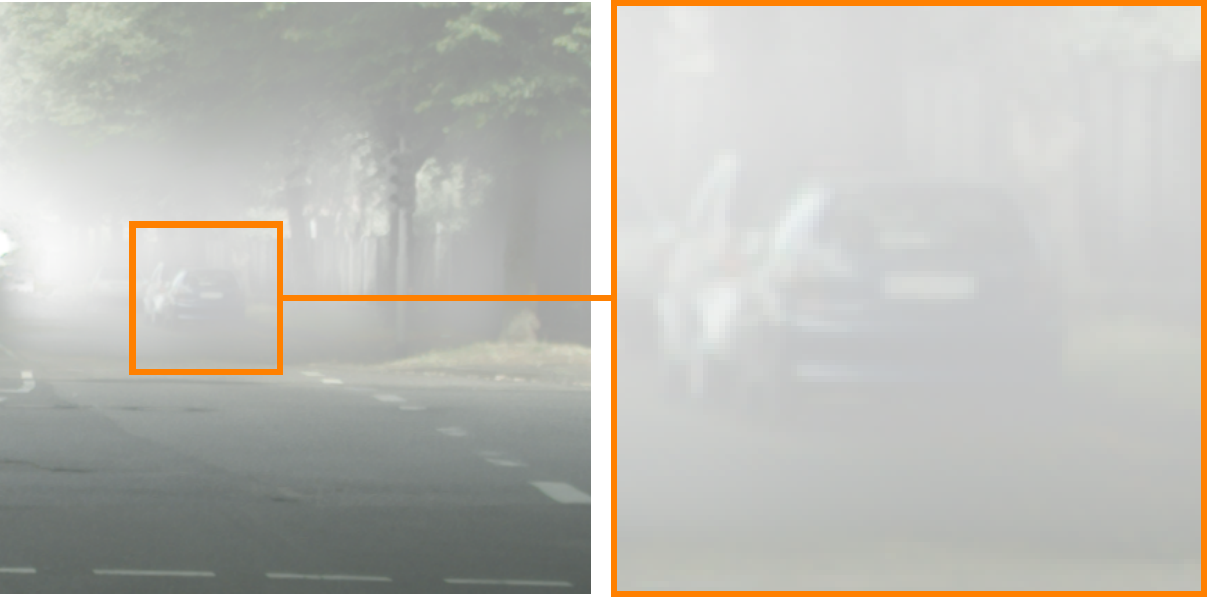}\hfill
\label{}\includegraphics[width=2.3cm]{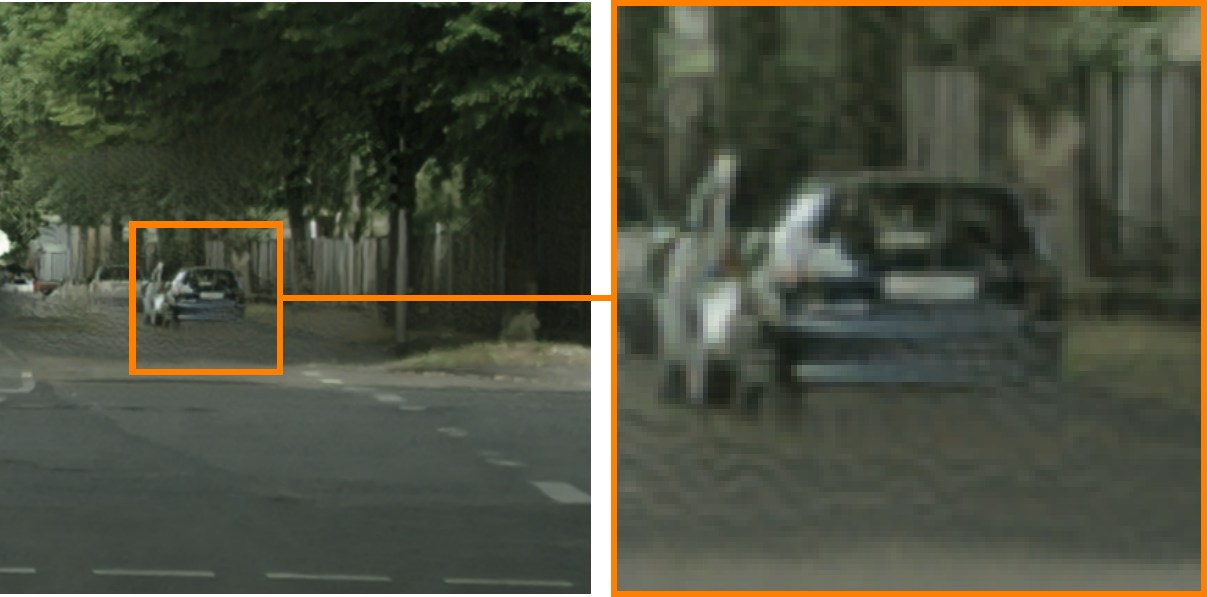}\hfill
\label{}\includegraphics[width=2.3cm]{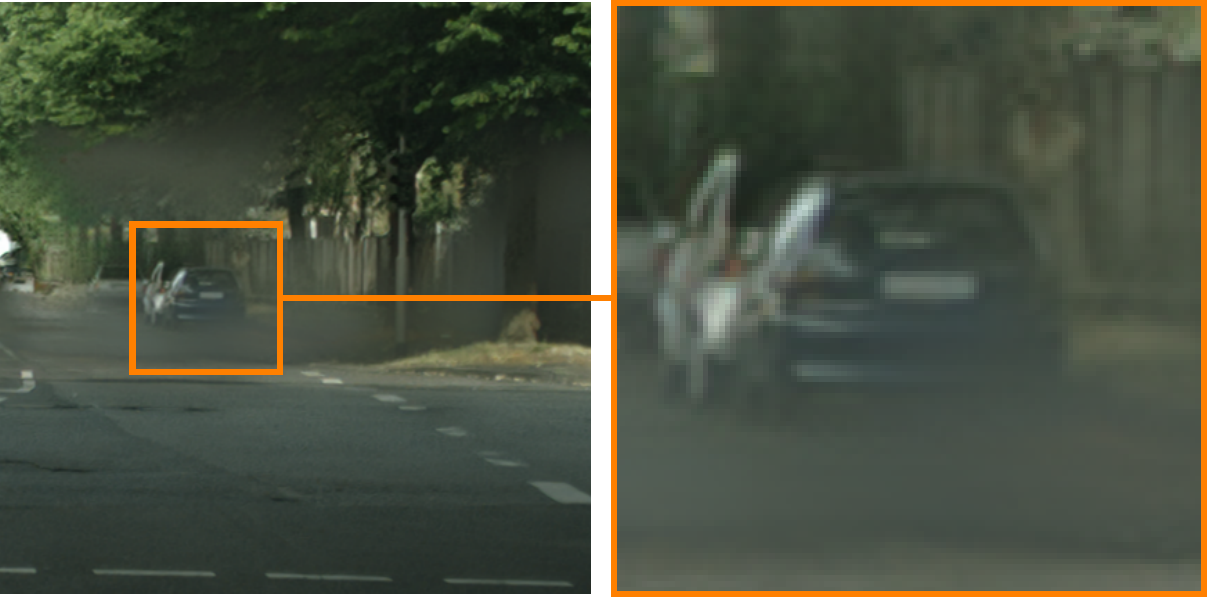}\hfill
\label{}\includegraphics[width=2.3cm]{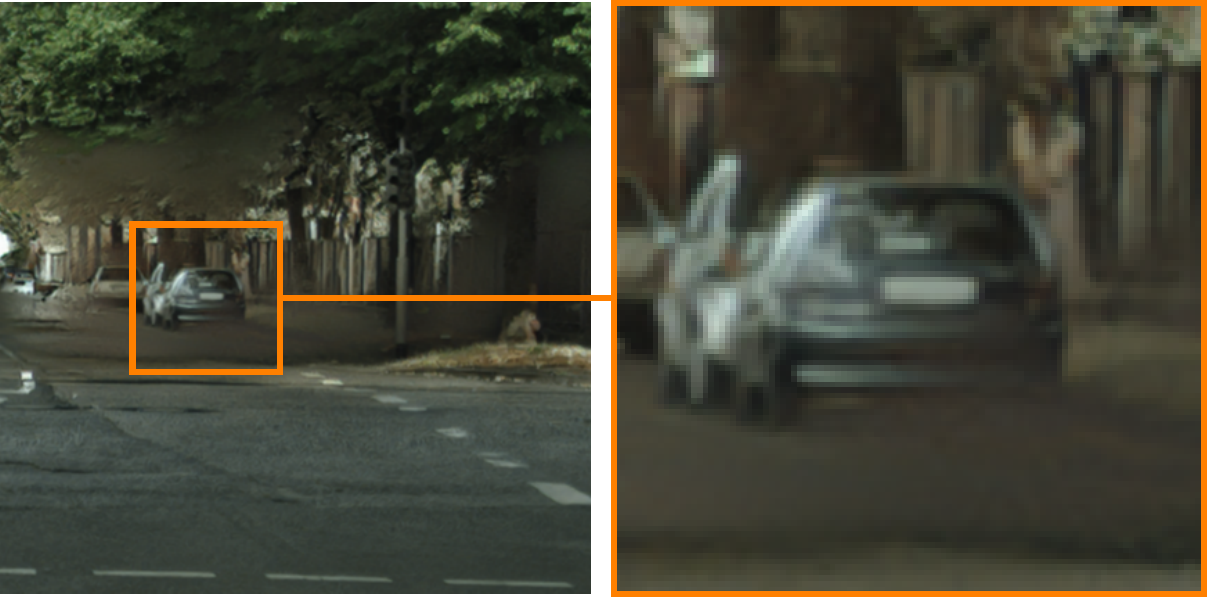}\hfill
\label{}\includegraphics[width=2.3cm]
{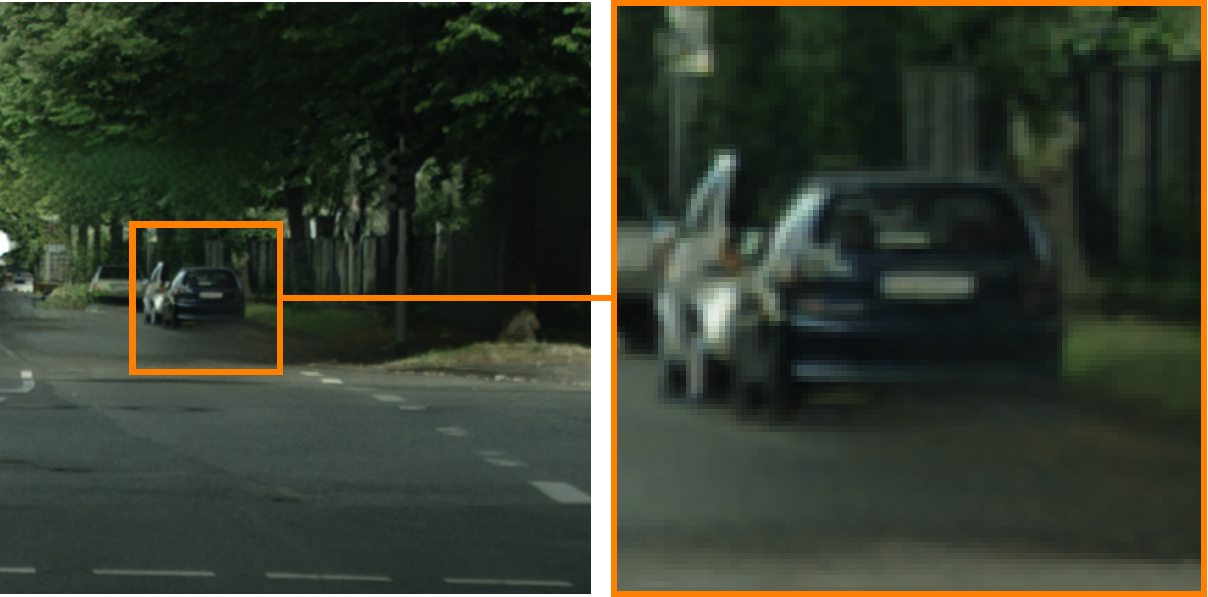} \\

\caption{Qualitative comparison with state-of-the-art contrastive learning based I2I methods. We compare against foggy input (Column 1), CUT~\cite{park2020contrastive} (Column 2), FeSeSim~\cite{zheng_cham_cai_2021} (Column 3), Qs-Attn~\cite{hu_zhou_huang_shi_sun_li_2022} (Column 4), Our method (Column 5).  Our approach achieves better translation in object regions through the proposed attention driven scheme. Best viewed with digital zoom.}
\label{fig:qualitative}
\end{figure}

We compare our translation results with CUT~\cite{park2020contrastive}, FeSeSim~\cite{zheng_cham_cai_2021} and Qs-Attn~\cite{hu_zhou_huang_shi_sun_li_2022} in Fig.~\ref{fig:qualitative}. It may be observed that while previous methods are successful in transferring the global style and appearance, they often struggle to capture instance-level details and result in poor translation quality in local object areas. By identifying salient object regions, our approach guides the translation task to optimise appearance of object instances and achieve superior image quality in the relevant image regions.

\begin{table}[h]

\footnotesize
\setlength\tabcolsep{4pt}
\centering
\begin{adjustbox}{scale=0.88}
\begin{tabular*}{\linewidth}{@{\extracolsep{\fill}} l | c c c c }
\toprule
Method  &  FID $\downarrow$  & KID $\downarrow$  & $\text{FID}_{\textsc{inst}}$  $\downarrow$ & $\text{KID}_\textsc{inst}$ $\downarrow$ \\ 
\hline

CUT$^{*\dag}$~\cite{park2020contrastive}  \textsc{\textit{{}     (ECCV '20) }}            &   0.21 &0.84 &0.61&2.77 \\ 
FeSeSim$^{*\dag}$~\cite{zheng_cham_cai_2021}   \textsc{\textit{{}     (CVPR '21) }}          & 0.20 & 0.74 & \underline{0.51} & \underline{1.97} \\
Qs-Att.$^{*\dag}$~\cite{hu_zhou_huang_shi_sun_li_2022}  \textsc{\textit{{}     (CVPR '22) }}         &  0.20 &  0.83  &0.55&2.23 \\ 
\hline

Ours - supervised         & \textbf{0.18} & \textbf{0.67} & \textbf{0.47} & \textbf{1.44} \\ 
Ours - local-global $^{\dag}$   & \underline{0.19} & \underline{0.70} & \underline{0.51} & 2.02 \\ 
\bottomrule

\end{tabular*}
\end{adjustbox}
\caption{Comparison of recent contrastive learning based image-to-image translation methods, across image quality metrics, under the \mbox{Foggy Cityscapes $\rightarrow$ Cityscapes} setting.}
\label{tab:results:foggy_ours} 
\end{table}

\begin{figure}
\centering
\includegraphics[width=3cm]{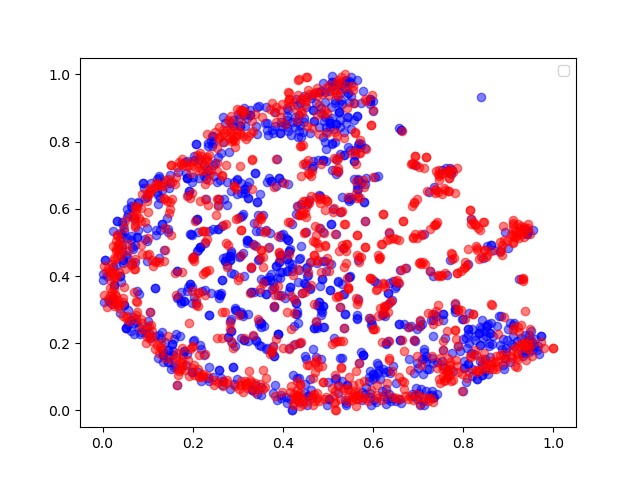}%
\hfill
\includegraphics[width=3cm]{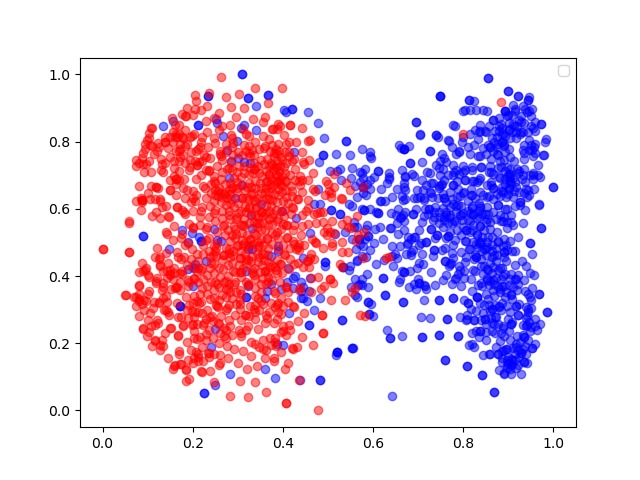}%
\hfill
\includegraphics[width=3cm]{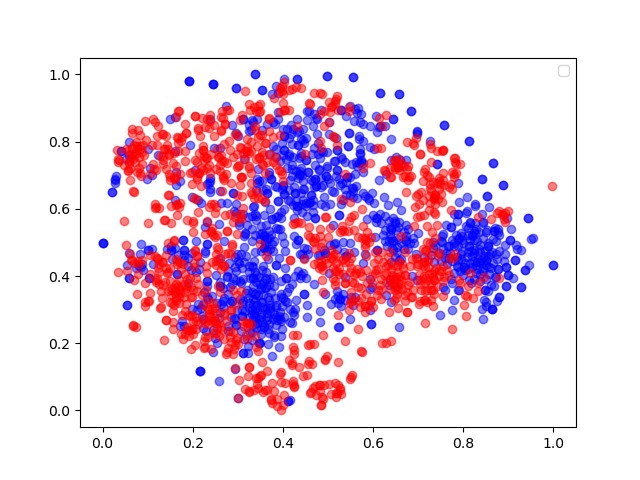}%
\hfill
\caption{Enlarged version of the main paper feature visuliazation via t-SNE. We randomly sample object features corresponding to salient object and background regions. We compare (a) baseline model \mbox{without $G_A$, (b) a model with supervised  $G_A$, (c) a model with self-supervised $G_A$ using the Eq.~\eqref{eq:detco} loss.}}
\label{fig:sup:tsne}
\end{figure}

\subsection{Implementation details}
\label{sec:supplementary:imp}

\textbf{Detector training details} All detectors are trained using identical hyperparameters and settings; we employ Stochastic Gradient Descent (SGD) with base learning rate $0.001$, batch size of $4$ and weight decay of $5{\times}10^{-4}$. Unless otherwise stated, we initialize the models with ImageNet weights and decrease the learning rate after 50000 and 70000 iteration steps with $\gamma=0.1$, for a total of 100,000 training iterations. Following common protocol, all images are resized such that the smallest side length (i.e. width or height) is 600 pixels both during training and test. All models are implemented using Detectron~\cite{wu2019detectron2} and PyTorch libraries ~\cite{NEURIPS2019_9015}.

\textbf{Image translation training details} We build our image-to-image (I2I) translation model using a patchwise multi-layer 
component, similar to~\cite{park2020contrastive}. For fair comparison, all models in Tab.~\ref{tab:results:foggy} and Tab.~\ref{tab:results:foggy_ours} are trained for a total of 400 epochs using an Adam optimizer~\cite{DBLP:journals/corr/KingmaB14} with momentum parameters \mbox{$b1{=}0.5$}, \mbox{$b2{=}0.99$} and an initial learning rate $1e{-}5$. The input images are resized such that the smallest size is 600 pixels during training. We perform inference on full resolution images during test.  

When training the self-supervised local-global translation model, unless stated otherwise, we follow training settings aligned with~\cite{xie_ding_wang_zhan_xu_sun_li_luo_2021}. For data augmentation, we apply random horizontal flip, gaussian
blur and color jittering related to brightness, contrast, saturation, hue and grayscale. Our local patch generation process follows the approach of ~\cite{Misra2019SelfSupervisedLO}. Namely, a random region is firstly cropped such that it covers at least 60\% of the original global image, followed by the aforementioned data augmentation operations. The image is divided into $4{\times}4$ grid areas which are randomly shuffled to obtain the final 16 local patches. Finally, we set weights of Eq.~\eqref{eq:detco} to $0.1$, $0.4$, $0.7$, $1.0$ for objective terms $w_1, w_2, w_3, w_4$ respectively, where each term pertains to a different convolutional layer of networks $G_A$ and $G_{Am}$. 

\textbf{Network Architectures} We denote a network convolutional layer that contains $f$ filters, with stride $x$ and a $y \times y$ kernel size to be a $cf$-$sx$-$ky$ layer. In this notation convention, $c64$-$s1$-$k3$ denotes a convolutional layer that applies 64 filters with a stride of 1 and kernel size $3 \times 3$. Futhermore, we denote a convolutional layer that applies the transposed convolution operation using $f$ filters, a stride of $x$ and kernel size $y \times y$ as $uf$-$sx$-$ky$. Unless stated otherwise, every $cf$-$sx$-$ky$ and $uf$-$sx$-$ky$ layer is followed by a ReLu~{Agarap2018DeepLU} activation function and InstanceNorm~\cite{Ulyanov2016InstanceNT} normalization layer.

We deploy a patchGAN discriminator~\cite{pix2pix2017} $D$ with an architecture that can be denoted by [$c64$-$s2$-$k4$, $c128$-$s2$-$k4$, $c256$-$s2$-$k4$, $512$-$s1$-$k4$]. Accordingly, we model the feature extractor network $E_B$ using layers [$c64$-$s2$-$k7$, $c128$-$s2$-$k3$, $c256$-$s2$-$k3$, $9{\cdot}r256$-$s1$-$k3$] where $9{\cdot}r256$-$s1$-$k3$ denotes 9 residual blocks with 2 convolutional layers, each. We implement network $G_C$ as [$u128$-$s2$-$k7$, $u64$-$s2$-$k3$, $u27$-$s1$-$k7$], where the $u27$-$s1$-$k7$ layer is followed by a $\tanh$ activation function, without a normalization layer. 

For the attention generator $G_A$ we use an architecture denoted by [$u128$-$s2$-$k7$, $u64$-$s2$-$k3$, $u10$-$s1$-$k7$] where the last layer, $u10$-$s1$-$k7$, is followed by a \textit{Softmax} function which generates the attention masks. The supervised model additionally trains two filters $c2$-$s1$-$k7$ in network $G_A$ which produce the object saliency prediction. Our fully self-supervised model follows the same architecture as the supervised model for $E_B$, $G_C$ and $G_A$ with the only exception being the object saliency filters, $c2$-$s1$-$k7$, in $G_A$. In the self-supervised case, momentum networks $E_{Bm}$, $G_{Am}$ follow identical architectural copies of $E_B$, $G_A$, respectively. More specifically, we optimize layers [$u128$-$s2$-$k7$, $u648$-$s2$-$k3$, $u10$-$s1$-$k7$] and [$um{*}128$-$s2$-$k7$, $um{*}64$-$s2$-$k3$, $um{*}10$-$s1$-$k7$] together, where $um{*}$ denotes the corresponding layers of $G_{Am}$. We additionally optimize layers $c256$-$s2$-$k3$ and $cm{*}256$-$s2$-$k3$ together, pertaining to $E_B$ and $E_{Bm}$ respectively, via Eq.~\eqref{eq:detco} of the main paper. We attach 4 global and 4 local MLP heads to each of these layers to obtain the final representations. The set of MLPs are implemented as a set of linear layers followed by ReLU activation function. All experiments are performed on four NVIDIA V100 GPUs, each with 32GB of RAM.



\end{document}